\documentclass[conference]{IEEEtran}
\IEEEoverridecommandlockouts
\usepackage{cite}
\usepackage{hyperref}
\usepackage{amsmath,amssymb,amsfonts}
\usepackage{algorithmic}
\usepackage{graphicx}
\usepackage{textcomp}
\usepackage{xcolor}
\usepackage{tabularray}
\usepackage[normalem]{ulem}
\def\hrulefill{\leavevmode\leaders\hrule height 1pt\hfill\kern0pt}

\usepackage[utf8]{inputenc} % allow utf-8 input
\usepackage[T1]{fontenc}    % use 8-bit T1 fonts
\usepackage{hyperref}       % hyperlinks
\usepackage{url}            % simple URL typesetting
\usepackage{amsfonts}       % blackboard math symbols
\usepackage{nicefrac}       % compact symbols for 1/2, etc.
\usepackage{microtype}      % microtypography

\usepackage{booktabs}
\usepackage{multirow}
\usepackage{subcaption}
\usepackage{utfsym} %打x
\usepackage{pifont} %打勾
\usepackage{multirow}
\usepackage{fontawesome} 
\usepackage{listings}
\usepackage{multicol}
\usepackage{helvet}
\usepackage{amssymb}
\usepackage{booktabs}
\usepackage[export]{adjustbox}
\usepackage[most]{tcolorbox}
\usepackage{mdframed}

\PassOptionsToPackage{table}{xcolor}
\usepackage[table]{xcolor}
\usepackage{colortbl}
\definecolor{c0}{HTML}{D3D3D3}
\definecolor{c1}{HTML}{0D92F4}
\definecolor{c2}{HTML}{EC8305}
\definecolor{c3}{HTML}{347928}
\definecolor{c4}{HTML}{FBFBFB}
\definecolor{c5}{HTML}{31511E}
\definecolor{c31}{HTML}{8B5DFF}
\definecolor{c32}{HTML}{6A42C2}
\definecolor{c33}{HTML}{563A9C}

\def\hrulefill{\leavevmode\leaders\hrule height 1pt\hfill\kern0pt}

\usepackage{marvosym}

\usepackage{subfiles}

\newcounter{changecounter}

\DeclareRobustCommand{\changemark}[2][]{%
  \refstepcounter{changecounter}% 使用 refstepcounter 而不是 stepcounter
  \if\relax\detokenize{#1}\relax
  \else
    \label{changemark:#1}% 使用特定前缀避免冲突
  \fi
  \begingroup
  \setlength{\fboxsep}{1pt}%
  \endgroup
  \hspace{0.05em}%
  #2
}
\DeclareRobustCommand{\changetable}[1]{%
  #1
}

\def\BibTeX{{\rm B\kern-.05em{\sc i\kern-.025em b}\kern-.08em
    T\kern-.1667em\lower.7ex\hbox{E}\kern-.125emX}}
\begin{document}
\IEEEoverridecommandlockouts
% \subfile{XRAG-Response-Letter}

\title{XRAG: eXamining the Core - Benchmarking Foundational Components in Advanced Retrieval-Augmented Generation\\ }
% \title{[Experiment, Analysis, and Benchmark] XRAG: eXamining the Core - Benchmarking Foundational Components in Advanced Retrieval-Augmented Generation\\ }

\author{
\IEEEauthorblockN{
Qili Zhang\IEEEauthorrefmark{2}\IEEEauthorrefmark{1}, 
Qianren Mao\Letter\IEEEauthorrefmark{2}
\thanks{This work is supported by Zhongguancun Laboratory.}
\thanks{\Letter Corresponding author.}, 
Yangyifei Luo\IEEEauthorrefmark{1}, 
Yashuo Luo\IEEEauthorrefmark{1},
Hanwen Hao\IEEEauthorrefmark{1}, \\
Zhilong Cao\IEEEauthorrefmark{1},
Weifeng Jiang\IEEEauthorrefmark{3}, 
Zhijun Chen\IEEEauthorrefmark{1},   
Junnan Liu\IEEEauthorrefmark{1},  \\
Feng Yan\IEEEauthorrefmark{1},
Xiaolong Wang\IEEEauthorrefmark{1},
Jinlong Zhang\IEEEauthorrefmark{1}, 
Zhenting Huang\IEEEauthorrefmark{1},\\
Zhixing Tan\IEEEauthorrefmark{2},  
Jie Sun\IEEEauthorrefmark{2},
Bo Li\IEEEauthorrefmark{1},
Jianxin Li\IEEEauthorrefmark{1},
and Philip S. Yu\IEEEauthorrefmark{4}
}
\IEEEauthorblockA{
\IEEEauthorrefmark{2}Zhongguancun Laboratory
\IEEEauthorrefmark{1}Beihang University
\IEEEauthorrefmark{3}Nanyang Technological University
\IEEEauthorrefmark{4}University of Illinois Chicago
}
\IEEEauthorblockA{
zhangqili@buaa.edu.cn, maoqr@zgclab.edu.cn, \{luoyangyifei, luoyashuo, haohanwen\}@buaa.edu.cn \\
zhilongcao66@gmail.com, weifeng001@e.ntu.edu.sg, zhijunchen@buaa.edu.cn, to.liujn@outlook.com \\
feng.yan@buaa.edu.cn, 3170153878@qq.com, zhangjinlong@buaa.edu.cn, zhentinghng@gmail.com \\
\{zxtan, sunjie\}@zgclab.edu.cn, 5481463@qq.com, lijx@buaa.edu.cn, psyu@uic.edu
}
}

\maketitle

\begin{abstract}
  Retrieval-augmented generation (RAG) synergizes the retrieval of pertinent database with the generative capabilities of Large Language Models (LLMs), ensuring that the generated output is contextually relevant but also accurate and current.
  We introduce XRAG, an open-source, modular codebase that facilitates exhaustive evaluation of the performance of foundational components of advanced RAG modules. 
  These components are categorized into four core phases: pre-retrieval, retrieval, post-retrieval, and generation.  
  We systematically analyze them across reconfigured datasets, providing a comprehensive benchmark for their effectiveness.  
  These components can be combined in different ways through an orchestrator, of which there are five types: sequential, conditional, iterative, parallel, and hybrid.
  Our work thoroughly evaluates the performance of advanced core components in RAG systems, offering actionable insights for optimizing RAG architectures in \changemark[change_abstract]{dataset-driven} AI applications.
\end{abstract}

\begin{IEEEkeywords}
Database, Retrieval-Augmented Generation, Evaluation, Benchmark, Data Management
\end{IEEEkeywords}

\section{Introduction}
\label{Introduction}
Retrieval-Augmented Generation (RAG)~\cite{JiangXGSLDYCN23,abs-2002-08909,abs-2312-10997,BorgeaudMHCRM0L22,AsaiWWSH24,singh2025agenticretrievalaugmentedgenerationsurvey,abs-2501-00309,mao2025privacy} represents a pivotal strategy in Q\&A tasks, demonstrating enhanced performance by delivering more informative and accurate answers compared to relying solely on large language models (LLMs). 
The efficacy of basic RAG systems~\cite{du2024stackingtransformerscloserlook, lu2024turboragacceleratingretrievalaugmentedgeneration} is contingent depends on the seamless operation of four core components: pre-retrieval, retrieval, post-retrieval, and generation.  
The pre-retrieval stage indexes the corpus and reforms queries for efficient retrieval. 
The retrieval stage focuses on identifying and extracting documents relevant to a given query.
The post-retrieval stage refines, summarizes, or compacts information to ensure contextual clarity. 
Finally, the generation stage employs the LLM to produce responses. 
In naive RAG systems, these stages are executed in a sequential manner.
However, when dealing with complex retrieval scenarios, naive RAG often suffers from inefficiency and incomplete retrieval. 
Therefore, advanced RAG systems have optimized the retrieval process instead of sequential.
Five RAG processes are as follows: sequential, conditional, iterative, parallel, and hybrid.
These stages and processes critically influence output quality, highlighting the RAG framework's interdependence.
% Additionally, advanced RAG modules (e.g., reranker, refiner) offer sophisticated algorithms for tailored search solutions, surpassing standardized methodologies. 

\begin{table*}[t]
  \centering
  % \footnotesize
  \scriptsize
\caption{Comparison of RAG Libraries. 
\texttt{{\footnotesize Modular Design}} (Mod.Dsgn) indicates toolkit modularity. 
\texttt{{\footnotesize Fair Comparison}}~\cite{zhang2024raglab} (Fair.Comp) indicates evaluation by aligning key components like seeds, generators, retrievers, and instructions.  
\texttt{{\footnotesize Unified Datasets}} (Unif.Data)  ensures unified dataset formats for retrieval and generation. 
\texttt{{\footnotesize Modular Evaluation}} (Mod.Eva) assesses RAG modular differences. 
% \texttt{{\footnotesize Failure Management}} (Fail.Mgmt) systematically implements strategies for identifying and mitigating RAG failure points.
ConR uses token-matching for evaluating retrieval. ConG uses token-matching for evaluating generation. CogL is based on LLM-based instructions for retrieval and generation evaluation. `u' refers to No. of unified metrics.}
\renewcommand\arraystretch{1.2}
\setlength{\tabcolsep}{6.38mm}{
  \begin{tabular}{@{}|l|c|c|c|c|c|c|c|}
    \specialrule{1pt}{0pt}{0pt}
  \textbf{{RAG Library}} 
  & \textbf{Mod.Dsgn}
  & \textbf{Fair.Comp}
  & \textbf{Unif.Data}
  & \textbf{Mod.Eva}
  % & \textbf{Fail.Mgmt}
  & \textbf{ConR}
  & \textbf{ConG}
  & \textbf{CogL}
  \\ \hline
  LangChain~\cite{Chase_LangChain_2022}  
  & \ding{52}  & \usym{2717} & \usym{2717} & \usym{2717}  & 0 & 0 & 0 \\ 
  LlamaIndex~\cite{Liu_LlamaIndex_2022} 
  & \ding{52}  & \usym{2717} & \usym{2717} & \ding{52}  &  6 &  0  &  7 \\ 
  FastRAG~\cite{Izsak_fastRAG_Efficient_Retrieval_2023}    
  & \ding{52} & \usym{2717}  & \usym{2717} &  \usym{2717}   & 0 & 0 & 0  \\ 
  RALLE~\cite{HoshiMNTMTD23}    & \ding{52} & \usym{2717} & \usym{2717} & \usym{2717} &  0  & 0 & 0 \\ 
  LocalRQA~\cite{abs-2403-00982} & \usym{2717}  & \usym{2717} & \ding{52} & \ding{52}  &  3 & 2 &  1 \\ 
  AutoRAG~\cite{bwookkim2024} & \ding{52} & \usym{2717}  & \ding{52} & \ding{52}  &  6 &  5 &  4\\ 
  FlashRAG~\cite{abs-2405-13576}  & \ding{52} & \usym{2717} & \ding{52} & \ding{52} &   4  &  5 & 0 \\ 
  RAGLAB~\cite{zhang2024raglab} & \ding{52} & \ding{52} & \ding{52} & \usym{2717} & 0 &   4 & 0 \\ 
  \textbf{XRAG} (ours) & \ding{52} & \ding{52} & \ding{52} & \ding{52}  & \(\textcolor{c5}{\textbf{7}^{ u}}\) & \(\textcolor{c5}{\textbf{10}^{ u}}\) & \(\textcolor{c5}{\textbf{23}^{ u}}\) \\
  \specialrule{1pt}{0pt}{0pt}
  \end{tabular}
\label{Comparation}
}
\end{table*}

In real-world applications characterized by complex dataset structures, existing RAG frameworks exhibit significant limitations.
% Toolkits like LangChain~\cite{Chase_LangChain_2022} and LlamaIndex~\cite{Liu_LlamaIndex_2022}, modularize the RAG process, increasing adaptability and broadening its applications. 
Modular RAG toolkits like LangChain~\cite{Chase_LangChain_2022} and LlamaIndex~\cite{Liu_LlamaIndex_2022} are typically cumbersome, making adaptation to new data challenging and validating or optimizing innovative methods inconvenient. 
% FastRAG~\cite{Izsak_fastRAG_Efficient_Retrieval_2023} and RALLE~\cite{HoshiMNTMTD23} allow users to assemble RAG systems with core components, fostering a more adaptable RAG implementation. 
% AutoRAG~\cite{bwookkim2024} further supports users by identifying optimal RAG pipelines for custom data, facilitating bespoke RAG systems. 
% LocalRQA~\cite{abs-2403-00982} and RAGLAB~\cite{zhang2024raglab} focus on RAG training, offering scripts for various component training. 
Efficient RAG Frameworks like FastRAG~\cite{Izsak_fastRAG_Efficient_Retrieval_2023}, RALLE~\cite{HoshiMNTMTD23}, AutoRAG~\cite{bwookkim2024}, and LocalRQA~\cite{abs-2403-00982} require users to independently reproduce published algorithms and offer limited component options, restricting the flexibility of RAG systems despite modular designs.  
% FlashRAG~\cite{abs-2405-13576} and RAGLAB~\cite{zhang2024raglab} advance algorithmic reproducibility in RAG systems by integrating numerous algorithms into a unified framework. This supports efficient replication of existing methods and promotes innovation in algorithm development. 
In module evaluation tasks, FlashRAG~\cite{abs-2405-13576} lacks uniformity in fundamental evaluation components, such as random seeds, generators, retrievers, and instructions, which hinders result comparability. 
RAGLAB~\cite{zhang2024raglab}, despite offering a fair experimental setup, lacks comprehensive evaluation strategies for assessing individual RAG components. 
% Similarly, FlashRAG~\cite{abs-2405-13576} exhibits the same lack of modularity analysis, restricting to derive meaningful conclusions.  
Although ongoing efforts address these challenges through modular RAG processes (e.g., retrieval engines and generative agents), an implicit gap persists in the comparative performance evaluations of these advanced RAG modules within the overall RAG workflow. 
Comprehensive assessments of these modules are notably absent, making it difficult for researchers to evaluate their approaches in consistent experimental conditions.

To address the above issues, we concentrate on the core components of advanced RAG modules and the orchestrators of agentic RAG, conducting comprehensive experiments.
We introduce XRAG~\footnote{Code available at: \url{https://github.com/DocAILab/XRAG}.}, an open-source, modular codebase designed to evaluate foundational components of advanced RAG modules comprehensively. 
% The key capabilities of XRAG are summarized as follows:
XRAG offers four principal capabilities:

\textbf{Modular RAG Architecture: Fine-Grained Comparative Analysis.}
Extensive experiments are conducted on the advanced RAG modules in four stages: pre-retrieval, retrieval, post-retrieval, and generation. 
The core components cover three query rewriting strategies, six retrieval units, three post-processing techniques, and LLM generators from different vendors — OpenAI, Meta, and DeepSeek. 
XRAG provides an in-depth understanding of the capabilities of RAG components.

\textbf{Agentic RAG Process: Multi-Orchestrator Integration and Workflow Design.}
% Furthermore, we conducted conventional generation evaluatio and analysis on all five categories of orchestrators.
The XRAG architecture incorporates five orchestrators: sequential, conditional, iterative, parallel, and hybrid.
It provides a streamlined framework to flexibly orchestrating agentic RAG processes and integrate advanced RAG modules.
Furthermore, to guide the development of RAG systems across various scenarios and data types, XRAG conducts performance evaluations of its orchestrators.

\begin{figure*}[htbp]
\centerline{\includegraphics[width=0.85\textwidth]{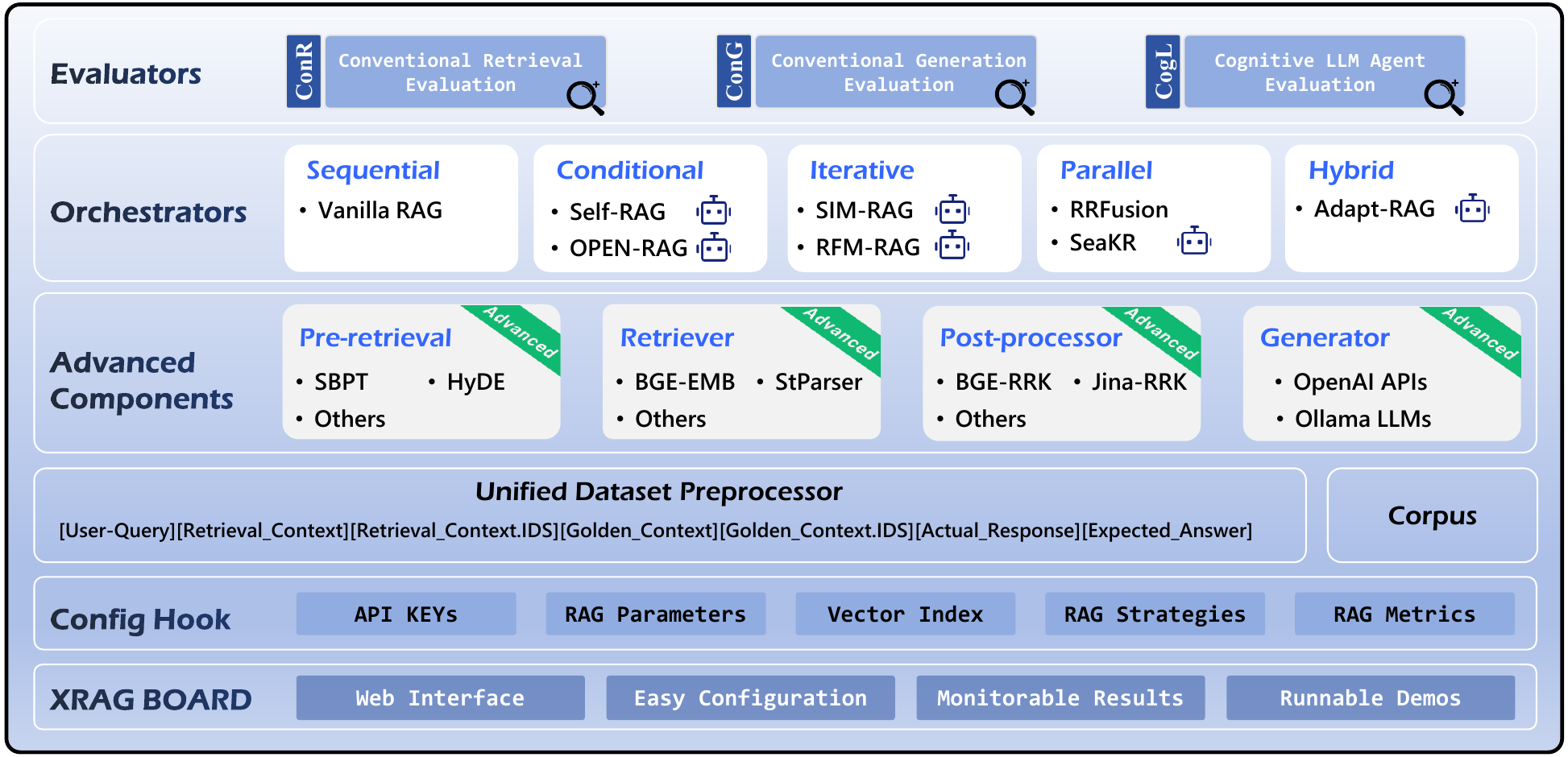}}
    \caption{Schematic overview of the XRAG framework. Icon ``\includegraphics[height=0.9em, valign=B]{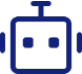}'' indicates that this is an agent-based method.}
    \label{framework}
\end{figure*}

\textbf{Unified Benchmark Datasets: Dual Assessment of Retrieval and Generation.}
To enhance the uniformity of datasets in RAG research, XRAG compiles and formats four prevalent benchmark datasets, preprocessing them into a unified format. 
This standardization enables concurrent assessment of both retrieval and generation, streamlining comparative evaluations across RAG modules and orchestrators.

\textbf{Comprehensive Testing Methodologies: Multidimensional Evaluation Framework.} 
To overcome the absence of a holistic evaluation system for RAG components, XRAG introduces an evaluation benchmark encompassing three perspectives thorough evaluation of retrieval and generation. . 
It comprises Conventional Retrieval Evaluation for retrieval-unit matching,  Conventional Generation Evaluation for generation tests based on generative-token matching, and Cognitive LLM  Evaluation for generation tests based on semantic understanding.

% XRAG ensures a standardized and thorough evaluation of retrieval and generation. 

% \textbf{Identification and Mitigation of RAG Failure Points: Systematic Analysis and Improvement.} 
% Recognizing the lack of systematic experiments and improvement methods addressing RAG failure points, XRAG develops a set of evaluation methods to pinpoint and rectify specific issues. 
% % Targeted enhancement strategies are proposed and employed to verify the resolution of identified problems. 
% Analyzing failure points and implementing feasible optimization and validation solutions can bolster the optimization of RAG components.

\begin{figure}[h]
  \centering
      \includegraphics[width=0.45\textwidth]{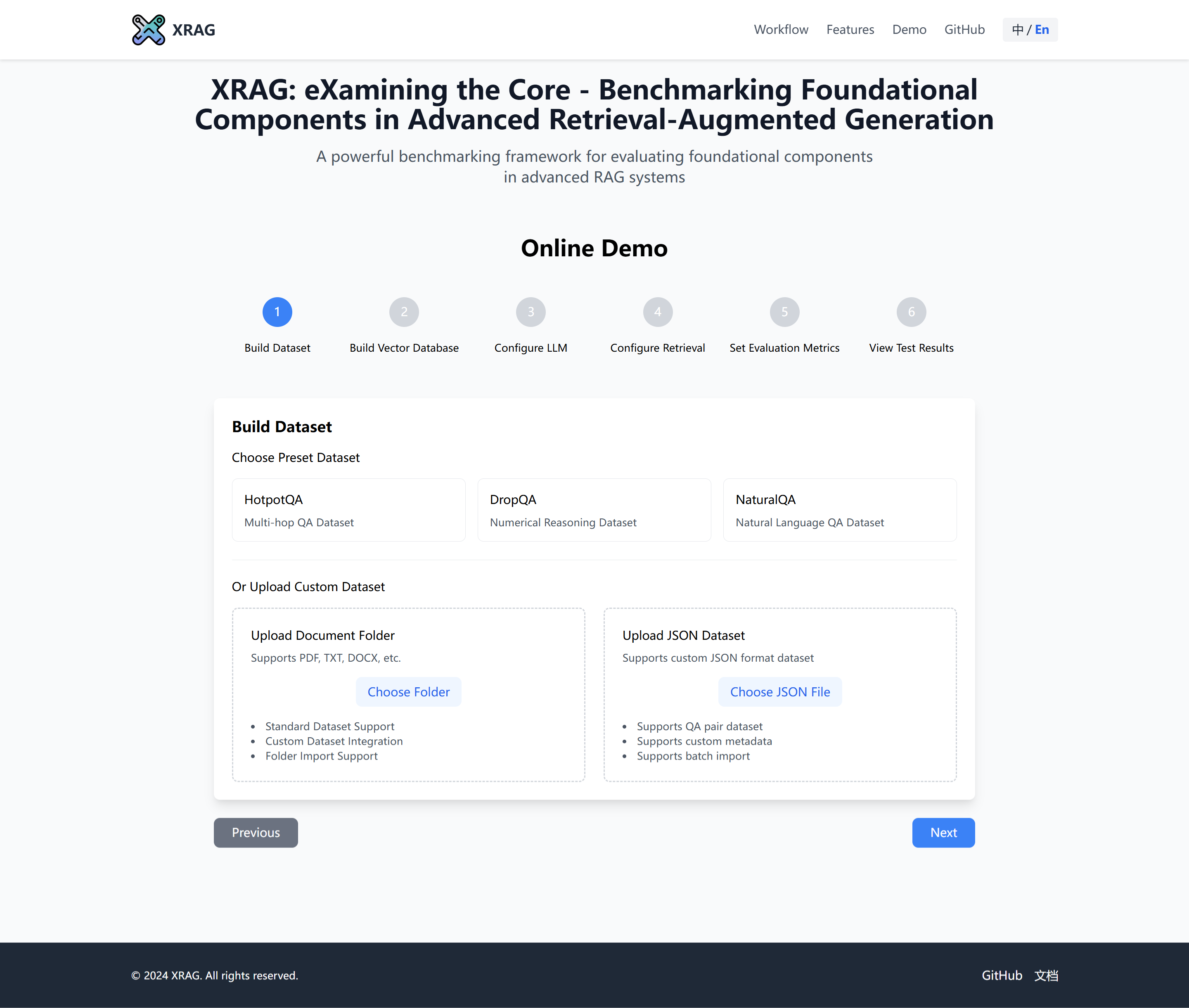}
      \caption{A Screenshot of the Development Web UI of XRAG. }
      \label{demo_pics}
\end{figure}

\section{Related Work}
\label{RELATED WORK}

\subsection{RAG used for Agentic Data Management}
% The field of data management has a rich history of developing benchmarks to ensure the fair and comparable evaluation of system performance.
% A core tenet of this discipline is that robust evaluation must go beyond simple end-to-end metrics and provide fine-grained insights into individual components, much like analyzing query optimizers separately from retrieval models.
In data management, benchmarks ensure fair performance evaluation by providing detailed insights beyond end-to-end metrics into components like query optimizers and retrieval models in RAG systems.
While RAG systems are often viewed through the lens of natural language processing, their effectiveness is fundamentally governed by the quality, organization, and accessibility of the underlying knowledge corpus—a classic data management concern~\cite{DBLP:journals/pvldb/KimSAP25}. 
Existing methods in database systems for AI, such as SAGE~\cite{DBLP:conf/icde/ZhangLS25}, ARAG~\cite{DBLP:conf/icde/XiaoCJZ25}, and scRAG~\cite{DBLP:conf/icde/MaoZLLYG25}, often lack standardized datasets and metrics to specifically diagnose the retrieval component. 
Our work bridges this gap by introducing a benchmark framework that applies data management's rigorous evaluation principles to the RAG pipeline, 
enabling a systematic assessment of how data curation, indexing, and retrieval strategies ultimately impact the quality of generated answers.
In certain database systems, such as HTAP databases~\cite{DBLP:journals/tkde/ZhangLZZF24,DBLP:journals/pvldb/ZhangLL24} and DB-GPT~\cite{DBLP:journals/dase/ZhouSL24}, XRAG can handle unstructured data and leverage RAG technology for retrieval and evaluation, enabling these database systems to be applied to scenarios such as generation and inference.

\subsection{RAG Systems or Toolkits}

Retrieval-Augmented Generation~\cite{JiangXGSLDYCN23,abs-2002-08909,abs-2312-10997,BorgeaudMHCRM0L22,AsaiWWSH24,singh2025agenticretrievalaugmentedgenerationsurvey,abs-2501-00309,mao2025privacy} has been widely adopted as an effective technique for mitigating hallucination issues in large language models across various complex generation and reasoning tasks. 
These tasks typically require retrieving critical information from large-scale databases to ensure accurate outcomes and logically sound reasoning processes. Toolkits like LangChain~\cite{Chase_LangChain_2022} and LlamaIndex~\cite{Liu_LlamaIndex_2022}, modularize the RAG process, increasing adaptability and broadening its applications. 
% However, they are typically cumbersome,  making adaptation to new data challenging and validating or optimizing innovative methods inconvenient. 
FastRAG~\cite{Izsak_fastRAG_Efficient_Retrieval_2023} and RALLE~\cite{HoshiMNTMTD23} allow users to assemble RAG systems with core components, fostering more flexible implementations. 
AutoRAG~\cite{bwookkim2024} further supports users by identifying optimal RAG pipelines for custom data, facilitating bespoke RAG systems. LocalRQA~\cite{abs-2403-00982} and RAGLAB~\cite{zhang2024raglab} focus on RAG training, offering scripts for various component training. 
% Nevertheless, FastRAG, RALLE, AutoRAG, and LocalRQA require users to reproduce published algorithms independently and offer limited component options, restricting the flexibility of RAG systems despite modular designs.  
FlashRAG~\cite{abs-2405-13576} and RAGLAB~\cite{zhang2024raglab} advance algorithmic reproducibility in RAG systems by integrating numerous algorithms into a unified framework, supporting efficient replication of existing methods and promoting innovation in algorithm development. 
% However, FlashRAG~\cite{abs-2405-13576} lacks uniformity in fundamental evaluation components, such as random seeds, generators, retrievers, and instructions, which hinders result comparability. 
% RAGLAB~\cite{zhang2024raglab}, despite offering a fair experimental setup, lacks comprehensive evaluation strategies for assessing individual RAG components. 
% Similarly, FlashRAG~\cite{abs-2405-13576} exhibits the same lack of modularity analysis, restricting to derive meaningful conclusions.  
% Although ongoing efforts address these challenges through modular RAG processes (e.g., retrieval engines and generative agents), an implicit gap persists in the comparative performance evaluation of these advanced RAG modules within the overall RAG workflow. 
% Comprehensive assessments of these modules are notably absent, making it challenging for researchers to evaluate their approaches in consistent experimental conditions. 
Despite their restricted scalability, undeveloped evaluation protocols, absent agentic RAG analysis, and non-standardized metrics, modular RAG frameworks pervade current engineering practice.
Meanwhile, researchers are progressively modularizing RAG workflows to address the dataset retrieval requirements in large language model generation tasks.
Inspired by these modular RAG frameworks, we introduce XRAG to evaluate foundational components of advanced RAG modules comprehensively.

\section{XRAG}
\label{XRAG}
Fig.~\ref{framework} delineates the integrated modules and schematic structure of the XRAG framework. 
The framework is stratified into datasets and corpus, advanced components, orchestrator, and evaluators, integrated through XRAG's board and config hook
\footnote{
As shown in the Fig.~\ref{demo_pics}, we have implemented the XRAG demo. These two parts are mainly functions within the demo and are not the focus of the experimental analysis.
}. 
The overall structure progresses from foundational to application-oriented components. 
Designed with a modular architecture, XRAG enables users to accomplish: preparation of normalized RAG datasets (Section~\ref{Unified_Datasets}), assembly of the RAG components (Section~\ref{Advanced_Component_Modules}), orchestration of the RAG workflow (Section~\ref{Advanced_Orchestrator_Modules}) and evaluation of the RAG system's core components (Section~\ref{Evaluation_Methods}).

\begin{figure*}[htbp]
\centerline{\includegraphics[width=1.0\textwidth]{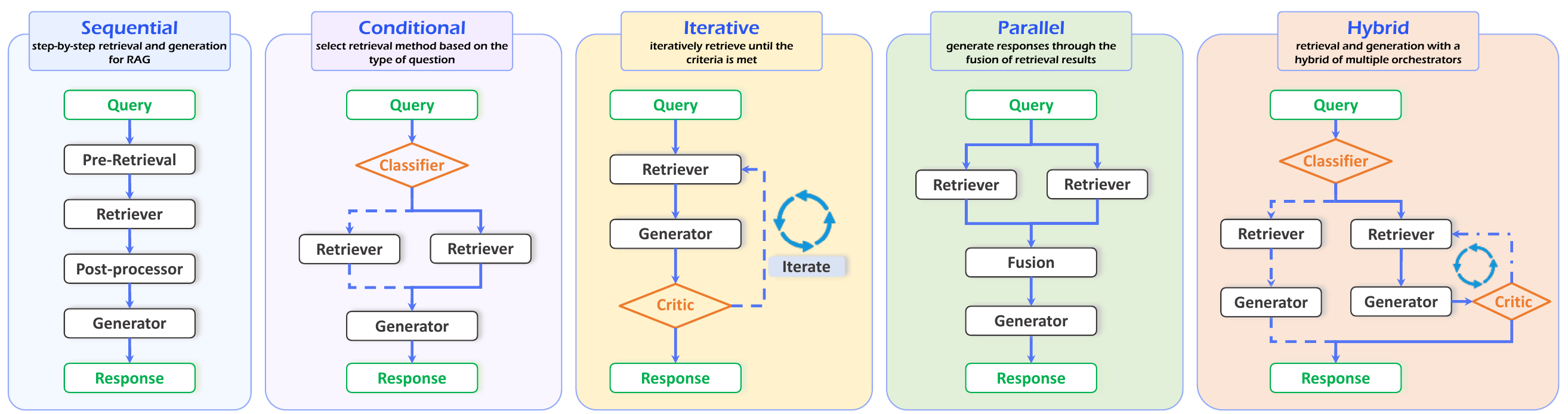}}
    \caption{Schematic of the Orchestrator's Logic in Coordinating Agentic RAG System Components.}
    \label{orchestrator}
\end{figure*}

\subsection{Basic \& Advanced RAG Components}
\label{Advanced_Component_Modules}
\textbf{\({\clubsuit}\) Pre-retrieval} 
Before retrieval, the pre-retrieval components leverage LLMs to refine user queries, enhancing the quality and relevance of the information retrieval process. 
Key methodologies include Step-back Prompting (\texttt{SBPT}~\cite{ZhengMCCCLZ24}): Broaden's queries to enrich contextual grounding for answers, enhancing the contextual foundation for answer generation. 
Hypothetical Document Embedding (\texttt{HyDE}~\cite{GaoMLC23}): Transmutes the original query into a form that better aligns with the indexed documents, improving retrieval alignment and efficacy.  

\textbf{\({\spadesuit}\) Retriever} 
%% TODO:这里不应该有RRFusion。
We use two basic retrieval models as benchmarks: BGE-Large and JINA-Large, along with their open-source and consistently embedding models.
For advanced retrieval strategies, we integrate the LlamaIndex to facilitate standard advanced methods. 
Reciprocal Rerank Fusion Retriever (\texttt{RRFusion}~\cite{CormackCB09}) fuses indexes with a BM25-based retriever, capturing both semantic relations and keyword relevance. 
Both retrievers assign scores, enabling reciprocal reranking for node sorting without additional models or excessive computation.  
For dense retrieval, SentenceWindow Retriever (\texttt{StParser}) parses documents into single sentences per node, incorporating surrounding sentences for added context.

\textbf{\({\blacktriangledown}\) Post-processor} 
To improve retrieval accuracy and efficiency, XRAG uses post-processors like rerankers to refine nodes before returning them. We integrate the BGE reranker (BGE-RRK), which computes similarity scores using a Cross-Encoder model. 
Additionally, JINA-Reranker-V2 (JINA-RRK) supports high-accuracy multilingual document reranking. 

\textbf{\({\blacktriangle}\) Generator} 
The XRAG framework integrates various LLM generators, including those from HuggingFace Transformers APIs~\footnote{\url{https://github.com/huggingface/transformers}}, ensuring compatibility with open-source LLMs. It also integrates the Ollama~\footnote{\url{https://github.com/ollama}} framework, supporting the localized use of LLMs.
% Anticipating the need for users to localize and adapt open-source models within private RAG algorithms, we have developed a system that supports deploying localized models on GPU and CPU. 
In addition to open-source models, the generator module of XRAG includes support for closed-source LLM APIs. Thus, users can access diverse capabilities while retaining the option to use proprietary models.

\subsection{Basic \& Advanced RAG Orchestrators}
\label{Advanced_Orchestrator_Modules}
% 随着RAG技术的发展，先进的RAG框架不再拘泥于pipeline的线性结构，而是以更复杂的编排方式编排基础模块以提升性能和适应多样化的应用场景。
% 我们使用编排器模块来组织和管理RAG组件的执行逻辑和运行流程。
With the advancement of RAG technology, modern agentic RAG frameworks are no longer confined to a linear pipeline structure. 
Instead, they employ more sophisticated orchestration methods to arrange fundamental RAG components, enhancing performance and adapting to diverse application scenarios. 
We use orchestrator modules to organize and manage the execution logic and workflow of RAG components.
As illustrated in Fig.~\ref{orchestrator}, the XRAG framework includes five types of orchestrators: sequential, conditional, iterative, parallel, and hybrid.

\textbf{\({\clubsuit}\) Sequential} 
% XRAG框架可以使用顺序编排器来实现传统的RAG方法。不论是在什么应用场景下，顺序编排器都会按照预检索-检索-后处理-生成的线性流程来组织RAG组件的执行逻辑和运行流程。
The sequential orchestrator operates through the sequential orchestration of advanced components to perform retrieval, and is suited for simple application scenarios.
% The XRAG framework can utilize a sequential orchestrator to implement the traditional RAG process. 
% The sequential orchestrator is suitable for simple application scenarios. 
Extending beyond Naive RAG, XRAG's sequential orchestrator leverages advanced modules to optimize retrieval performance while managing computational costs.
With this orchestrator, the XRAG framework's workflow strictly follows a linear sequence of steps—``pre-retrieval→retrieval→post-processing→generation''—executed step-by-step to retrieve contents and generate answers.

\textbf{\({\spadesuit}\) Conditional} 
% 在XRAG框架中，我们使用条件编排器去实现需要条件分支的RAG方法。条件编排器允许根据特定条件或规则动态调整RAG组件的执行顺序和逻辑。
% 我们集成了Self-RAG~\cite{abs-2307-03155}作为条件编排器的典型应用。另外，Open-RAG~\cite{abs-2307-03155}也可以作为条件编排器的一个实例。
In the XRAG framework, we employ a conditional orchestrator to implement RAG methods that require conditional branching. 
This orchestrator selects the retrieval method according to the type of query, 
and dynamically adjusts the execution sequence and logic of RAG components based on specific conditions or rules.
We have integrated Self-RAG~\cite{asai2024self} as a typical application of the conditional orchestrator, which employs self-reflection to determine whether retrieval is necessary. 
Additionally, Open-RAG~\cite{islam2024open}, which performs targeted retrieval by first classifying queries, can also serve as an example of this conditional orchestration approach.

\textbf{\({♥}\)\:\ Iterative} 
% XRAG框架支持使用迭代编排器来实现需要多轮迭代的RAG方法。迭代编排器允许RAG组件在多个轮次中反复执行，直到逐步完善检索结果并生成内容。
% SIM-RAG作为一个经典的迭代编排器被集成到XRAG框架中。
% 另外，我们集成了RFM-RAG作为另一个迭代编排器的实例。
The XRAG framework supports the use of an iterative orchestrator to implement RAG methods requiring multiple iterations. 
This orchestrator enables RAG components to execute repeatedly across rounds, progressively refining retrieval results and generated content until the criteria are met.
As an advanced iterative orchestrator, SIM-RAG~\cite{yang2025knowing}, which is capable of performing multiple retrieval steps and refining prompts during the iterative process, has been integrated into the XRAG framework. 
We have incorporated RFM-RAG~\cite{li2025retrieval} as another implementation of the iterative orchestrator. 
RFM-RAG dynamically constructs an evidence pool through iterative retrieval and formulates targeted queries until termination.

\textbf{\({\blacktriangledown}\) Parallel} 
% 为了实现需要并行逻辑的RAG方法，XRAG框架支持使用并行编排器。并行编排器允许多个RAG组件在逻辑上的并行运行。
% 作为一个经典的并行编排的RAG技术，Reciprocal Rerank Fusion Retriever（RRFusion）被集成在XRAG框架中。
To enable RAG methods requiring parallel logic, the XRAG framework supports the use of a parallel orchestrator. 
This orchestrator allows multiple RAG components to operate in logical parallelism, as opposed to the conditional orchestrator that relies on selection logic.
As a classic parallel-orchestrated RAG technique, the retrieval logic of Reciprocal Rerank Fusion Retriever (\texttt{RRFusion}~\cite{CormackCB09}) has been integrated into the XRAG framework as an orchestrator.
% Reciprocal Rerank Fusion Retriever fuses indexes with a BM25-based retriever, capturing both semantic relations and keyword relevance. 
Both retrievers in RRFusion assign scores in parallel, enabling reciprocal reranking for node sorting without additional models or excessive computation.
% SeaKR方法并行多组检索选择最优检索生成结果，同样被集成在了XRAG的框架中。
The SeaKR~\cite{yao2024seakr} method, which performs parallel retrievals with multiple groups and selects the optimal retrieval for generation, has also been integrated into the XRAG framework.

\textbf{\({\blacktriangle}\) Hybrid} 
% 面对一些较为复杂的检索场景，单一编排器无法很好的解决问题，XRAG框架支持使用混合编排器来实现复杂的RAG方法。混合编排器结合了多种编排逻辑，以适应复杂的RAG需求。
% XRAG框架集成了Adaptive-RAG方法作为混合编排器的一个实例。Adaptive-RAG包含条件编排和迭代编排的逻辑，以实现更灵活和高效的检索流程。
The XRAG framework supports the use of a hybrid orchestrator to address complex RAG scenarios where a single orchestrator proves insufficient. 
The hybrid orchestrator can combine multiple orchestration logics to accommodate sophisticated RAG requirements.
The framework integrates the Adapt-RAG~\cite{jeong2024adaptive} method as an instance of this hybrid orchestrator. 
Adapt-RAG incorporates both conditional and iterative orchestration logic, dynamically switching between direct and iterative retrieval depending on the complexity of the query, to achieve a flexible and efficient retrieval process.

\addtocounter{changecounter}{-1}
\begin{table*}[t]
  \centering
    % \footnotesize
    \scriptsize
    \caption{
      \changemark[Summary of Benchmark Dataset and Corpus]{
        Summary of Benchmark Datasets \& Corpus.} Original  training and validation sets, are retained for potential future fine-tuning of RAG systems or other customized tasks, although XRAG does not currently support fine-tuning.
      }
  \renewcommand\arraystretch{1.4}
  \setlength{\tabcolsep}{3.15mm}{
    \begin{tabular}{|l|lll|ll|c|c|c|c|}
      \specialrule{1pt}{0pt}{0pt}
    \multirow{2}{*}{\textbf{Datasets}} &
      \multicolumn{3}{c|}{\textbf{Size}} &
      \multicolumn{2}{c|}{\textbf{Corpus}} &
      \multirow{2}{*}{\textbf{Multi-hop}} &
      \multirow{2}{*}{\textbf{Constrained}} &
      \multirow{2}{*}{\textbf{Numerical}} &
      \multirow{2}{*}{\textbf{Set-logical}} \\ \cline{2-6}
              & \multicolumn{1}{l|}{\textbf{Train}}   & \multicolumn{1}{l|}{\textbf{Validation}} & \textbf{Test}  & \multicolumn{1}{l|}{\textbf{Documents}} & \textbf{Source}    &  &  &  &  \\ \hline
              \textbf{HotpotQA (HQA)}  & \multicolumn{1}{l|}{86,830}  & \multicolumn{1}{l|}{8,680}      & 968   & \multicolumn{1}{l|}{508,826}   & Wikipedia & \ding{52} & \usym{2717}  & \ding{52} & \usym{2717} \\ \hline
              \textbf{DropQA (DQA)}    & \multicolumn{1}{l|}{78,241}  & \multicolumn{1}{l|}{7,824}      & 870   & \multicolumn{1}{l|}{6,147}     & Wikipedia & \ding{52} & \usym{2717} &\ding{52}  & \ding{52} \\ \hline
              \textbf{NaturalQA (NQA)} & \multicolumn{1}{l|}{100,093} & \multicolumn{1}{l|}{10,010}     & 1,112 & \multicolumn{1}{l|}{49,815}    & Wikipedia &\ding{52}  &\ding{52}  & \ding{52} & \usym{2717} \\  \hline
              % \rowcolor{blue!30}
              \textbf{\changetable{FinanceBench (FQA)}} & \multicolumn{1}{l|}{43,658} & \multicolumn{1}{l|}{50}     & 100 & \multicolumn{1}{l|}{37,885}    & SEC Filings &\ding{52}  & \usym{2717}  & \ding{52} & \ding{52} \\  \specialrule{1pt}{0pt}{0pt}
    \end{tabular}
  }
  \label{Summary_of_Datasets}
\end{table*}

\subsection{Unified Benchmark Datasets \& Corpus }
\label{Unified_Datasets}
We collect and preprocess four benchmark datasets for the XRAG framework, emphasising rigorous experimental validation of RAG systems. We develop a unified dataset structured to facilitate performance testing for both retrieval and generation modules, incorporating standardized formats: 

\begin{mdframed}
\footnotesize{\texttt{\textbf{User-Query} || \textbf{Retrieval-Context} || \textbf{Retrieval-Context.IDS} || \textbf{Golden-Context} || \textbf{Golden-Context.IDS} || \textbf{Actual-Response} || \textbf{Expected-Answer}}}.
\end{mdframed}

% \begin{figure*}[t]
%     \centering    
%     % First row: Retrieval metrics
%     \begin{subfigure}[t]{0.3\linewidth}
%         \centering
%         \includegraphics[width=\linewidth]{pics/Figure_hotpot.pdf}
%         \caption{\scriptsize HotpotQA}
%     \end{subfigure}
%     \hfill
%     \begin{subfigure}[t]{0.3\linewidth}
%         \centering
%         \includegraphics[width=\linewidth]{pics/Figure_drop.pdf}
%         \caption{\scriptsize DropQA}
%     \end{subfigure}
%     \hfill
%     \begin{subfigure}[t]{0.3\linewidth}
%         \centering
%         \includegraphics[width=\linewidth]{pics/Figure_nq.pdf}
%         \caption{\scriptsize NaturalQA}
%     \end{subfigure}
%     \caption{The golden contextual distribution of the corpora across three datasets providing the quantity and length of annotated contexts. This distribution aids in analyzing the contextual structure, enabling a clearer understanding of how contexts vary in detail for each dataset.}
%     \label{dataset_vue}
% \end{figure*}

XRAG provides a unified architecture for retrieval and question-answering datasets
\footnote{The dataset is CC BY-NC-SA 4.0 licensed (we maintained the same license agreement as the original dataset.), accessible via the URL \url{https://huggingface.co/datasets/DocAILab/XRAG_Dataset}. We also provide a documentation for the XRAG: \url{https://docailab.github.io/XRAG/} on how to use the data and code.
}, 
specifically HoppotQA~\cite{yang2018hotpotqa}, DropQA~\cite{Dua2019DROP}, NaturalQA~\cite{KwiatkowskiPRCP19} and FinanceBench~\cite{mao2025privacy}. The corpus used for indexing is derived from the metadata of these datasets' training, validation, and test sets. This methodology aligns with previous works~\cite{ShiMYS0LZY24} and supports the efficient deployment of vector databases. 
Table~\ref{Summary_of_Datasets} shows that the HotpotQA corpus contains the highest document count, followed by NaturalQA, indicating that retrieval difficulty escalates with the number of documents required per query.
\changemark[FedE4FIN Dataset]{
  The FinanceBench~\cite{islam2023financebenchnewbenchmarkfinancial, mao2025privacy} dataset consists of 37,885 financial documents covering 40 companies that are publicly traded in the USA.
}
  Moreover, these datasets address complex RAG question-answering scenarios, including Multi-hop Queries, Constrained Queries, Numerical Reasoning, and Logical Reasoning tasks.  
Multi-hop questions depend on iteratively retrieved documents and their interrelations to deduce the answer. Constrained Q\&A generates each answer alongside a corresponding constraint or condition rather than providing a standalone response. Numerical reasoning involves performing arithmetic operations such as addition, subtraction, sorting, and counting. Set-logical reasoning addresses complex logical problems involving relationships among retrieved chunks.

% The corpora from HQA, DQA and NQA originate from Wikipedia. 
We constructed retrievable documents using metadata from the original datasets, standardizing retrieval objects as document IDs for testing. Retrieval is successful if the retrieved chunk node corresponds to the annotated document ID. This method ensures consistency in retrieval labels and eliminates discrepancies caused by varying document chunking strategies. 
% 删图对应的文字
% Moreover, Fig.~\ref{dataset_vue} illustrates the distribution of context lengths across three Q\&A datasets: HotpotQA, DropQA, and NaturalQA. 
% HotpotQA shows a tight clustering of context lengths between 0.1 to 0.3k tokens, focusing on shorter contexts. 
% DropQA presents a broader range, with most contexts falling within the 0.1 to 0.5k tokens range but with a more extended tail. 
% NaturalQA exhibits the most diverse distribution, with context lengths spanning from very short to as long as 140k tokens, reflecting a design that accommodates various text lengths for Q\&A tasks. 
For the test set, we limit the number of samples to mitigate the high token cost associated with RAG and LLM evaluations. 
In HQA, NQA, and DQA, instead of evaluating the entire test set, we apply a sampling-based averaging method, reducing token usage while preserving reliability.
For FQA, which has a smaller test set, we use all test data for evaluation.

The respective prompts of XRAG for the HQA, DQA, NQA and FQA datasets are listed below.

\begin{tcolorbox}[
  colback=gray!10,
  colframe=gray!166,
  width=\linewidth,
  arc=2mm, auto outer arc,
  title={\scriptsize Prompt for XRAG suitable for HotpotQA and FinanceBench Datasets},
  breakable,
  before upper={\setlength{\parskip}{0pt}}
  ]
      \scriptsize
      {Context information is below.
        
      \(\cdots\) \(\cdots\) \(\cdots\) \{context\_str\}  \(\cdots\) \(\cdots\) \(\cdots\)

        Given the context information and no prior knowledge, answer the question: 
        
        \{query\_str\}

        We have the opportunity to refine the original answer.
        
        (only if needed) with some more context below.

        \(\cdots\) \(\cdots\) \(\cdots\) \{context\_msg\} \(\cdots\) \(\cdots\) \(\cdots\)

        Given the new context, refine the original answer to better.
        
        Answer the question: {query\_str}

        If the context isn't proper, output the original answer again.
        
        Original Answer: {existing\_answer}
      }
\end{tcolorbox}

\begin{tcolorbox}[
  colback=gray!10,
  colframe=gray!166,
  width=\linewidth,
  arc=2mm, auto outer arc,
  title={\scriptsize Prompt for XRAG suitable for DropQA and NaturalQA Datasets},
  breakable
  ]
      \scriptsize
      {
        Context information is below.
        
       \(\cdots\) \(\cdots\) \(\cdots\) \{context\_str\}  \(\cdots\) \(\cdots\) \(\cdots\)

        Given the context information and no prior knowledge. Please provide a brief, shortest possible answer, ideally just one word for the following question: \

        Question:  who has sold more records Oasis or Coldplay?

        Expected Answer: Oasis

        We have the opportunity to refine the original answer.
        
        (only if needed) with some more context below.

        \(\cdots\) \(\cdots\) \(\cdots\) \{context\_msg\} \(\cdots\) \(\cdots\) \(\cdots\)

        Given the new context, refine the original answer to better.
        
        Answer the question: {query\_str}

        If the context isn't proper, output the original answer again.
        
        Original Answer: {existing\_answer}

      }
\end{tcolorbox}

\subsection{Evaluation Methods}
\label{Evaluation_Methods}
To assess the quality of the RAG components, we integrate the Jury~\cite{cavusoglu2023jury}, a comprehensive package for the evaluation of NLG systems, with RAG community evaluation tools such as UpTrain\footnote{\url{https://github.com/uptrain-ai/uptrain}} and DeepEval\footnote{\url{https://github.com/confident-ai/deepeval}}. 
XRAG evaluators are pivotal in determining the effectiveness of both the retrieval and generation components. They are categorized into three groups: Conventional Retrieval Evaluation, Conventional Generation Evaluation, and Cognitive LLM Evaluation. 

\textbf{Conventional Retrieval Evaluation (ConR Evaluator)}. 
It supports six primary metrics: F1, Mean Reciprocal Rank (MRR), and Mean Average Precision (MAP), along with Hit@1 and Hit@10. Additionally, it includes the DCG family of metrics, which assesses the effectiveness of ranking models by evaluating the quality of ordered results (\texttt{{\footnotesize retrieval-context.ids}}). This family comprises Discounted Cumulative Gain (DCG), Normalized Discounted Cumulative Gain (NDCG), and Ideal Discounted Cumulative Gain (IDCG).  IDCG is the maximum DCG that can be obtained if the results are ideally ranked – arranged in descending order of their relevance (\texttt{{\footnotesize golden-context.ids}}). 

\textbf{Conventional Generation Evaluation (ConG Evaluator)}.  
These generative-token matching metrics can be classified into three broad categories. N-gram similarity metrics: ChrF~\cite{Popovic15}, ChrF++~\cite{Popovic17}, METEOR~\cite{BanerjeeL05}, ROUGE F1~\cite{rouge2004package} (R1, R2, RL) focus on overlap in n-grams between generation (\texttt{{\footnotesize actual-response}}) and reference (\texttt{{\footnotesize expected-answer}}).
Divergence-based metrics: MAUVE~\cite{PillutlaSZTWCH21}, Perplexity~\cite{jelinek1977perplexity} measure content quality, diversity, and model learning by comparing the distribution between the generation and reference.
Error-based accuracy metrics: Word Error Rate (WER)~\cite{MorrisMG04}, Character Error Rate (CER)~\cite{MorrisMG04}, assess the accuracy of \texttt{{\footnotesize actual-response}} by calculating the differences or errors when compared with the \texttt{{\footnotesize expected-answer}}.

\begin{figure}[t]
\centering
\begin{tcolorbox}[
  colback=white,
  colframe=black,
  arc=0mm,
  boxrule=1pt,
  left=6pt,
  right=6pt,
  top=6pt,
  title={\scriptsize Interpretation of Cognitive LLM Evaluation Metrics},
  bottom=6pt,
  width=0.95\linewidth,
  fonttitle=\bfseries
]
\scriptsize
\begin{itemize}[ ]
  \item \textbf{Up-CRel} [\textcolor{c31}{\textbf{Retrieval}}]: \textcolor{c31}{\textbf{Uptrain-Context-Relevance}}
  \begin{itemize}[ ]
    \item \(\rm Pa.\) <\texttt{Query}, \texttt{Retrieval-Context}>
  \end{itemize}
  \item \textbf{Up-CCns} [\textcolor{c31}{\textbf{Retrieval}}]: \textcolor{c31}{\textbf{Uptrain-Context-Conciseness}}
  \begin{itemize}[ ]
    \item \(\rm Pa.\) <\texttt{Query}, \texttt{Golden-Context}, \texttt{Retrieval-Context}> 
  \end{itemize}
  \item \textbf{Dp-ARel} [\textcolor{c32}{\textbf{Generation}}]: \textcolor{c32}{\textbf{DeepEval-Response-Relevancy}}
  \begin{itemize}[ ]
    \item \(\rm Pa.\) <\texttt{Query}, \texttt{Actual-Response}>
  \end{itemize}
  \item \textbf{Up-RCmp} [\textcolor{c32}{\textbf{Generation}}]: \textcolor{c32}{\textbf{Uptrain-Response-Completeness}}
  \begin{itemize}[ ]
    \item \(\rm Pa.\) <\texttt{Query}, \texttt{Actual-Response}> 
  \end{itemize}
  \item \textbf{Up-RCnc} [\textcolor{c32}{\textbf{Generation}}]: \textcolor{c32}{\textbf{Uptrain-Response-Conciseness}}
  \begin{itemize}[ ]
    \item \(\rm Pa.\) <\texttt{Query}, \texttt{Actual-Response}> 
  \end{itemize}
  \item \textbf{Up-RRel} [\textcolor{c32}{\textbf{Generation}}]: \textcolor{c32}{\textbf{Uptrain-Response-Relevance}}
  \begin{itemize}[ ]
    \item \(\rm Pa.\) <\texttt{Query}, \texttt{Actual-Response}> 
  \end{itemize}
  \item \textbf{Up-RVal} [\textcolor{c32}{\textbf{Generation}}]: \textcolor{c32}{\textbf{Uptrain-Response-Valid}}
  \begin{itemize}[ ]
    \item \(\rm Pa.\) <\texttt{Query}, \texttt{Actual-Response}> 
  \end{itemize}
  \item \textbf{Up-RMch} [\textcolor{c32}{\textbf{Generation}}]: \textcolor{c32}{\textbf{Uptrain-Response-Matching}}
  \begin{itemize}[ ]
    \item \(\rm Pa.\) <\texttt{Query}, \texttt{Actual-Response}, \texttt{Expected-Answer}> 
  \end{itemize}
  \item \textbf{Dp-CPre} [\textcolor{c33}{\textbf{Retrieval \& Response}}]: \textcolor{c33}{\textbf{DeepEval-Context-Precision}} 
  \begin{itemize}[ ]
    \item \(\rm Pa.\) <\texttt{Query}, \texttt{Actual-Response}, \texttt{Expected-Answer}, \texttt{Retrieval-Context}> 
  \end{itemize}
  \item \textbf{Dp-CRec} [\textcolor{c33}{\textbf{Retrieval \& Response}}]: \textcolor{c33}{\textbf{DeepEval-Context-Recall}}
  \begin{itemize}[ ]
    \item \(\rm Pa.\) <\texttt{Query}, \texttt{Actual-Response}, \texttt{Expected-Answer}, \texttt{Retrieval-Context}> 
  \end{itemize}
  \item \textbf{Dp-CRel} [\textcolor{c33}{\textbf{Retrieval \& Response}}]: \textcolor{c33}{\textbf{DeepEval-Context-Relevance}} 
  \begin{itemize}[ ]
    \item \(\rm Pa.\) <\texttt{Query}, \texttt{Actual-Response},  \texttt{Retrieval-Context}> 
  \end{itemize}
  \item \textbf{Up-RCns} [\textcolor{c33}{\textbf{Retrieval \& Response}}]: \textcolor{c33}{\textbf{Uptrain-Context-Consistency}} 
  \begin{itemize}[ ]
    \item \(\rm Pa.\) <\texttt{Query}, \texttt{Actual-Response},  \texttt{Retrieval-Context}> 
  \end{itemize}
  \item \textbf{Up-CUti} [\textcolor{c33}{\textbf{Retrieval \& Response}}]: \textcolor{c33}{\textbf{Uptrain-Context-Utilization}}
  \begin{itemize}[ ]
    \item \(\rm Pa.\) <\texttt{Query}, \texttt{Actual-Response},  \texttt{Retrieval-Context}> 
  \end{itemize}
  \item \textbf{Up-FAcc} [\textcolor{c33}{\textbf{Retrieval \& Response}}]: \textcolor{c33}{\textbf{Uptrain-Factual-Accuracy}}
  \begin{itemize}[ ]
    \item \(\rm Pa.\) <\texttt{Query}, \texttt{Actual-Response},  \texttt{Retrieval-Context}> 
  \end{itemize}
  \item \textbf{Dp-Fath} [\textcolor{c33}{\textbf{Retrieval \& Response}}]: \textcolor{c33}{\textbf{DeepEval-Context-Faithfulness}}
  \begin{itemize}[ ]
    \item \(\rm Pa.\) <\texttt{Query}, \texttt{Actual-Response},  \texttt{Retrieval-Context}> 
  \end{itemize}
  \item \textbf{Dp-Hall} [\textcolor{c33}{\textbf{Retrieval \& Response}}]: \textcolor{c33}{\textbf{DeepEval-Context-Hallucination}}
  \begin{itemize}[ ]
    \item \(\rm Pa.\) <\texttt{Query}, \texttt{Actual-Response},  \texttt{Golden-Context}> 
  \end{itemize}
\end{itemize}
\end{tcolorbox}
\caption{Interpretation of Cognitive LLM Evaluation Metrics}
\label{congl_detail_2}
\end{figure}

\addtocounter{changecounter}{-1}
\begin{table*}[t]
% \footnotesize
\scriptsize
\caption{
\changemark[Hyper-Parameters_Orchestrator]{
Fixed and distinctive modules in Agentic RAG Orchestrators.}
SelfRAG-Llama2-7B and SIM-RAG-Llama3-2B are fine-tuned models used in the SelfRAG and SIM-RAG;
T5-Large is a 770M-parameter model released by Google; 
and RRFusion employs the Reciprocal Rerank Fusion Retriever from the LlamaIndex toolkit.
}
\label{orchestrator_hyperparams}
\setlength{\tabcolsep}{2.4mm}
\centering
\setlength{\arrayrulewidth}{0.6pt}{
\renewcommand{\arraystretch}{1.2}
\begin{tabular}{@{}>{\centering\arraybackslash}p{1.6cm}|>{\centering\arraybackslash}p{2cm}>{\centering\arraybackslash}p{3.5cm}|>{\centering\arraybackslash}p{2.2cm}>{\centering\arraybackslash}p{2.3cm}>{\centering\arraybackslash}p{1.8cm}>{\centering\arraybackslash}p{1.8cm}@{}}
\hline
% \textbf{Method} & \textbf{Retrieval Model} & \textbf{Generative Model} & \textbf{Classifier}& \textbf{Critic}& \textbf{Fusion} \\
\multirow{2}{*}{\textbf{Method}} & \multicolumn{2}{c|}{\textbf{Fixed modules}} & \multicolumn{4}{c}{\textbf{Distinctive modules}} \\
\cline{2-3} \cline{4-7}
& \textbf{Retrieval Model} & \textbf{Generative Model} & \textbf{Classifier}& \textbf{Critic}& \textbf{Fusion} & \textbf{Adapter}\\
\hline
% ORACLE  & N/A & deepseek-r1-distill-llama-70b & N/A & N/A & N/A& N/A \\
BGE-Large  & BGE-Large-en-v1.5 & deepseek-r1-distill-llama-70b & \usym{2717} & \usym{2717} & \usym{2717} & \usym{2717}  \\
Self-RAG  & BGE-Large-en-v1.5 & deepseek-r1-distill-llama-70b &  SelfRAG-Llama2-7B   & \usym{2717} & \usym{2717} & \usym{2717}\\
SIM-RAG & BGE-Large-en-v1.5 & deepseek-r1-distill-llama-70b & \usym{2717} & SIM-RAG-Llama3-2B   & \usym{2717} & \usym{2717} \\
RRFusion & BGE-Large-en-v1.5 & deepseek-r1-distill-llama-70b  & \usym{2717} & \usym{2717} &  RRFusion~\cite{CormackCB09} & \usym{2717}\\
Adapt-RAG  & BGE-Large-en-v1.5 & deepseek-r1-distill-llama-70b  & \usym{2717} & \usym{2717} & \usym{2717} & T5-Large~\cite{JMLR:v21:20-074} \\
\hline
\end{tabular}
}
\end{table*}

\addtocounter{changecounter}{-1}
\begin{table*}[tb]
\caption{
    \changemark[Basic Retriever Table]{Upstream RAG \textbf{Retrieval Performance}} using \textbf{Basic Retriever} of BGE-Large and JINA-Large. The conventional retrieval evaluation, denoted as ConR, is performed. The symbol \(^{\circ}\) signifies metric values ranging from 0 to positive infinity, while all other values are within the \(\left [0,1 \right ] \) interval (\%).}
  \centering
  % \footnotesize
  \scriptsize
\renewcommand\arraystretch{1.25}
\setlength{\tabcolsep}{5.5mm}{
  \begin{tabular}{|l|l|cccccccc|}
    \specialrule{1pt}{0pt}{0pt}
  \multicolumn{2}{|l|}{\multirow{2}{*}{\textbf{Dataset \& Methods}}} & \multicolumn{8}{c|}{\textbf{Conventional Retrieval Evaluation (ConR Evaluator)}}    \\ \cline{3-10}
  \multicolumn{2}{|l|}{}                                   & \multicolumn{1}{c|}{\textbf{F1}} & \multicolumn{1}{c|}{\textbf{MRR}} & \multicolumn{1}{c|}{\textbf{Hit@1}} & \multicolumn{1}{c|}{\textbf{Hit@10}} & \multicolumn{1}{c|}{\textbf{MAP}} & \multicolumn{1}{c|}{\textbf{NDCG}} & \multicolumn{1}{c|}{\textbf{DCG}\(^{\circ}\)} & \multicolumn{1}{c|}{\textbf{IDCG}\(^{\circ}\)}                      \\ \hline
  \multirow{2}{*}{\rotatebox{90}{\texttt{\textbf{HQA}}}}           
  & \textbf{JINA-Large}         & \multicolumn{1}{c|}{49.80}   & \multicolumn{1}{c|}{64.23}    & \multicolumn{1}{c|}{21.42}     & \multicolumn{1}{c|}{94.44}      & \multicolumn{1}{c|}{55.17}    & \multicolumn{1}{c|}{70.44}     & \multicolumn{1}{c|}{0.9823}    & \multicolumn{1}{c|}{1.1795}                         \\ 
  & \textbf{BGE-Large}        & \multicolumn{1}{c|}{60.66}   & \multicolumn{1}{c|}{72.84}    & \multicolumn{1}{c|}{18.29}     & \multicolumn{1}{c|}{97.71}      & \multicolumn{1}{c|}{64.13}    & \multicolumn{1}{c|}{78.64}     & \multicolumn{1}{c|}{1.1644}    & \multicolumn{1}{c|}{1.3656}                         \\ 
\hline
\multirow{2}{*}{\rotatebox{90}{\texttt{\textbf{DQA}}}}            
& \textbf{JINA-Large}          & \multicolumn{1}{c|}{20.88}   & \multicolumn{1}{c|}{24.42}    & \multicolumn{1}{c|}{06.90}     & \multicolumn{1}{c|}{34.47}      & \multicolumn{1}{c|}{24.42}    & \multicolumn{1}{c|}{26.76}     & \multicolumn{1}{c|}{0.3835}    & \multicolumn{1}{c|}{0.4482}                         \\ 
& \textbf{BGE-Large}        & \multicolumn{1}{c|}{22.58}   & \multicolumn{1}{c|}{27.10}    & \multicolumn{1}{c|}{07.12}     & \multicolumn{1}{c|}{42.30}      & \multicolumn{1}{c|}{27.10}    & \multicolumn{1}{c|}{29.58}     & \multicolumn{1}{c|}{0.4160}    & \multicolumn{1}{c|}{0.4849}                        \\ 
\hline
\multirow{2}{*}{\rotatebox{90}{\texttt{\textbf{NQA}}}}         
& \textbf{JINA-Large}           & \multicolumn{1}{c|}{49.20}   & \multicolumn{1}{c|}{62.44}    & \multicolumn{1}{c|}{30.31}     & \multicolumn{1}{c|}{88.49}      & \multicolumn{1}{c|}{62.44}    & \multicolumn{1}{c|}{66.98}     & \multicolumn{1}{c|}{0.8843}    & \multicolumn{1}{c|}{1.0120}                         \\ 
& \textbf{BGE-Large}        & \multicolumn{1}{c|}{49.90}   & \multicolumn{1}{c|}{63.82}    & \multicolumn{1}{c|}{25.68}     & \multicolumn{1}{c|}{93.24}      & \multicolumn{1}{c|}{63.82}    & \multicolumn{1}{c|}{69.12}     & \multicolumn{1}{c|}{0.8799}    & \multicolumn{1}{c|}{1.0300}    \\ 
\hline
 \multirow{2}{*}{\rotatebox{90}{\texttt{\changetable{\textbf{FQA}}}}}         
& \textbf{JINA-Large}           & \multicolumn{1}{c|}{9.65}   & \multicolumn{1}{c|}{40.75}    & \multicolumn{1}{c|}{38.00}     & \multicolumn{1}{c|}{43.00}      & \multicolumn{1}{c|}{40.75}    & \multicolumn{1}{c|}{41.39}     & \multicolumn{1}{c|}{0.5076}    & \multicolumn{1}{c|}{0.5268}                         \\ 
 & \textbf{BGE-Large}        & \multicolumn{1}{c|}{6.05}   & \multicolumn{1}{c|}{26.83}    & \multicolumn{1}{c|}{23.00}     & \multicolumn{1}{c|}{29.00}      & \multicolumn{1}{c|}{26.83}    & \multicolumn{1}{c|}{27.45}     & \multicolumn{1}{c|}{0.3104}    & \multicolumn{1}{c|}{0.3278}    \\ 
  
\specialrule{1pt}{0pt}{0pt}
  % \hline
  \end{tabular}
}
\label{main_conr_retrieval}
\end{table*}

\textbf{Cognitive LLM Evaluation (CogL Evaluator)}. 
It is classified into three categories: Retrieval-oriented, Generation-oriented, and Combined Retrieval \& Generation. 
Cognitive LLM Evaluation metrics, derived from UpTrain(prefixed with `Up') and DeepEval(prefixed with `Dp'), are classified based on the parameters unified by our XRAG framework. 
Response-oriented metrics involve response-related parameters including Response Relevance (Dp-ARel), Response Completeness (Up-RCmp) from DeepEval, Response Conciseness (Up-RCnc), Response Relevance (Up-RRel), and Response Validity (Up-RVal) and Response Matching (Up-RMch) from Uptrain. 
Retrieval-oriented metrics that lack response-related parameters and include retrieval-related parameters, assess context quality and consist of Context Relevance (Up-CRel) and Context Conciseness (Up-CCns), which come from UpTrain.  
Combined metrics evaluate the impact of retrieval on final responses and include Context Precision (Dp-CPre), Context Recall (Dp-CRec), Context Relevance (Dp-CRel), Response Consistency (Up-RCns),  Context Utilization (Up-CUti), and Factual Accuracy (Up-FAcc), Faithfulness (Dp-Faith), and Hallucination (Dp-Hall). 
% Metrics involving response-related parameters, such as \texttt{{\footnotesize actual-response}} or \texttt{{\footnotesize expected-answer}}, are Generation-oriented Metrics. 
% Conversely, metrics that lack response-related parameters, including retrieval-related parameters, such as \texttt{{\footnotesize retrieval-context}} or \texttt{{\footnotesize retrieval-context.ids}}, are designated as Retrieval-oriented Metrics. 
% The rest are Combined Retrieval \& Generation Metrics. 
% Moreover, we utilize \textbf{GPT-3.5 Turbo} as the LLM agent for CogL evaluator. 
% Retrieval-oriented metrics assess context quality and consist of Context Relevance (Up-CRel) and Context Conciseness (Up-CCns), which come from UpTrain.  
% Response-oriented metrics include Response Relevance (Dp-ARel), Response Completeness (Up-RCmp) from DeepEval, Response Conciseness (Up-RCnc), Response Relevance (Up-RRel), and Response Validity (Up-RVal) and Response Matching (Up-RMch) from Uptrain. 
% Combined metrics evaluate the impact of retrieval on final responses and include Context Precision (Dp-CPre), Context Recall (Dp-CRec), Context Relevance (Dp-CRel), Response Consistency (Up-RCns),  Context Utilization (Up-CUti), and Factual Accuracy (Up-FAcc), Faithfulness (Dp-Faith), and Hallucination (Dp-Hall). 
% The metrics prefixed with `Up' originate from UpTrain, while those prefixed with `Dp' are from DeepEval. 
Detailed usage patterns of the CogL evaluation are illustrated in Fig.~\ref{congl_detail_2}.
% In Cognitive LLM Evaluation, the costs associated with tokens are considerable, stemming from the input and output tokens that participate in the model's reasoning and testing phases. 
Conventional rule-based metrics like Exact Match (EM) and ROUGE concentrate on n-gram matching, which sometimes falls short in capturing the comprehensive accuracy of language expressions; 
however, they remain fitting for tasks that necessitate distinct and singular responses. 
In contrast, metrics based on Cognitive LLMs evaluate the quality of generative language, which is particularly beneficial when the complexity of evaluative comprehension surpasses the capabilities of rule-based metrics.

% Since LLM evaluations are billed per token, we conducted experimental discussions for our foundational modules of RAG in Appendix~\ref{CONGL}.  

XRAG evaluators have obvious advantages:

\begin{itemize}
    \item \textbf{Evaluating with Multiple RAG Metrics in One Go}: The XRAG evaluator allows users to simultaneously assess various RAG-specific metrics. This capability streamlines the evaluation process, enabling comprehensive performance analysis without sequential evaluations.
    \item \textbf{Standardizes the Structure of Evaluation Metrics}: A unified data format simplifies the comparison between different RAG components in both retrieval and generation.
    \item \textbf{Character and Semantic \(\times\) Retrieval and Generation}:  
    It encompasses a 4-fold cross-dimensional analysis, including character-level matching tests and semantic-level understanding tests for both retrieval and generation.
\end{itemize}

\addtocounter{changecounter}{-1}
\begin{table*}[t]
  \caption{\changemark[AR_table]{Upstream RAG \textbf{Retrieval Performance}} about advanced \textbf{Pre-retrieval (PR)}, \textbf{Advanced Retrieval (AR)} and \textbf{Post-processor (PP)}, where the BGE-Large model constitutes the fixed foundational retrieval model. Results \underline{underlined} indicate superiority over Basic Retriever of BGE-Large, while \textbf{bold} results denote the best performance.
  }
  \centering
  % \footnotesize
  \scriptsize
\renewcommand\arraystretch{1.2}
\setlength{\tabcolsep}{3.6mm}{
  \begin{tabular}{|l|l|ccc|@{\hspace{0.5mm}}c@{\hspace{0.5mm}}|cccccccc|}
    \specialrule{1pt}{0pt}{0pt}
  \multicolumn{2}{|l|}{\multirow{2}{*}{\textbf{Dataset \& Methods}}}& \multicolumn{3}{c|}{\textbf{Distinctive}}& \multicolumn{1}{@{\hspace{0.5mm}}c@{\hspace{0.5mm}}|}{\textbf{Fixed}} & \multicolumn{8}{c|}{\textbf{Conventional Retrieval Evaluation (ConR Evaluator)}}    \\ \cline{3-14}
  \multicolumn{2}{|l|}{}              & \multicolumn{1}{c|}{\textbf{PR}} & \multicolumn{1}{c|}{\textbf{AR}} & \multicolumn{1}{c|}{\textbf{PP}} & \multicolumn{1}{@{\hspace{0.5mm}}c@{\hspace{0.5mm}}|}{\textbf{R}}                   & \multicolumn{1}{c|}{\textbf{F1}} & \multicolumn{1}{c|}{\textbf{MRR}} & \multicolumn{1}{c|}{\textbf{Hit@1}} & \multicolumn{1}{c|}{\textbf{Hit@10}} & \multicolumn{1}{c|}{\textbf{MAP}} & \multicolumn{1}{c|}{\textbf{NDCG}} & \multicolumn{1}{c|}{\textbf{DCG}\(^{\circ}\)} & \multicolumn{1}{c|}{\textbf{IDCG}\(^{\circ}\)}                      \\ \hline
  \multirow{6}{*}{\rotatebox{90}{\texttt{\textbf{HQA}}}}          
  & \textbf{SBPT}     & \multicolumn{1}{c}{\ding{52}} & \multicolumn{1}{c}{\usym{2717}} & \multicolumn{1}{c|}{\usym{2717} } & \multirow{6}{*}{\rotatebox{90}{\texttt{\textbf{BGE-Large}}}}    & \multicolumn{1}{c|}{38.60}   & \multicolumn{1}{c|}{53.86}    & \multicolumn{1}{c|}{16.96}     & \multicolumn{1}{c|}{96.43}      & \multicolumn{1}{c|}{44.99}    & \multicolumn{1}{c|}{59.61}     & \multicolumn{1}{c|}{0.7174}    & \multicolumn{1}{c|}{0.8992}                         \\ 
  & \textbf{HyDE}     & \multicolumn{1}{c}{\ding{52}} & \multicolumn{1}{c}{\usym{2717}} & \multicolumn{1}{c|}{\usym{2717} } &   & \multicolumn{1}{c|}{53.07}   & \multicolumn{1}{c|}{67.39}    & \multicolumn{1}{c|}{17.86}    & \multicolumn{1}{c|}{\textbf{97.32}}      & \multicolumn{1}{c|}{58.32}    & \multicolumn{1}{c|}{72.44}     & \multicolumn{1}{c|}{0.9947}    & \multicolumn{1}{c|}{1.1753}                         \\ 
  & \textbf{RRFusion}  & \multicolumn{1}{c}{\usym{2717}} & \multicolumn{1}{c}{\ding{52}} & \multicolumn{1}{c|}{\usym{2717} } &       & \multicolumn{1}{c|}{59.99}   & \multicolumn{1}{c|}{\underline{76.92}}    & \multicolumn{1}{c|}{12.24}     & \multicolumn{1}{c|}{96.94}      & \multicolumn{1}{c|}{\underline{66.25}}    & \multicolumn{1}{c|}{\underline{79.72}}     & \multicolumn{1}{c|}{1.0401}    & \multicolumn{1}{c|}{1.1192}                         \\
  & \textbf{StParser} & \multicolumn{1}{c}{\usym{2717}} & \multicolumn{1}{c}{\ding{52}} & \multicolumn{1}{c|}{\usym{2717} } &      & \multicolumn{1}{c|}{42.92}   & \multicolumn{1}{c|}{59.54}    & \multicolumn{1}{c|}{14.02}     & \multicolumn{1}{c|}{97.20}      & \multicolumn{1}{c|}{49.82}    & \multicolumn{1}{c|}{65.17}     & \multicolumn{1}{c|}{0.8069}    & \multicolumn{1}{c|}{0.9921}                         \\
  & \textbf{BGE-RRK}   & \multicolumn{1}{c}{\usym{2717}} & \multicolumn{1}{c}{\usym{2717}} & \multicolumn{1}{c|}{\ding{52} } &        & \multicolumn{1}{c|}{\textbf{\underline{66.90}}}   & \multicolumn{1}{c|}{\textbf{\underline{80.19}}}    & \multicolumn{1}{c|}{\underline{\textbf{77.03}}}     & \multicolumn{1}{c|}{97.30}      & \multicolumn{1}{c|}{\textbf{\underline{70.62}}}    & \multicolumn{1}{c|}{\textbf{\underline{83.14}}}     & \multicolumn{1}{c|}{\textbf{1.1354}}    & \multicolumn{1}{c|}{\textbf{1.2183}}                         \\
  & \textbf{JINA-RRK}   & \multicolumn{1}{c}{\usym{2717}} & \multicolumn{1}{c}{\usym{2717}} & \multicolumn{1}{c|}{\ding{52} } &       & \multicolumn{1}{c|}{\underline{64.68}}   & \multicolumn{1}{c|}{\underline{76.89}}    & \multicolumn{1}{c|}{\underline{76.92}}     & \multicolumn{1}{c|}{96.43}      & \multicolumn{1}{c|}{\underline{68.83}}    & \multicolumn{1}{c|}{\underline{79.60}}     & \multicolumn{1}{c|}{1.1096}    & \multicolumn{1}{c|}{1.1733}                         \\
  \hline
\multirow{6}{*}{\rotatebox{90}{\texttt{\textbf{DQA}}}}            
& \textbf{SBPT}   & \multicolumn{1}{c}{\ding{52}} & \multicolumn{1}{c}{\usym{2717}} & \multicolumn{1}{c|}{\usym{2717} } & \multirow{6}{*}{\rotatebox{90}{\texttt{\textbf{BGE-Large}}}}      & \multicolumn{1}{c|}{22.57}   & \multicolumn{1}{c|}{27.03}    & \multicolumn{1}{c|}{06.01}     & \multicolumn{1}{c|}{42.08}      & \multicolumn{1}{c|}{27.03}    & \multicolumn{1}{c|}{29.45}     & \multicolumn{1}{c|}{0.4149}    & \multicolumn{1}{c|}{0.4823}                         \\
& \textbf{HyDE} & \multicolumn{1}{c}{\ding{52}} & \multicolumn{1}{c}{\usym{2717}} & \multicolumn{1}{c|}{\usym{2717} } &      & \multicolumn{1}{c|}{\underline{22.74}}   & \multicolumn{1}{c|}{26.40}    & \multicolumn{1}{c|}{06.81}     & \multicolumn{1}{c|}{41.48}      & \multicolumn{1}{c|}{26.40}    & \multicolumn{1}{c|}{28.89}     & \multicolumn{1}{c|}{\underline{0.4213}}    & \multicolumn{1}{c|}{\underline{0.4906}}                       \\ 
&  \textbf{RRFusion} & \multicolumn{1}{c}{\usym{2717}} & \multicolumn{1}{c}{\ding{52}} & \multicolumn{1}{c|}{\usym{2717} } &         & \multicolumn{1}{c|}{\underline{26.36}}   & \multicolumn{1}{c|}{\underline{29.31}}    & \multicolumn{1}{c|}{\underline{14.29}}     & \multicolumn{1}{c|}{19.05}      & \multicolumn{1}{c|}{\underline{29.31}}    & \multicolumn{1}{c|}{\underline{30.18}}     & \multicolumn{1}{c|}{0.3885}    & \multicolumn{1}{c|}{0.4135}                         \\
& \textbf{StParser}   & \multicolumn{1}{c}{\usym{2717}} & \multicolumn{1}{c}{\ding{52}} & \multicolumn{1}{c|}{\usym{2717} } &        & \multicolumn{1}{c|}{\underline{27.70}}   & \multicolumn{1}{c|}{\underline{30.34}}    & \multicolumn{1}{c|}{05.13}     & \multicolumn{1}{c|}{\underline{\textbf{46.15}}}      & \multicolumn{1}{c|}{\underline{30.34}}    & \multicolumn{1}{c|}{\underline{31.28}}     & \multicolumn{1}{c|}{0.4092}    & \multicolumn{1}{c|}{0.4355}                         \\
& \textbf{BGE-RRK}   & \multicolumn{1}{c}{\usym{2717}} & \multicolumn{1}{c}{\usym{2717}} & \multicolumn{1}{c|}{\ding{52} } &      & \multicolumn{1}{c|}{\underline{31.72}}   & \multicolumn{1}{c|}{\underline{34.27}}    & \multicolumn{1}{c|}{\underline{28.46}}     & \multicolumn{1}{c|}{38.88}      & \multicolumn{1}{c|}{\underline{34.27}}    & \multicolumn{1}{c|}{\underline{35.44}}     & \multicolumn{1}{c|}{\underline{0.4663}}    & \multicolumn{1}{c|}{\underline{0.4990}}                        \\
& \textbf{JINA-RRK}    & \multicolumn{1}{c}{\usym{2717}} & \multicolumn{1}{c}{\usym{2717}} & \multicolumn{1}{c|}{\ding{52} } &       & \multicolumn{1}{c|}{\underline{\textbf{34.20}}}   & \multicolumn{1}{c|}{\underline{\textbf{37.66}}}    & \multicolumn{1}{c|}{\underline{\textbf{31.17}}}     & \multicolumn{1}{c|}{\underline{44.16}}      & \multicolumn{1}{c|}{\underline{\textbf{37.66}}}    & \multicolumn{1}{c|}{\underline{\textbf{38.68}}}     & \multicolumn{1}{c|}{\underline{\textbf{0.5097}}}    & \multicolumn{1}{c|}{\underline{\textbf{0.5385}}}                         \\
\hline
\multirow{6}{*}{\rotatebox{90}{\texttt{\textbf{NQA}}}}         
& \textbf{SBPT} & \multicolumn{1}{c}{\ding{52}} & \multicolumn{1}{c}{\usym{2717}} & \multicolumn{1}{c|}{\usym{2717} } &   \multirow{6}{*}{\rotatebox{90}{\texttt{\textbf{BGE-Large}}}}     & \multicolumn{1}{c|}{49.22}   & \multicolumn{1}{c|}{62.27}    & \multicolumn{1}{c|}{22.86}     & \multicolumn{1}{c|}{88.57}      & \multicolumn{1}{c|}{62.27}    & \multicolumn{1}{c|}{67.53}     & \multicolumn{1}{c|}{0.8737}    & \multicolumn{1}{c|}{1.0210}                         \\
& \textbf{HyDE}  & \multicolumn{1}{c}{\ding{52}} & \multicolumn{1}{c}{\usym{2717}} & \multicolumn{1}{c|}{\usym{2717} } &     & \multicolumn{1}{c|}{\underline{50.12}}   & \multicolumn{1}{c|}{62.94}    & \multicolumn{1}{c|}{\underline{41.03}}     & \multicolumn{1}{c|}{\underline{94.87}}     & \multicolumn{1}{c|}{62.94}    & \multicolumn{1}{c|}{68.69}     & \multicolumn{1}{c|}{\underline{0.8864}}    & \multicolumn{1}{c|}{\underline{\textbf{1.0483}}}    \\ 
& \textbf{RRFusion}   & \multicolumn{1}{c}{\usym{2717}} & \multicolumn{1}{c}{\ding{52}} & \multicolumn{1}{c|}{\usym{2717} } &       & \multicolumn{1}{c|}{\underline{\textbf{62.39}}}   & \multicolumn{1}{c|}{\underline{\textbf{70.35}}}    & \multicolumn{1}{c|}{\underline{42.86}}     & \multicolumn{1}{c|}{\underline{\textbf{95.24}}}       & \multicolumn{1}{c|}{\underline{\textbf{70.35}}}    & \multicolumn{1}{c|}{\underline{\textbf{73.02}}}     & \multicolumn{1}{c|}{\underline{\textbf{0.8949}}}    & \multicolumn{1}{c|}{0.9700}                         \\
& \textbf{StParser} & \multicolumn{1}{c}{\usym{2717}} & \multicolumn{1}{c}{\ding{52}} & \multicolumn{1}{c|}{\usym{2717} } &       & \multicolumn{1}{c|}{\underline{59.79}}   & \multicolumn{1}{c|}{\underline{68.42}}    & \multicolumn{1}{c|}{\underline{28.95}}     & \multicolumn{1}{c|}{\underline{94.74}}      & \multicolumn{1}{c|}{\underline{68.42}}    & \multicolumn{1}{c|}{\underline{71.29}}     & \multicolumn{1}{c|}{0.8430}    & \multicolumn{1}{c|}{0.9239}                         \\
& \textbf{BGE-RRK}  & \multicolumn{1}{c}{\usym{2717}} & \multicolumn{1}{c}{\usym{2717}} & \multicolumn{1}{c|}{\ding{52} } &        & \multicolumn{1}{c|}{\underline{59.14}}   & \multicolumn{1}{c|}{\underline{68.26}}    & \multicolumn{1}{c|}{\underline{50.91}}     & \multicolumn{1}{c|}{76.36}      & \multicolumn{1}{c|}{\underline{68.26}}    & \multicolumn{1}{c|}{\underline{71.20}}     & \multicolumn{1}{c|}{0.8284}    & \multicolumn{1}{c|}{0.9113}                         \\
& \textbf{JINA-RRK}   & \multicolumn{1}{c}{\usym{2717}} & \multicolumn{1}{c}{\usym{2717}} & \multicolumn{1}{c|}{\ding{52} } &        & \multicolumn{1}{c|}{\underline{58.97}}   & \multicolumn{1}{c|}{\underline{66.67}}    & \multicolumn{1}{c|}{\underline{\textbf{56.41}}}     & \multicolumn{1}{c|}{79.49}      & \multicolumn{1}{c|}{\underline{66.67}}   & \multicolumn{1}{c|}{68.68}     & \multicolumn{1}{c|}{0.8648}    & \multicolumn{1}{c|}{0.9215}                         \\
\hline
\multirow{6}{*}{\rotatebox{90}{\texttt{\changetable{\textbf{FQA}}}}}         
& \textbf{SBPT}  & \multicolumn{1}{c}{\ding{52}} & \multicolumn{1}{c}{\usym{2717}} & \multicolumn{1}{c|}{\usym{2717} } &   \multirow{6}{*}{\rotatebox{90}{\texttt{\textbf{BGE-Large}}}}    & \multicolumn{1}{c|}{\underline{6.65}}   & \multicolumn{1}{c|}{\underline{27.17}}    & \multicolumn{1}{c|}{21.00}     & \multicolumn{1}{c|}{\underline{30.00}}      & \multicolumn{1}{c|}{\underline{27.16}}    & \multicolumn{1}{c|}{\underline{28.03}}     & \multicolumn{1}{c|}{\underline{0.3367}}    & \multicolumn{1}{c|}{\underline{0.3618}}                         \\
& \textbf{HyDE}  & \multicolumn{1}{c}{\ding{52}} & \multicolumn{1}{c}{\usym{2717}} & \multicolumn{1}{c|}{\usym{2717} } &     & \multicolumn{1}{c|}{4.75}   & \multicolumn{1}{c|}{21.00}    & \multicolumn{1}{c|}{18.00}     & \multicolumn{1}{c|}{23.00}     & \multicolumn{1}{c|}{21.00}    & \multicolumn{1}{c|}{21.58}     & \multicolumn{1}{c|}{0.2441}    & \multicolumn{1}{c|}{0.2602}    \\ 
& \textbf{RRFusion} & \multicolumn{1}{c}{\usym{2717}} & \multicolumn{1}{c}{\ding{52}} & \multicolumn{1}{c|}{\usym{2717} } &        & \multicolumn{1}{c|}{\underline{\textbf{12.66}}}   & \multicolumn{1}{c|}{18.00}    & \multicolumn{1}{c|}{17.00}     & \multicolumn{1}{c|}{19.00}       & \multicolumn{1}{c|}{18.00}    & \multicolumn{1}{c|}{18.26}     & \multicolumn{1}{c|}{0.1889}    & \multicolumn{1}{c|}{0.1963}                         \\
& \textbf{StParser}   & \multicolumn{1}{c}{\usym{2717}} & \multicolumn{1}{c}{\ding{52}} & \multicolumn{1}{c|}{\usym{2717} } &       & \multicolumn{1}{c|}{\underline{7.83}}   & \multicolumn{1}{c|}{12.12}    & \multicolumn{1}{c|}{9.00}     & \multicolumn{1}{c|}{12.00}      & \multicolumn{1}{c|}{12.00}    & \multicolumn{1}{c|}{12.12}     & \multicolumn{1}{c|}{0.1275}    & \multicolumn{1}{c|}{0.1275}                         \\
& \textbf{BGE-RRK}   & \multicolumn{1}{c}{\usym{2717}} & \multicolumn{1}{c}{\usym{2717}} & \multicolumn{1}{c|}{\ding{52} } &       & \multicolumn{1}{c|}{\underline{8.86}}   & \multicolumn{1}{c|}{\underline{35.83}}    & \multicolumn{1}{c|}{\underline{35.00}}     & \multicolumn{1}{c|}{\underline{39.00}}      & \multicolumn{1}{c|}{\underline{35.84}}    & \multicolumn{1}{c|}{\underline{36.71}}     & \multicolumn{1}{c|}{\underline{0.4785}}    & \multicolumn{1}{c|}{\underline{0.5033}}                         \\
& \textbf{JINA-RRK}   & \multicolumn{1}{c}{\usym{2717}} & \multicolumn{1}{c}{\usym{2717}} & \multicolumn{1}{c|}{\ding{52} } &       & \multicolumn{1}{c|}{\underline{10.20}}   & \multicolumn{1}{c|}{\underline{\textbf{42.33}}}    & \multicolumn{1}{c|}{\underline{\textbf{40.00}}}     & \multicolumn{1}{c|}{\underline{\textbf{46.00}}}      & \multicolumn{1}{c|}{\underline{\textbf{41.83}}}   & \multicolumn{1}{c|}{\underline{\textbf{43.41}}}     & \multicolumn{1}{c|}{\underline{\textbf{0.5524}}}    & \multicolumn{1}{c|}{\underline{\textbf{0.5822}}}                         \\
\specialrule{1pt}{0pt}{0pt}
  \end{tabular}
}
\label{advanced_conr_retrieval}
\end{table*}

\addtocounter{changecounter}{-1}
\begin{table*}[t]
  \centering
% \footnotesize
  \scriptsize
  \caption{
  \changemark[Downstream RAG Generation Performance]{  
  Downstream RAG \textbf{Generation Performance}
  } of LLMs, utilizing \textit{Oracle} as the retrieval results. The conventional generation evaluation (ConG) is conducted. The top scores are emphasized in \textbf{bold}. The symbol \(^{\circ}\) signifies metric values ranging from 0 to positive infinity, while all other values are within the \(\left [0,1 \right ] \) interval (\%). The symbol \(\downarrow\) denotes that lower metric values are preferable. The error value is the average of three trials, with a constant LLM Temperature of 0.}
\renewcommand\arraystretch{1.2}
\setlength{\tabcolsep}{2.55mm}{
\begin{tabular}{|ll|lllllllll|}
  \specialrule{1pt}{0pt}{0pt}
\multicolumn{2}{|l|}{\multirow{1}{*}{\textbf{{Datasets \& Methods}}}} &
  \multicolumn{9}{c|}{\textbf{Conventional Generation Evaluation (ConG Evaluator)}} \\
  \cline{3-11}
  \multicolumn{2}{|l|}{\textbf{\textcolor{red}{Oracle: Golden-Context}}} &
  \multicolumn{1}{c|}{\textbf{ChrF}} &
  \multicolumn{1}{c|}{\textbf{ChrF++}} &
  \multicolumn{1}{c|}{\textbf{METEOR}} &
  \multicolumn{1}{c|}{\textbf{R1}} &
  \multicolumn{1}{c|}{\textbf{R2}} &
  \multicolumn{1}{c|}{\textbf{RL}} &
  \multicolumn{1}{c|}{\textbf{PPL}\(^{\circ}_{\downarrow}\)} &
  \multicolumn{1}{c|}{\textbf{CER}\(^{\circ}_{\downarrow}\)} &
  \multicolumn{1}{c|}{\textbf{WER}\(^{\circ}_{\downarrow}\)}  \\
  \hline
  \multicolumn{1}{|l|}{\multirow{4}{*}{\rotatebox{90}{\texttt{\textbf{HQA}}}}} &
  \textbf{GPT-4o mini} &
\multicolumn{1}{c|}{21.75$_{ \pm \scalebox{0.7}{$.024$}}$} &
\multicolumn{1}{c|}{20.01$_{ \pm \scalebox{0.7}{$.012$}}$} &
\multicolumn{1}{c|}{21.62$_{ \pm \scalebox{0.7}{$.008$}}$} &
\multicolumn{1}{c|}{12.93$_{ \pm \scalebox{0.7}{$.007$}}$} &
\multicolumn{1}{c|}{05.32$_{ \pm \scalebox{0.7}{$.003$}}$} &
\multicolumn{1}{c|}{12.71$_{ \pm \scalebox{0.7}{$.005$}}$} &
\multicolumn{1}{c|}{111.98$_{ \pm \scalebox{0.7}{$1.368$}}$} &
\multicolumn{1}{c|}{\textbf{11.08}$_{ \pm \scalebox{0.7}{$.019$}}$} &
\textbf{09.75}$_{ \pm \scalebox{0.7}{$.002$}}$ \\
\end{tabular}
}
\label{main_cong_generation_Oracle}
\end{table*}

\addtocounter{changecounter}{-1}
\begin{table*}[t]
  \centering
%  \footnotesize
  \scriptsize
  \caption{
  \changemark[Downstream RAG Generation Performance 2]{  
  Downstream RAG \textbf{Generation Performance}
   } of LLMs, utilizing \textit{JINA-Large} as the retriever. 
  Results \underline{underlined} indicate superiority over RAG generation based on Oracle context, while \textbf{bold} results denote the best performance.} 
\renewcommand\arraystretch{1.2}
\setlength{\tabcolsep}{2.55mm}{
  \begin{tabular}{|ll|lllllllll|}
    \specialrule{1pt}{0pt}{0pt}
  \multicolumn{2}{|l|}{\multirow{1}{*}{\textbf{{Datasets \& Methods}}}} &
    \multicolumn{9}{c|}{\textbf{Conventional Generation Evaluation (ConG Evaluator)}} 
    \\ \cline{3-11} 
    \multicolumn{2}{|l|}{\textbf{\textcolor{red}{Retriever: JINA-Large}}} &
    \multicolumn{1}{c|}{\textbf{ChrF}} &
    \multicolumn{1}{c|}{\textbf{ChrF++}} &
    \multicolumn{1}{c|}{\textbf{METEOR}} &
    \multicolumn{1}{c|}{\textbf{R1}} &
    \multicolumn{1}{c|}{\textbf{R2}} &
    \multicolumn{1}{c|}{\textbf{RL}} &
    \multicolumn{1}{c|}{\textbf{PPL}\(^{\circ}_{\downarrow}\)} &
    \multicolumn{1}{c|}{\textbf{CER}\(^{\circ}_{\downarrow}\)} &
    \multicolumn{1}{c|}{\textbf{WER}\(^{\circ}_{\downarrow}\)}  \\ \hline
  \multicolumn{1}{|l|}{\multirow{4}{*}{\rotatebox{90}{\texttt{\textbf{HQA}}}}} &
  \textbf{GPT-4o mini} &
    \multicolumn{1}{c|}{\underline{26.65}$_{ \pm \scalebox{0.7}{$.030$}}$} &
    \multicolumn{1}{c|}{\underline{25.20}$_{ \pm \scalebox{0.7}{$.025$}}$} &
    \multicolumn{1}{c|}{\underline{29.81}$_{ \pm \scalebox{0.7}{$.020$}}$} &
    \multicolumn{1}{c|}{\underline{16.87}$_{ \pm \scalebox{0.7}{$.015$}}$} &
    \multicolumn{1}{c|}{\underline{08.82}$_{ \pm \scalebox{0.7}{$.010$}}$} &
    \multicolumn{1}{c|}{\underline{16.72}$_{ \pm \scalebox{0.7}{$.005$}}$} &
    \multicolumn{1}{c|}{\underline{95.72}$_{ \pm \scalebox{0.7}{$1.245$}}$} &
    \multicolumn{1}{c|}{14.51$_{ \pm \scalebox{0.7}{$.025$}}$} &
    11.32$_{ \pm \scalebox{0.7}{$.010$}}$ \\
\end{tabular}
}
\label{main_cong_generation_JINA}
\end{table*}

\addtocounter{changecounter}{-1}
\begin{table*}[t]
  \centering
  % \footnotesize
  \scriptsize
  \caption{
  \changemark[Downstream RAG Generation Performance 3]{  
  Downstream RAG \textbf{Generation Performance}
   } of LLMs, utilizing \textit{BGE-Large} as the retriever.
  Results \underline{underlined} indicate superiority over RAG generation based on Oracle context, while \textbf{bold} results denote the best performance.} 
\renewcommand\arraystretch{1.2}
\setlength{\tabcolsep}{2.55mm}{
  \begin{tabular}{|ll|ccccccccc|}
    \specialrule{1pt}{0pt}{0pt}
  \multicolumn{2}{|l|}{\multirow{1}{*}{\textbf{{Datasets \& Methods}}}} &
    \multicolumn{9}{c|}{\textbf{Conventional Generation Evaluation (ConG Evaluator)}} 
    \\ \cline{3-11} 
    \multicolumn{2}{|l|}{\textbf{\textcolor{red}{Retriever: BGE-Large}}} &
    \multicolumn{1}{c|}{\textbf{ChrF}} &
    \multicolumn{1}{c|}{\textbf{ChrF++}} &
    \multicolumn{1}{c|}{\textbf{METEOR}} &
    \multicolumn{1}{c|}{\textbf{R1}} &
    \multicolumn{1}{c|}{\textbf{R2}} &
    \multicolumn{1}{c|}{\textbf{RL}} &
    \multicolumn{1}{c|}{\textbf{PPL}\(^{\circ}_{\downarrow}\)} &
    \multicolumn{1}{c|}{\textbf{CER}\(^{\circ}_{\downarrow}\)} &
    \multicolumn{1}{c|}{\textbf{WER}\(^{\circ}_{\downarrow}\)}  \\ \hline
  \multicolumn{1}{|l|}{\multirow{4}{*}{\rotatebox{90}{\texttt{\textbf{HQA}}}}} &
  \textbf{GPT-4o mini} &
    \multicolumn{1}{c|}{\underline{28.23}$_{ \pm \scalebox{0.7}{.030}}$} &
    \multicolumn{1}{c|}{\underline{26.85}$_{ \pm \scalebox{0.7}{.025}}$} &
    \multicolumn{1}{c|}{\underline{33.07}$_{ \pm \scalebox{0.7}{.020}}$} &
    \multicolumn{1}{c|}{\underline{17.96}$_{ \pm \scalebox{0.7}{.015}}$} &
    \multicolumn{1}{c|}{\underline{09.51}$_{ \pm \scalebox{0.7}{.010}}$} &
    \multicolumn{1}{c|}{\underline{17.86}$_{ \pm \scalebox{0.7}{.005}}$} &
    \multicolumn{1}{c|}{\underline{90.51}$_{ \pm \scalebox{0.7}{1.145}}$} &
    \multicolumn{1}{c|}{13.60$_{ \pm \scalebox{0.7}{.025}}$} &
   11.02$_{ \pm \scalebox{0.7}{.010}}$ \\
\end{tabular}
}
\label{main_cong_generation_BGE}
\end{table*}

\begin{table*}[t]
  \centering
  % \footnotesize
  \scriptsize
  \caption{Upstream \& Downstream performance of the RAG framework using Cognitive LLM Evaluation focusing on Separate Retrieval and Generation Metrics. The second row indicates the mean success rate of API requests \({P}_{sc}\) from three 100-entry samples, with green-highlighted results representing the average across 3-times sampling tests. }
  \renewcommand\arraystretch{1.25}
\setlength{\tabcolsep}{2.9mm}{
  \begin{tabular}{|ll|cc|cccccc|}
    \specialrule{1pt}{0pt}{0pt}
  \multicolumn{2}{|l|}{\multirow{1}{*}{\textbf{Datasets \& Methods}}} &
    \multicolumn{2}{c|}{\textbf{CogL [Retrieval]}} &
    \multicolumn{6}{c|}{\textbf{CogL [Generation]}} \\ \cline{3-10} 
    \multicolumn{2}{|l|}{\textbf{\textcolor{red}{Retriever: BGE-Large}}} &
    \multicolumn{1}{c|}{\textbf{Up-CRel}} &
    \textbf{Up-CCns} &
    \multicolumn{1}{c|}{\textbf{Dp-ARel}} &
    \multicolumn{1}{c|}{\textbf{Dp-RCmp}} &
    \multicolumn{1}{c|}{\textbf{Dp-RCnc}} &
    \multicolumn{1}{c|}{\textbf{Dp-RRel}} &
    \multicolumn{1}{c|}{\textbf{Dp-RVal}} &
    \multicolumn{1}{c|}{\textbf{Dp-RMch}} 
     \\ \hline 
     \multicolumn{1}{|l|}{\multirow{2}{*}{\rotatebox{90}{\texttt{\textbf{HQA}}}}} &
     \multirow{2}{*}{GPT-3.5 Turbo} &
     \multicolumn{1}{c|}{\cellcolor{c5}\textcolor{c4}{0.7352$_{ \pm \scalebox{0.7}{0.012}}$}} &
     \multicolumn{1}{c|}{\cellcolor{c5}\textcolor{c4}{0.9201$_{ \pm \scalebox{0.7}{0.023}}$}} &
     \multicolumn{1}{c|}{\cellcolor{c5}\textcolor{c4}{0.9367$_{ \pm \scalebox{0.7}{0.034}}$}} &
     \multicolumn{1}{c|}{\cellcolor{c5}\textcolor{c4}{0.7667$_{ \pm \scalebox{0.7}{0.045}}$}} &
     \multicolumn{1}{c|}{\cellcolor{c5}\textcolor{c4}{0.9358$_{ \pm \scalebox{0.7}{0.056}}$}} &
     \multicolumn{1}{c|}{\cellcolor{c5}\textcolor{c4}{0.7856$_{ \pm \scalebox{0.7}{0.067}}$}} &
     \multicolumn{1}{c|}{\cellcolor{c5}\textcolor{c4}{0.9733$_{ \pm \scalebox{0.7}{0.078}}$}} &
     \cellcolor{c5}\textcolor{c4}{0.1854$_{ \pm \scalebox{0.7}{0.089}}$} \\
\end{tabular}
}
\label{main_conl_generation_1}
\end{table*}

\begin{table*}[t]
  \centering
  % \footnotesize
  \scriptsize
  \caption{Upstream \& Downstream performance of the RAG framework using Cognitive LLM Evaluation focusing on Combined Retrieval and Generation Metrics.}
  \renewcommand\arraystretch{1.25}
\setlength{\tabcolsep}{2.88mm}{
  \begin{tabular}{|ll|cccccccc|}
    \specialrule{1pt}{0pt}{0pt}
    \multicolumn{2}{|l|}{\multirow{1}{*}{\textbf{Datasets \& Methods}}} &
    \multicolumn{8}{c|}{\textbf{CogL [Combined Retrieval \& Generation]}} \\ \cline{3-10} 
    \multicolumn{2}{|l|}{\textbf{\textcolor{red}{Retriever: BGE-Large}}} &
    \multicolumn{1}{c|}{\textbf{Dp-CPre}} &
    \multicolumn{1}{c|}{\textbf{Dp-CRec}} &
    \multicolumn{1}{c|}{\textbf{Dp-CRel}} &
    \multicolumn{1}{c|}{\textbf{Up-RCns}} &
    \multicolumn{1}{c|}{\textbf{Up-CUti}} &
    \multicolumn{1}{c|}{\textbf{Up-FAcc}} &
    \multicolumn{1}{c|}{\textbf{Dp-Fath}} &
    \textbf{Dp-Hall} \\ \hline
    
  \multicolumn{1}{|l|}{\multirow{2}{*}{\rotatebox{90}{\texttt{\textbf{HQA}}}}} &
    \multirow{2}{*}{GPT-3.5 Turbo} &
    \multicolumn{1}{c|}{\cellcolor{c5}\textcolor{c4}{0.9457$_{ \pm \scalebox{0.7}{0.022}}$}} &
    \multicolumn{1}{c|}{\cellcolor{c5}\textcolor{c4}{0.6455$_{ \pm \scalebox{0.7}{0.034}}$}} &
    \multicolumn{1}{c|}{\cellcolor{c5}\textcolor{c4}{0.6812$_{ \pm \scalebox{0.7}{0.046}}$}} &
    \multicolumn{1}{c|}{\cellcolor{c5}\textcolor{c4}{0.8318$_{ \pm \scalebox{0.7}{0.058}}$}} &
    \multicolumn{1}{c|}{\cellcolor{c5}\textcolor{c4}{0.6870$_{ \pm \scalebox{0.7}{0.070}}$}} &
    \multicolumn{1}{c|}{\cellcolor{c5}\textcolor{c4}{0.7516$_{ \pm \scalebox{0.7}{0.082}}$}} &
    \multicolumn{1}{c|}{\cellcolor{c5}\textcolor{c4}{0.9197$_{ \pm \scalebox{0.7}{0.094}}$}} &
    \cellcolor{c5}\textcolor{c4}{0.5537$_{ \pm \scalebox{0.7}{0.106}}$} \\
\end{tabular}
}
\label{main_conl_generation_2}
\end{table*}

\addtocounter{changecounter}{-1}
\begin{table*}[tb]
  \centering
  % \footnotesize
  \scriptsize
  % \tiny
    \caption{
      \changemark[Orchestrator_results]{
        Downstream RAG \textbf{Generation Performance} of various Orchestrators, utilizing \textit{deepseek-r1-distill-llama-70b}, which has a 128K context window, as the generative model.
    }}

\renewcommand\arraystretch{1.2}
\setlength{\tabcolsep}{1.85mm}{
  \begin{tabular}{|ll|l|ccccccccc|}
    \specialrule{1pt}{0pt}{0pt}
  \multicolumn{2}{|l|}{\multirow{2}{*}{\textbf{Datasets \& Methods}}} & \multicolumn{1}{|l|}{\multirow{2}{*}{\textbf{Orchestrator}}} & \multicolumn{9}{c|}{\textbf{Conventional Generation Evaluation (ConG Evaluator)}}    \\ \cline{4-12}
  {}  &{}  &{}                                 
  & \multicolumn{1}{c|}{\textbf{ChrF}} &
  \multicolumn{1}{c|}{\textbf{ChrF++}} &
  \multicolumn{1}{c|}{\textbf{METEOR}} &
  \multicolumn{1}{c|}{\textbf{R1}} &
  \multicolumn{1}{c|}{\textbf{R2}} &
  \multicolumn{1}{c|}{\textbf{RL}} &
  \multicolumn{1}{c|}{\textbf{PPL}\(^{\circ}_{\downarrow}\)} &
  \multicolumn{1}{c|}{\textbf{CER}\(^{\circ}_{\downarrow}\)} &
  \multicolumn{1}{c|}{\textbf{WER}\(^{\circ}_{\downarrow}\)} 
  \\ \hline
         
  \multicolumn{1}{|l|}{\multirow{6}{*}{\rotatebox{90}{\texttt{\textbf{HQA}}}}} & \textbf{BGE-Large}         &\textbf{Sequential }         &
      \multicolumn{1}{c|}{26.06$_{ \pm \scalebox{0.7}{$.023$}}$} &
      \multicolumn{1}{c|}{24.78$_{ \pm \scalebox{0.7}{$.012$}}$} &
      \multicolumn{1}{c|}{31.08$_{ \pm \scalebox{0.7}{$.015$}}$} &
      \multicolumn{1}{c|}{17.22$_{ \pm \scalebox{0.7}{$.010$}}$} &
      \multicolumn{1}{c|}{08.95$_{ \pm \scalebox{0.7}{$.017$}}$} &
      \multicolumn{1}{c|}{17.17$_{ \pm \scalebox{0.7}{$.010$}}$} &
      \multicolumn{1}{c|}{\textbf{79.91}$_{ \pm \scalebox{0.7}{$1.260$}}$} &
      \multicolumn{1}{c|}{18.39$_{ \pm \scalebox{0.7}{$.008$}}$} &
      14.26$_{ \pm \scalebox{0.7}{$.011$}}$ \\
\hline
\end{tabular}
}
\label{agentic_rag}
\end{table*}

\section{Experimental Setup}
\label{Experimental_Setup}
% \textbf{Settings}
% Both the retriever and Q\&A LLMs are essential modules of the RAG system. To focus on evaluating the Q\&A capabilities of different LLMs, we fixed the retriever to the BGE-LARGE model version, as the retriever serves as the primary entry point influencing RAG performance.  
% For document preprocessing, we utilized SentenceSplitter to divide documents into chunks and construct a vector index. SentenceSplitter was configured with a chunk size of 128 tokens, representing the maximum sequence length and a chunk overlap of 20 tokens, denoting the overlap between consecutive chunks. 
For document preprocessing, we used SentenceSplitter(chunk size: 128 tokens, overlap: 20 tokens) to segment documents into chunks and build a vector index.
We adhered to LlamaIndex configurations for other RAG components, including the refine module for response synthesis.
To ensure compatibility and efficiency, XRAG integrates Huggingface Transformers. All Q\&A LLMs were set with Temperature \(=0\) to ensure experimental consistency~\footnote{Nonetheless, using large models for generation or evaluation always results in minor variations, possibly due to randomness in MoE training or prompt interpretation biases~\cite{kaddour2023}.}. 
% For each query, five chunks were retrieved as contextual data. 
Consequently, the evaluation metrics measure retrieval accuracy based on 3 retrieval nodes, fully encompassing the assumptions of most datasets that typically consider only one or two golden contextual nodes. 
% The metrics include DCG, NDCG, and IDCG, with a search depth \(K=3\). 
The generator model's context window—encompassing the query, prompt, retrieved context, and response content—is configured to 4096 tokens. 
% This study conducts RAG testing across the entire test set using two rule-based evaluations: Conventional Retrieval Evaluation (ConR) and Conventional Generation Evaluation (ConG). 
% For the testing the orchestrators, to ensure fairness, we uniformly employ the large language model Deepseek-r1-distill-llama-70b as the generative model. 
% For components unique to certain methods (such as the critic module in SIM-RAG), we adhere to the original configurations established in the respective works.
% For details, please refer to Table~\ref{orchestrator_hyperparams}.
To ensure a fair comparison when testing the orchestrators, we used identical modules where feasible, as detailed in Table~\ref{orchestrator_hyperparams}.
% 说在框架外被argue了
% Notably, Cognitive LLM Evaluation (CogL) metrics are omitted from the main experimental framework, appearing only in pilot studies due to budget constraints from high computational costs of token processing during LLM reasoning and testing. 
\changemark[CogL_costs]{
Due to the high cost associated with token processing during LLM (OpenAI) reasoning and evaluation, Cognitive LLM Evaluation (CogL) metrics were included as a pilot study to demonstrate the XRAG framework's capability in semantic assessment.
}
% Analyses of RAG pilot studies using CogL are in Appendix~\ref{analysis_conl}. 
% Computational resources are detailed in Appendix~\ref{computation_resource}.
\changemark[standard deviations]{
  Experiments involving LLM generation were repeated three times, and standard deviations are reported in the tables of generation results.
}
\textbf{Computational Resource} Due to inherent design and size constraints, the deployment environments for each model are distinct. 
For the DQA dataset, testing was conducted on a single NVIDIA V100 GPU (32GB); the indexing phase (single-threaded) and the search-and-generation phase (four-threaded) required approximately one and two hours, respectively, with the BGE-Large index occupying 496 MB and the JINA-Large index occupying 391 MB of disk space. 
Similarly, on the NQA dataset using the same V100 hardware, the indexing and search phases took three and four hours, with index sizes of 5.4 GB for BGE-Large and 4.2 GB for JINA-Large. 
For the HQA dataset, an NVIDIA L20 GPU (48GB) was employed; both indexing and search processes were executed in a single-threaded manner, lasting four and eight hours, while the BGE-Large and JINA-Large indices utilized 18 GB and 14 GB of storage, respectively. 
Finally, the FQA dataset was evaluated on an NVIDIA RTX 4090 GPU (24GB), both single-threaded indexing and search/generation were completed in one hour each, with the BGE-Large and JINA-Large indices requiring 8.8 GB and 6.9 GB of disk space.

\section{Experimental Results}

% \textbf{Performance of modulor used for Retrieval}
% \subsection{Retrieval Evaluation on ConR}
\subsection{Performance of modulor used for Retrieval}
\label{Retrieval Evaluation on ConR}
The retrieval performance varies sensibly across the four datasets, with the poorest quality observed on DQA and FQA in Table~\ref{main_conr_retrieval}. 
Since DQA requires advanced discrete reasoning~\cite{RanLLZL19} and FQA demands paragraph understanding of financial documents, it presents a greater retrieval challenge. 

\textbf{i. Different datasets should focus on different testing metrics.} 
The NDCG metric reflects that the basic retrieval model performs reasonably well in relevance and ranking accuracy on HotpotQA and NaturalQA datasets. 
Ignoring ranking accuracy, a high Hit@10 score (>0.8) indicates a substantial likelihood of retrieving semantically relevant document blocks. 
The Hit@1 metric measures the ability to return the correct answer as the top-ranked result, making it suitable for tasks with a single retrieval target (e.g., the golden single passage in the NaturalQA dataset).
However, it is less effective for scenarios requiring multiple equally prioritized context targets, as seen in the HotpotQA dataset. Overall, the retrieval system performs well on HotpotQA and NaturalQA but faces challenges on DropQA, due to the complexity of the answer (Low Hit@1 scores reflect difficulty in retrieving the most relevant results.). 

% % 不同的检索模型适合不同domain的数据集
% \textbf{i. Different basic retrieval models are suited for datasets from different domains.} 
% As shown in Table~\ref{main_conr_retrieval}, the bge model performs well on the HQA, DQA, and NQA datasets, while the jina model performs better on the FQA dataset.
% By referring to Table~\ref{Summary_of_Datasets}, it can be observed that HQA, DQA, and NQA are all based on Wikipedia data, while FQA is financial domain data.
% This may be because the training data of the BGE-Large primarily consists of general-domain data such as news articles, video transcripts, and paragraph rankings, which allows it to perform well on wiki datasets. 
% In contrast, the training data for the JINA-Large is a multilingual and multi-domain comprehensive dataset, providing retrieval capabilities across various domains. 
% Although JINA-Large performs slightly less effectively on wiki datasets, which predominantly consist of knowledge-based data, it excels on the financial FQA dataset.

\textbf{ii. To improve retrieval performance, priority should be given to retrieval and post-retrieval.} %pre和post有用，可以精细化设计，特别是基于rrk的模型，可以给大模型更有用的信息，。
Compared to the basic retriever in Table~\ref{main_conr_retrieval} and the advanced retriever in Table~\ref{advanced_conr_retrieval}, the highlighted results indicate that adopting advanced retrieval modules, which integrate additional matching information, can enhance retrieval effectiveness. 
For example, RRFusion combines dense vectors with keyword retrieval, and the sentence window includes the context of chunk nodes in the retrieval objects. 
Specifically, it is observed that reranking methods can markedly improve performance, particularly for challenging retrieval tasks such as DQA and FQA. 
For instance, both BGE-RRK and JINA-RRK outperform the basic retriever.
This is because the reranker can evaluate the relevance of retrieved data, prioritizing content most likely to provide accurate and relevant answers. 
By allowing the LLM to focus on these top-ranked contexts during answer generation, it significantly enhances the accuracy and quality of responses.

\begin{figure*}[t]
    \centering    
    % First row: Retrieval metrics
    \begin{subfigure}[t]{0.16\linewidth}
        \centering
        \includegraphics[width=\linewidth]{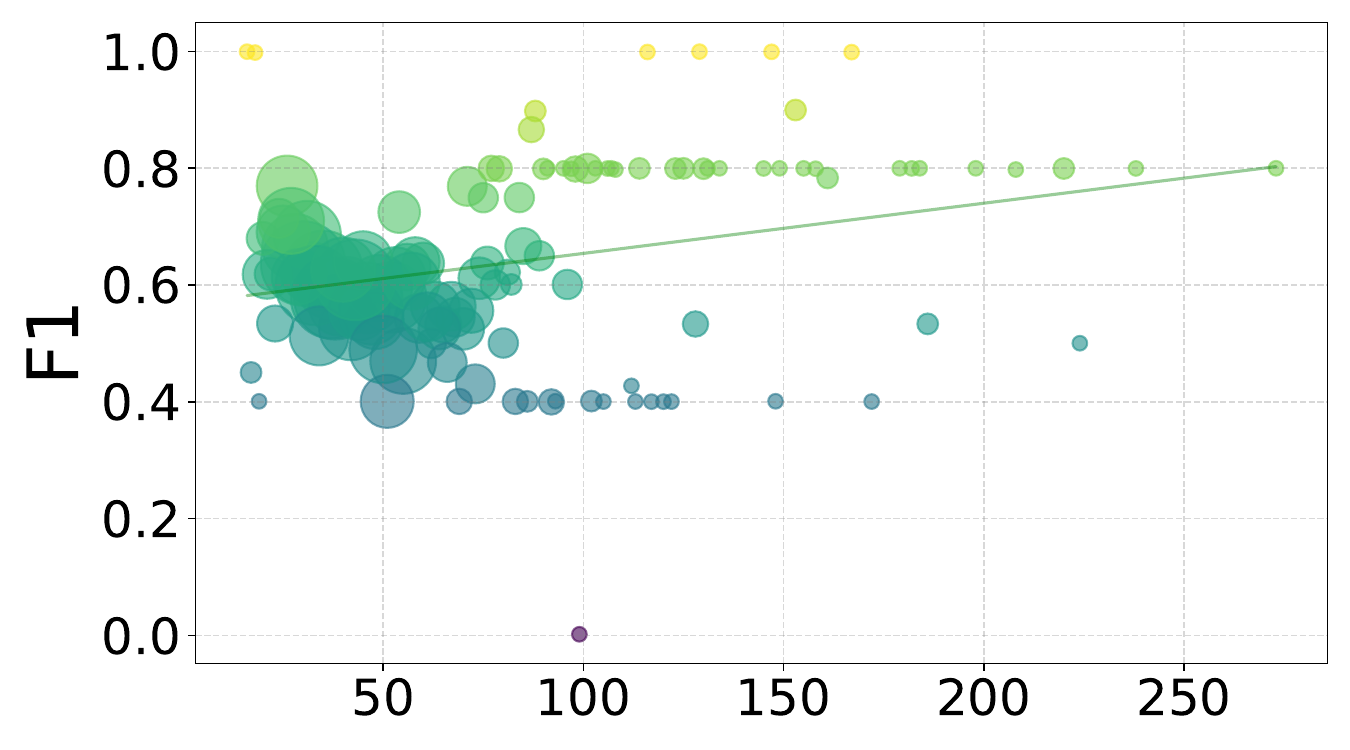}
        % \caption{\scriptsize F1}
    \end{subfigure}
    \hfill
    \begin{subfigure}[t]{0.16\linewidth}
        \centering
        \includegraphics[width=\linewidth]{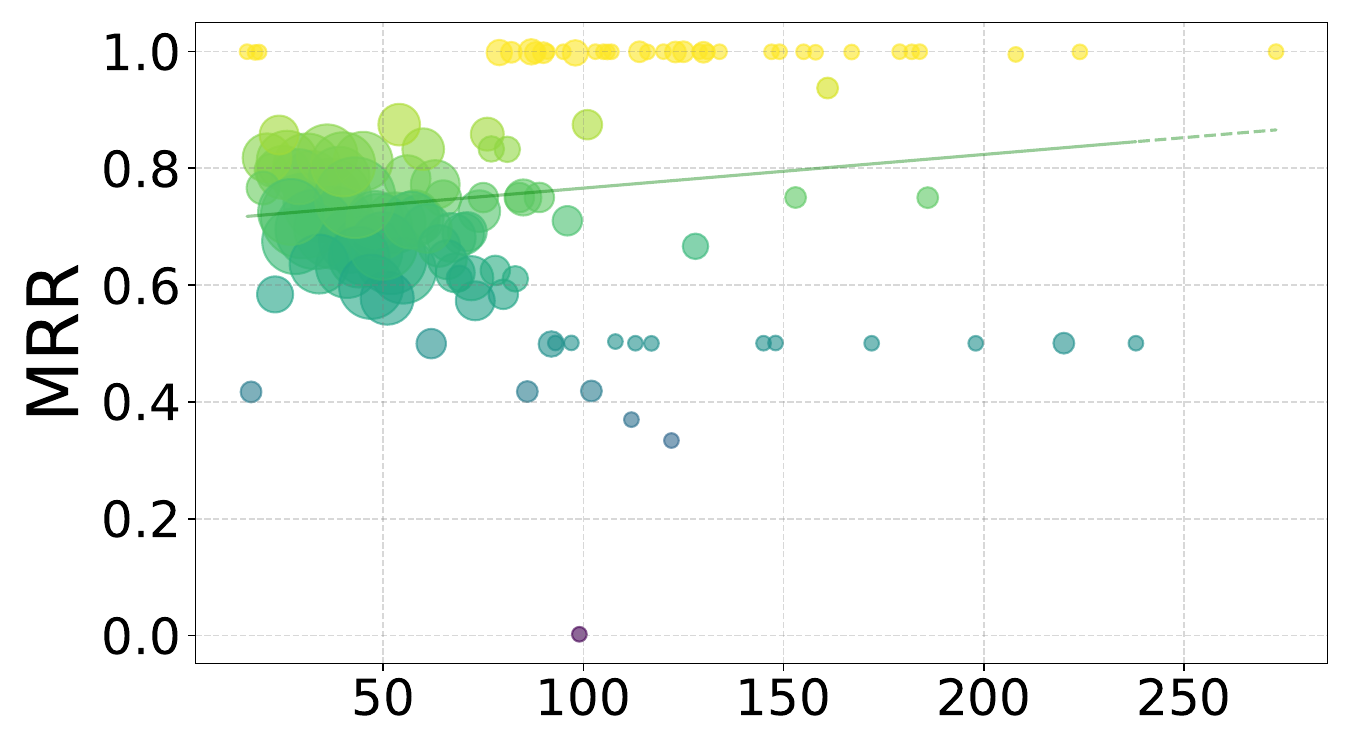}
        % \caption{\scriptsize MRR}
    \end{subfigure}
    \hfill
    \begin{subfigure}[t]{0.16\linewidth}
        \centering
        \includegraphics[width=\linewidth]{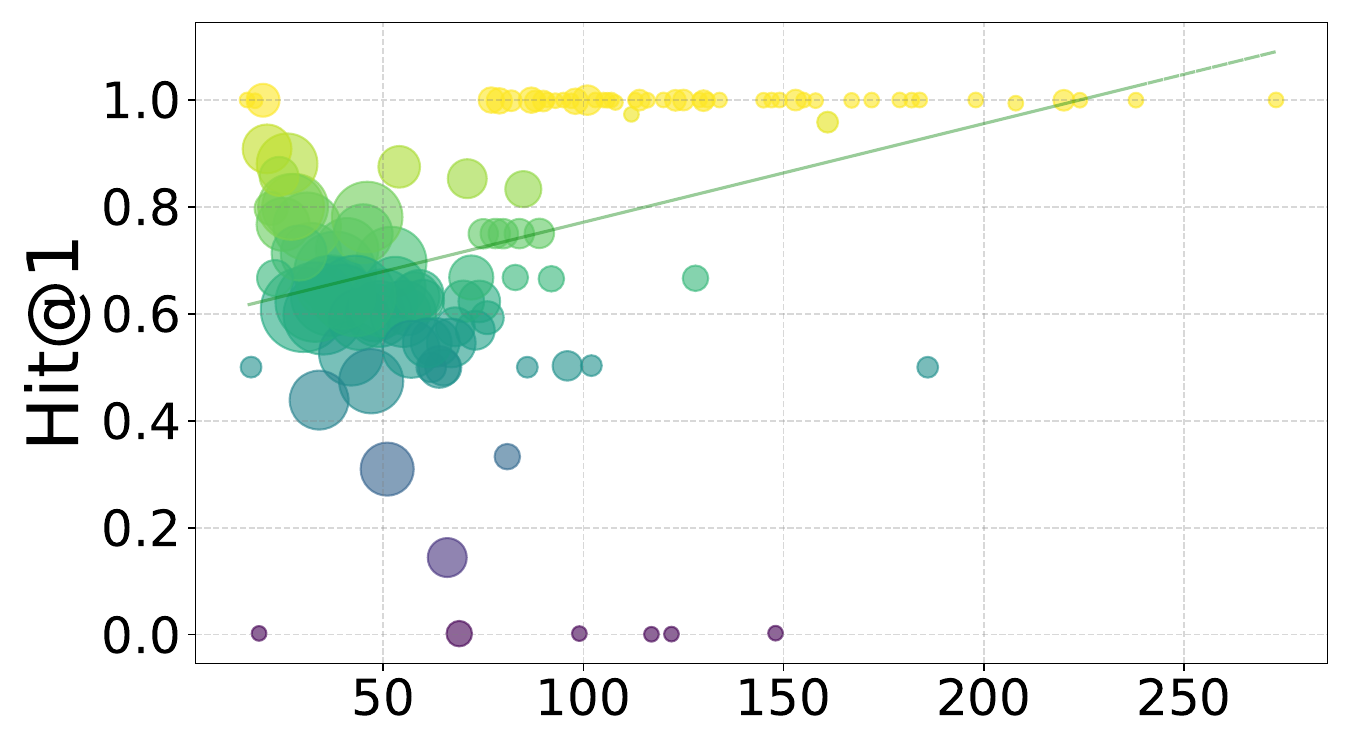}
        % \caption{\scriptsize Hit@1}
    \end{subfigure}
    \hfill
    \begin{subfigure}[t]{0.16\linewidth}
        \centering
        \includegraphics[width=\linewidth]{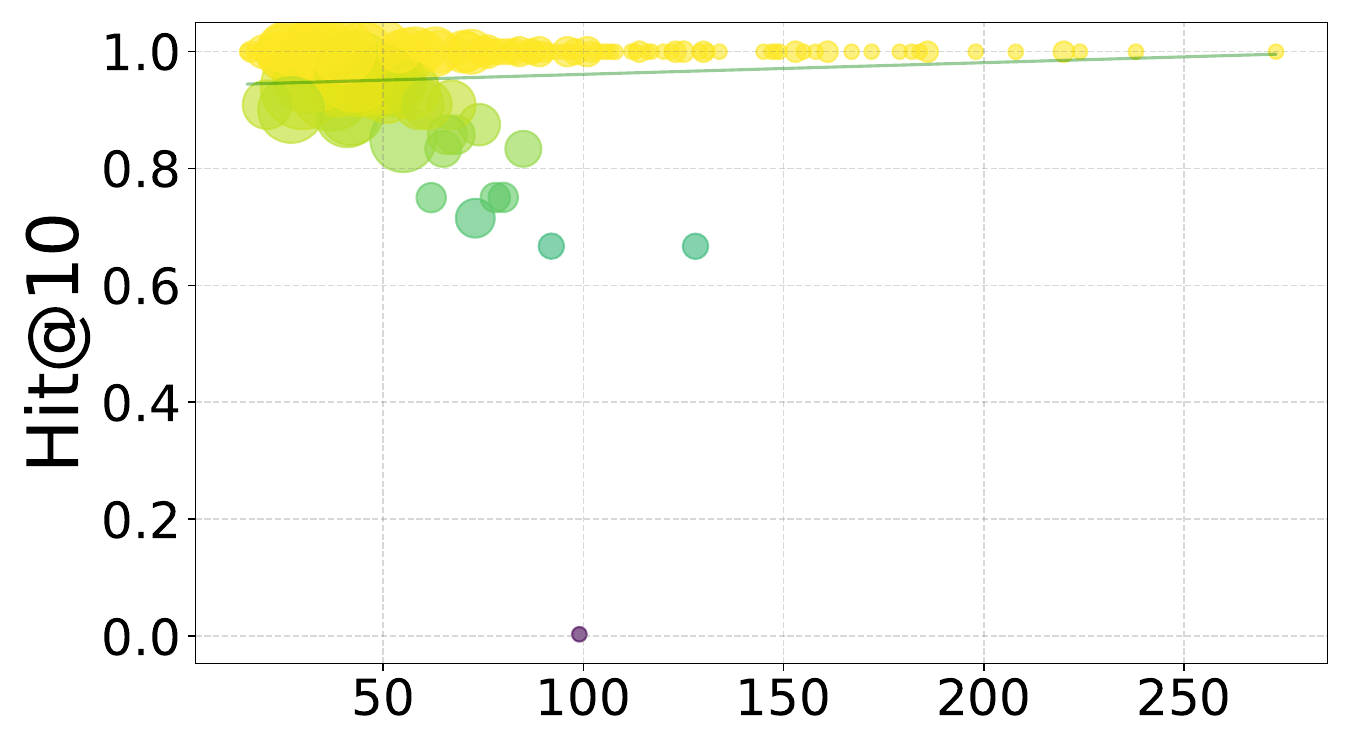}
        % \caption{\scriptsize Hit@10}
    \end{subfigure}
    \hfill
    \begin{subfigure}[t]{0.16\linewidth}
        \centering
        \includegraphics[width=\linewidth]{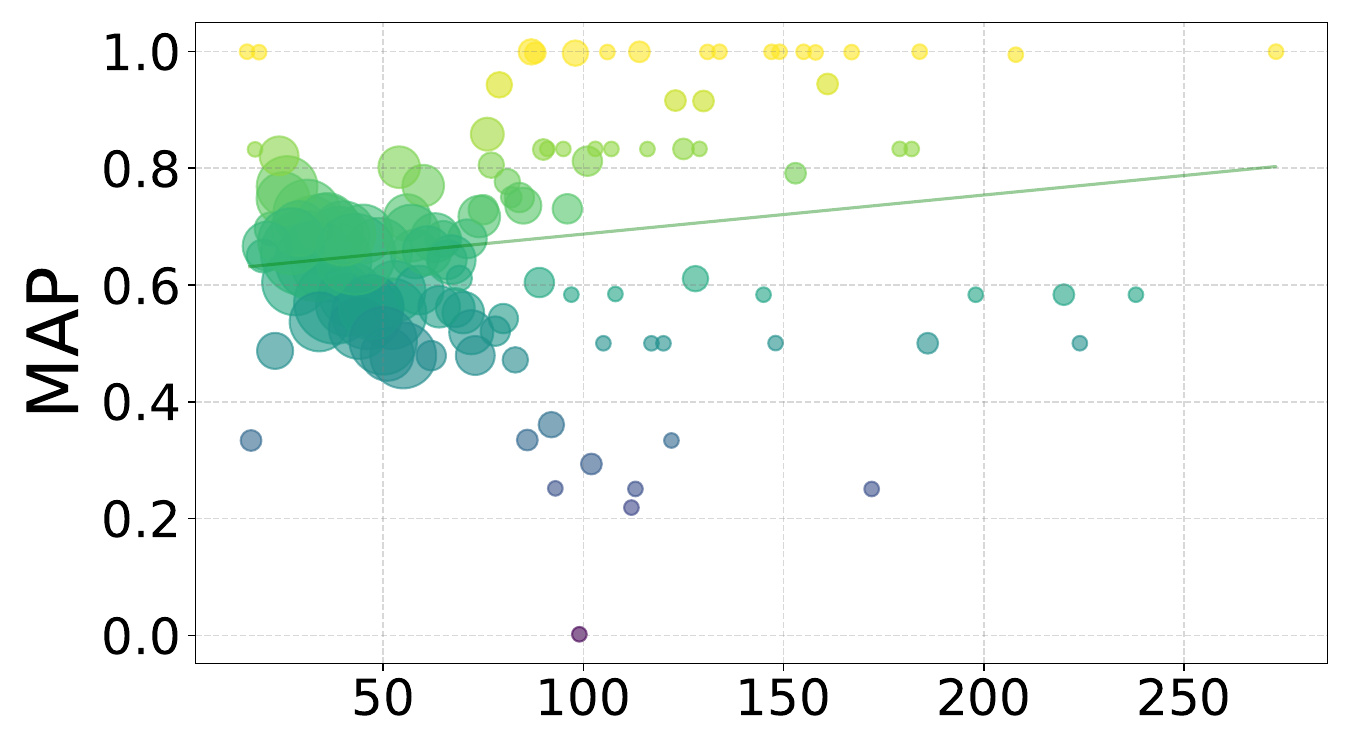}
        % \caption{\scriptsize MAP}
    \end{subfigure}
    \hfill
    \begin{subfigure}[t]{0.16\linewidth}
        \centering
        \includegraphics[width=\linewidth]{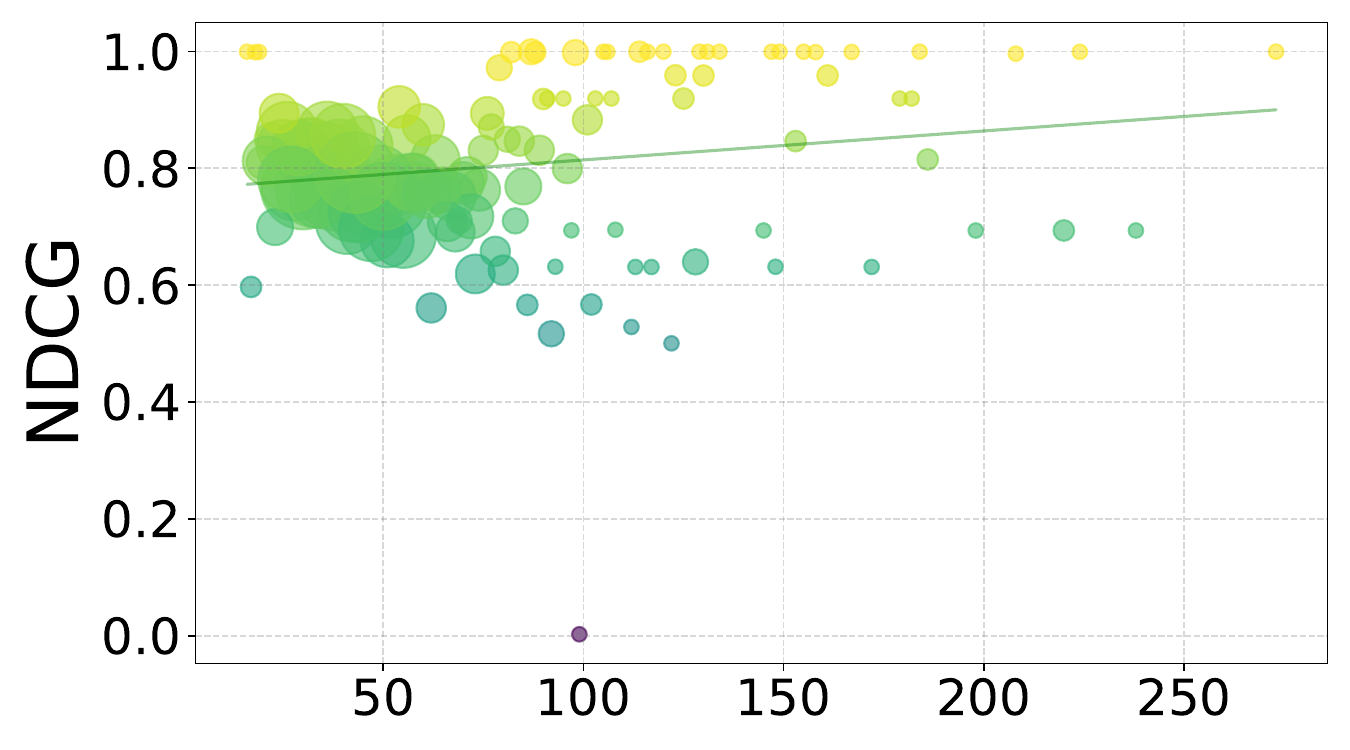}
        % \caption{\scriptsize NDCG}
    \end{subfigure}
  
    \vspace{0.1in}

    % \hfill
    % \begin{subfigure}[t]{0.185\linewidth}
    %     \centering
    %     \includegraphics[width=\linewidth]{pics/sec3-chrf.pdf}
    %     \caption{\scriptsize ChrF}
    % \end{subfigure}
    % \begin{subfigure}[t]{0.185\linewidth}
    %     \centering
    %     \includegraphics[width=\linewidth]{pics/sec3-R2.pdf}
    %     \caption{\scriptsize R-2}
    % \end{subfigure}
    % \hfill
    % \begin{subfigure}[t]{0.185\linewidth}
    %     \centering
    %     \includegraphics[width=\linewidth]{pics/sec3-RL.pdf}
    %     \caption{\scriptsize R-L}
    % \end{subfigure}
    
    \begin{subfigure}[t]{0.16\linewidth}
        \centering
        \includegraphics[width=\linewidth]{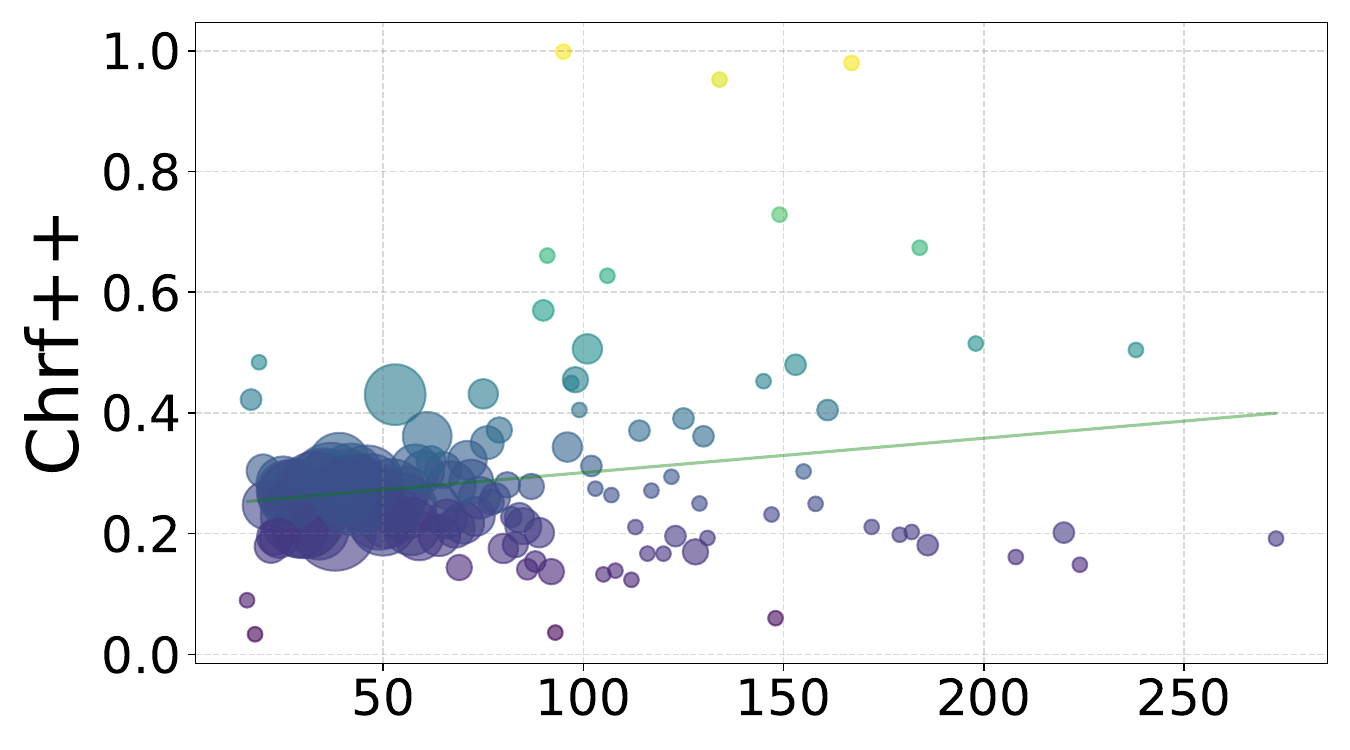}
        % \caption{\scriptsize ChrF++}
    \end{subfigure}
    \hfill
    \begin{subfigure}[t]{0.16\linewidth}
        \centering
        \includegraphics[width=\linewidth]{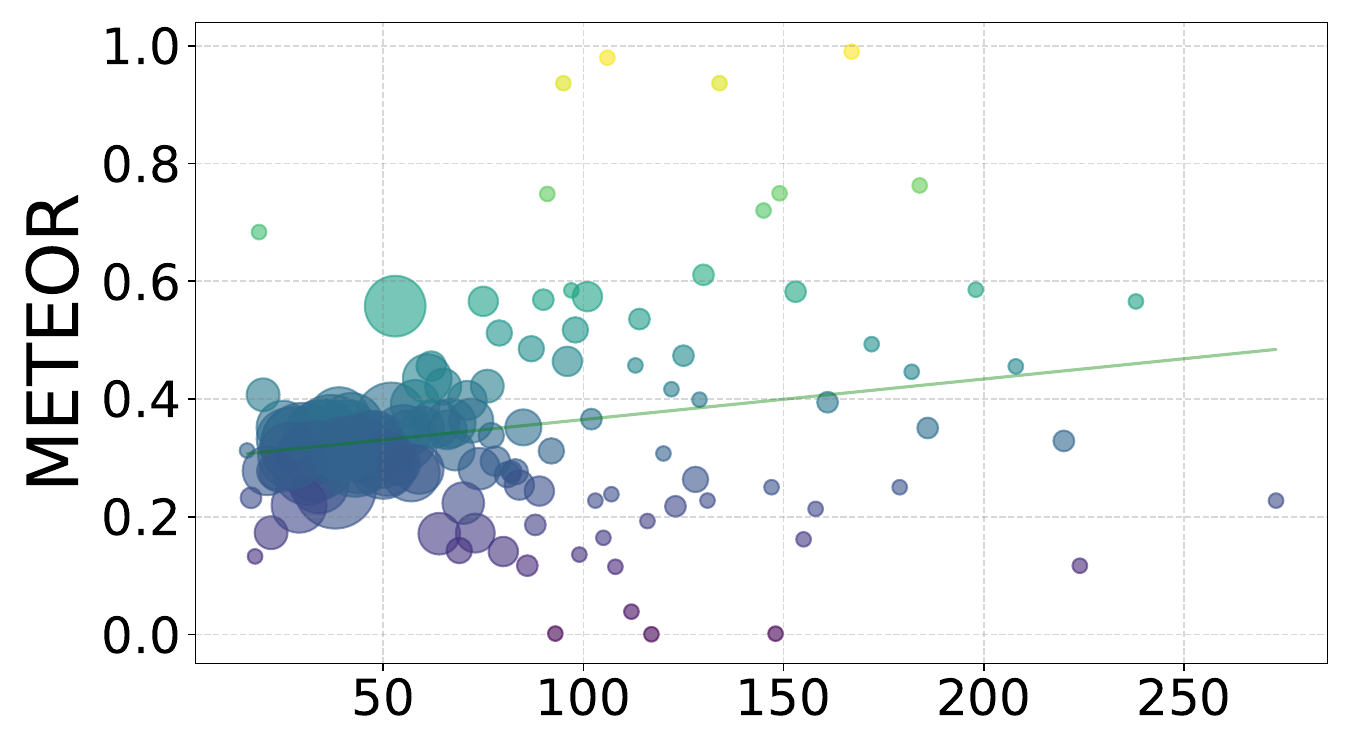}
        % \caption{\scriptsize METEOR}
    \end{subfigure}
    \hfill
    \begin{subfigure}[t]{0.16\linewidth}
        \centering
        \includegraphics[width=\linewidth]{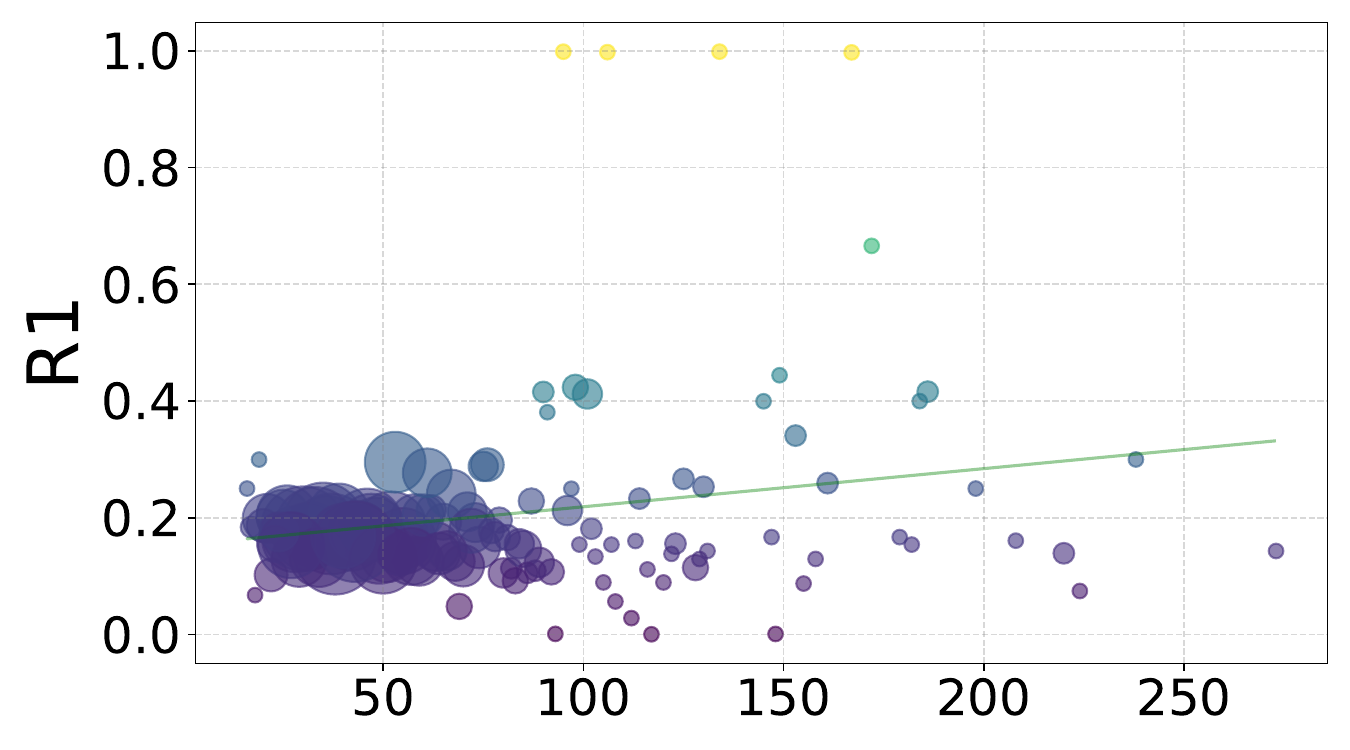}
        % \caption{\scriptsize R-1}
    \end{subfigure}
    \hfill
    \begin{subfigure}[t]{0.16\linewidth}
        \centering
        \includegraphics[width=\linewidth]{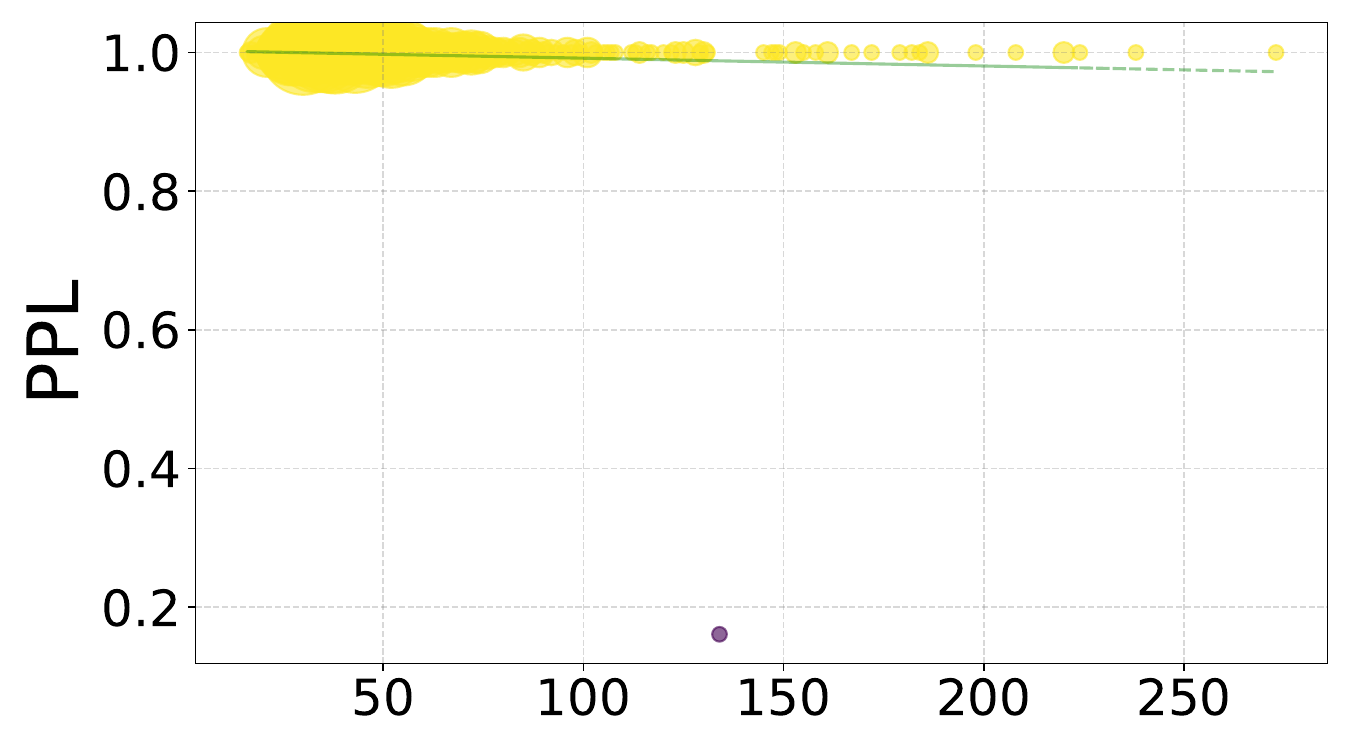}
        % \caption{\scriptsize PPL}
    \end{subfigure}
    \hfill
    \begin{subfigure}[t]{0.16\linewidth}
        \centering
        \includegraphics[width=\linewidth]{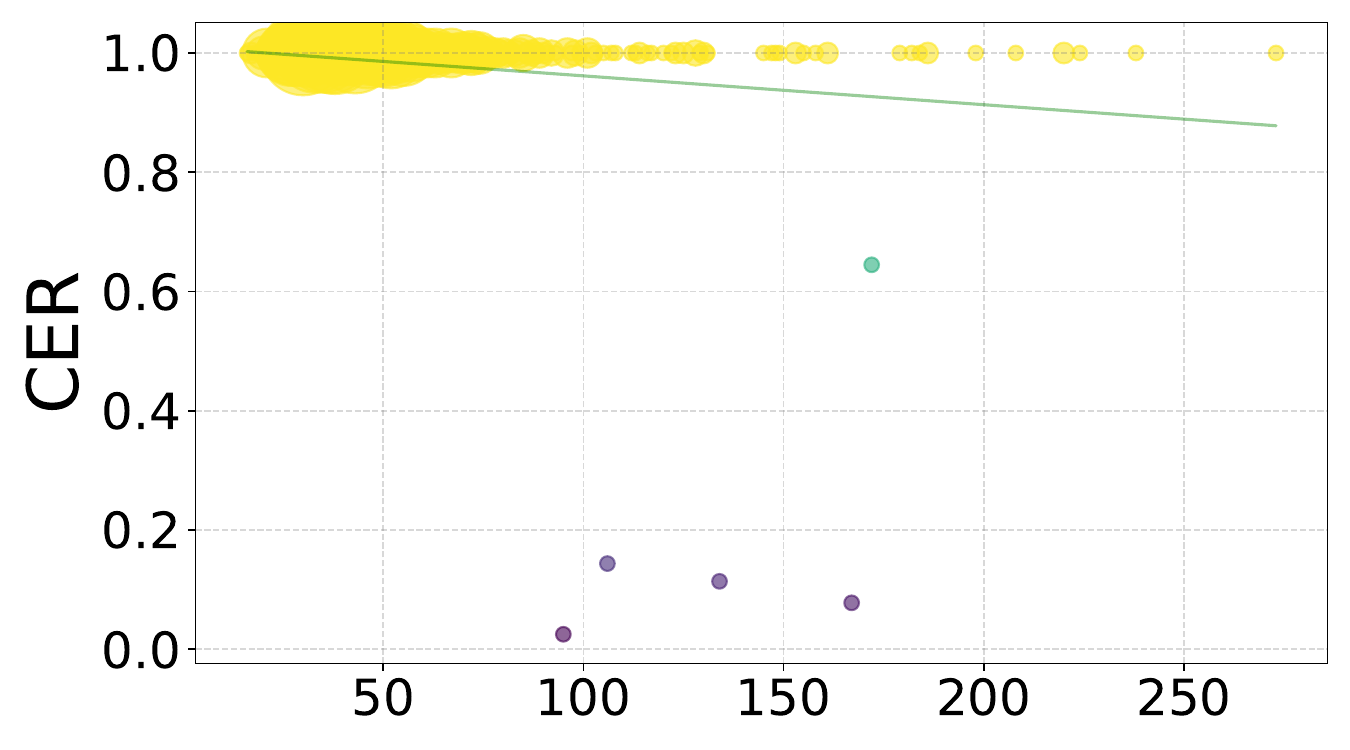}
        % \caption{\scriptsize CER}
    \end{subfigure}
    \hfill
    \begin{subfigure}[t]{0.16\linewidth}
        \centering
        \includegraphics[width=\linewidth]{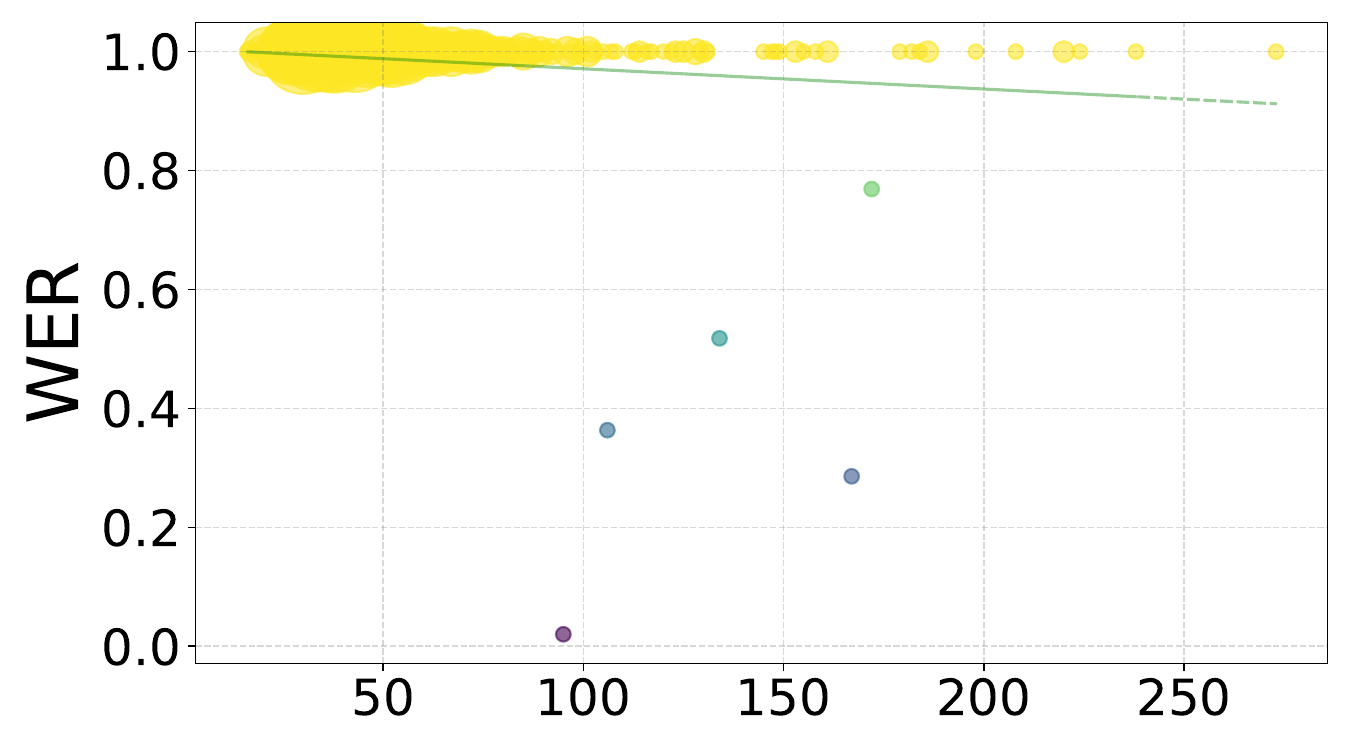}
        % \caption{\scriptsize WER}
    \end{subfigure}
    
    \caption{Comparison of retrieval and generation effects as query length varies. Experiments conducted on the HQA dataset (Test).}
    \label{dataset_vue_sec3}
\end{figure*}

\begin{figure*}[t]
    \centering
    % First row
    \begin{minipage}[t]{0.16\linewidth}
        \centering
        \includegraphics[width=\linewidth]{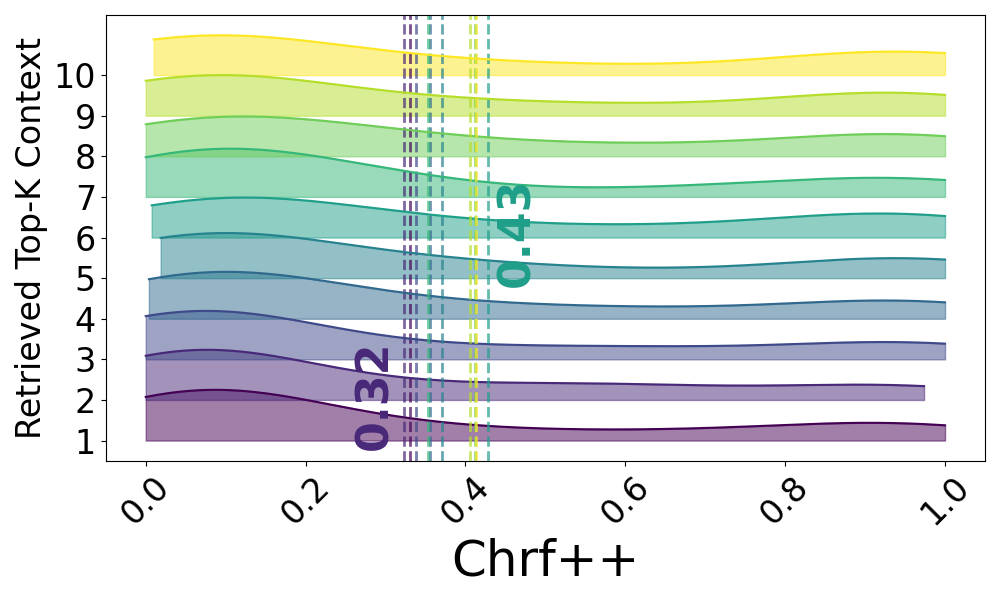}
        % \par\vspace{2pt}
        % {\scriptsize (a) ChrF++}
    \end{minipage}
    \hfill
    \begin{minipage}[t]{0.16\linewidth}
        \centering
        \includegraphics[width=\linewidth]{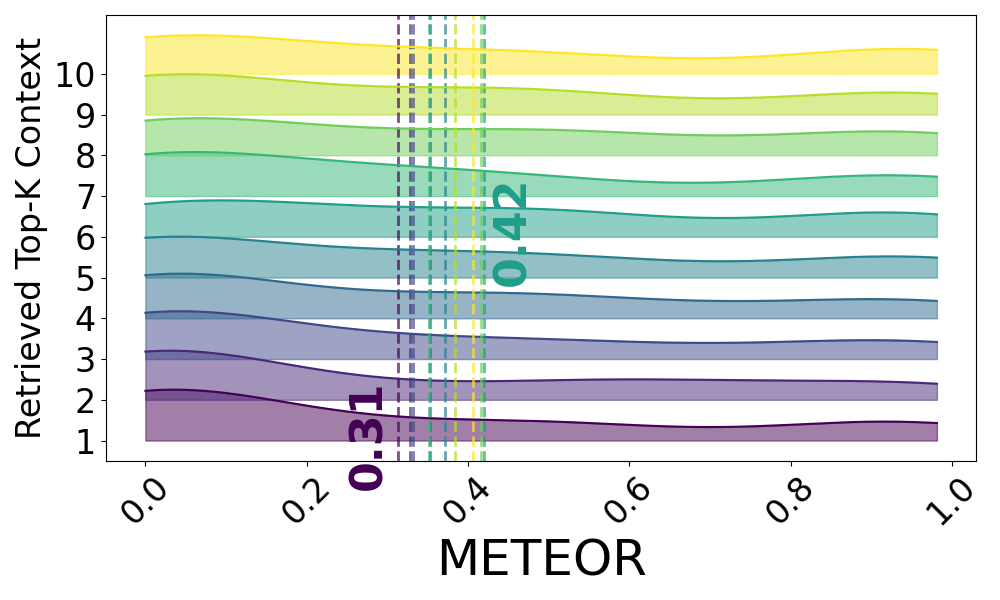}
        % \par\vspace{2pt}
        % {\scriptsize (b) METEOR}
    \end{minipage}
    \hfill
    \begin{minipage}[t]{0.16\linewidth}
        \centering
        \includegraphics[width=\linewidth]{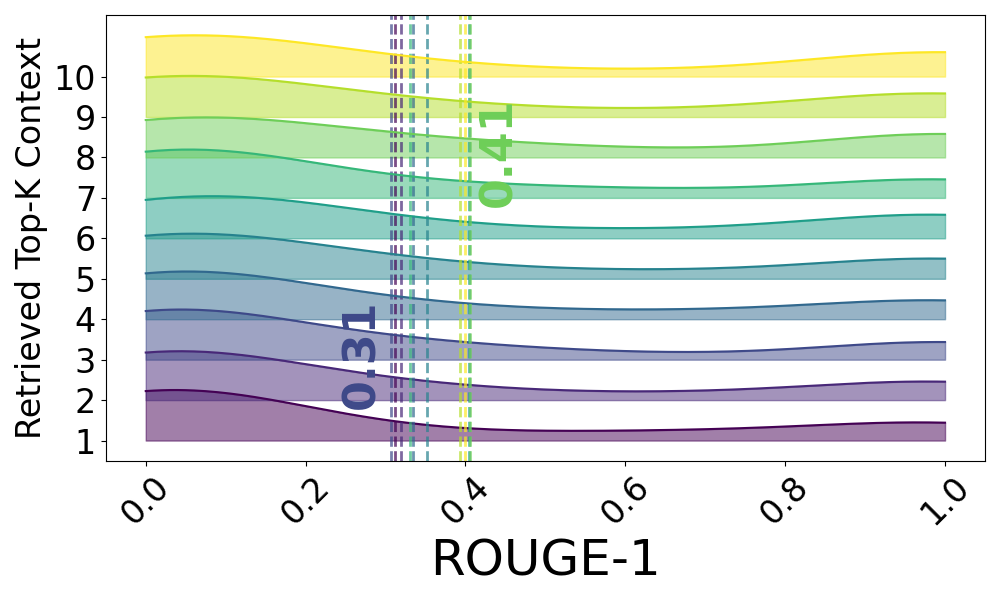}
        % \par\vspace{2pt}
        % {\scriptsize (c) ROUGE-1}
    \end{minipage}
    \hfill
    \begin{minipage}[t]{0.16\linewidth}
        \centering
        \includegraphics[width=\linewidth]{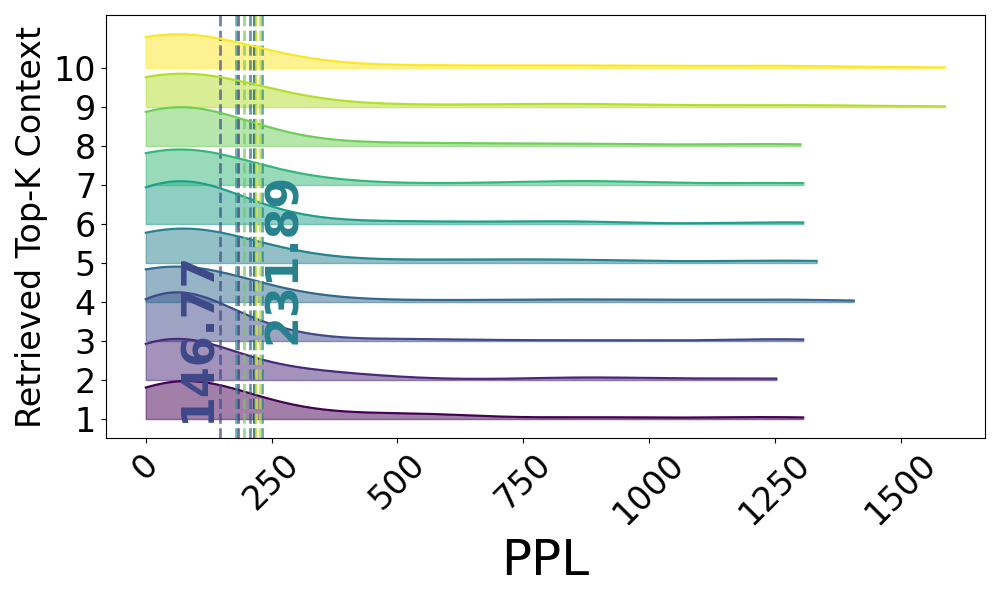}
        % \par\vspace{2pt}
        % {\scriptsize (d) PPL}
    \end{minipage}
    \hfill
    \begin{minipage}[t]{0.167\linewidth}
        \centering
        \includegraphics[width=\linewidth]{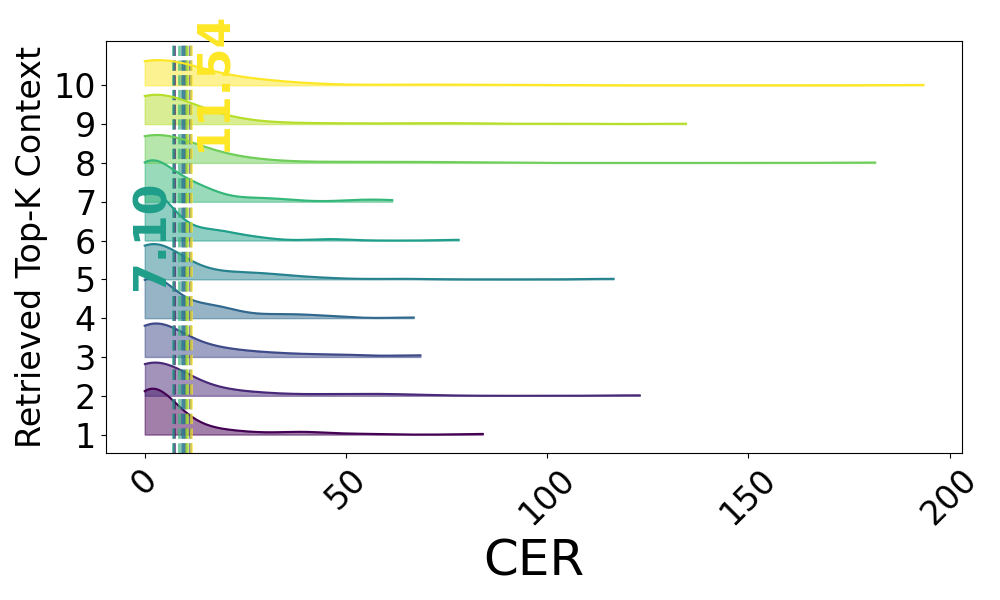}
        % \par\vspace{2pt}
        % {\scriptsize (e) CER}
    \end{minipage}
    \hfill
    \begin{minipage}[t]{0.16\linewidth}
        \centering
        \includegraphics[width=\linewidth]{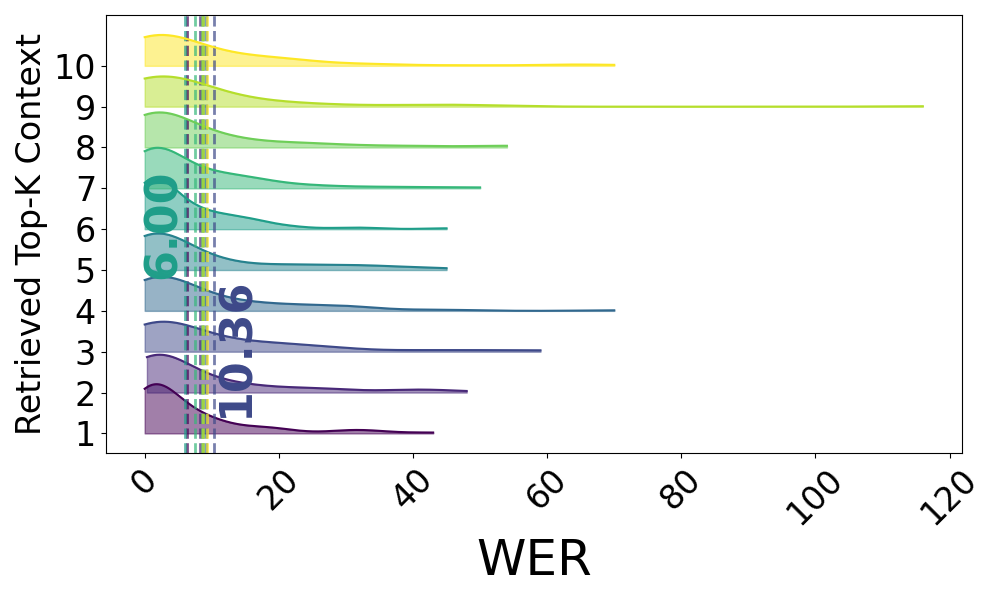}
        % \par\vspace{2pt}
        % {\scriptsize (f) WER}
    \end{minipage}
    % \hfill
    % \begin{minipage}[t]{0.23\linewidth}
    %     \centering
    %     \includegraphics[width=\linewidth]{pics/Figure_riddle_rouge2.pdf}
    %     \par\vspace{2pt}
    %     {\scriptsize (d) ROUGE-2}
    % \end{minipage}
    
    % % Second row
    % \vspace{0.1in}
    % \hfill
    % \begin{minipage}[t]{0.23\linewidth}
    %     \centering
    %     \includegraphics[width=\linewidth]{pics/Figure_riddle_rougel.pdf}
    %     \par\vspace{2pt}
    %     {\scriptsize (e) ROUGE-L}
    % \end{minipage}

    \caption{Ridge plots of the impact of varying numbers of retrieved contexts on Q\&A performance in the HQA dataset (Test). The chart shows maximum and minimum values.}
    \label{dataset_vue_sec4}
\end{figure*}

\subsection{Performance of modulor used for Generation}
% \subsection{Generation Evaluation on ConG}
\label{Generation Evaluation on ConG}
% \textbf{Performance of modulor used for Generation}
In examining the question-and-answer (Q\&A) component, it is essential to collectively consider Tables~\ref{main_cong_generation_Oracle}, ~\ref{main_cong_generation_JINA}, ~\ref{main_cong_generation_BGE} and ~\ref{agentic_rag}. 
Table~\ref{main_cong_generation_Oracle} provides the test results when the human-annotated retrieval output is directly entered into the large model for Q\&A purposes. 
% Table~\ref{main_cong_generation_JINA} illustrates the outcomes when the retriever employs the JINA-Large model to identify relevant sentences, which are subsequently fed into the large model. 
% Similarly, Table~\ref{main_cong_generation_BGE} demonstrates the Q\&A performance when the retriever uses the BGE-Large model to acquire pertinent sentences for the large model.
Table~\ref{main_cong_generation_JINA} and Table~\ref{main_cong_generation_BGE} illustrate the outcomes when the retriever employs the JINA-Large and BGE-Large models, respectively, to identify relevant sentences for the large model.
\changemark[Conl_analysis]{
Tables~\ref{main_conl_generation_1} and ~\ref{main_conl_generation_2} present the results of a Cognitive LLM evaluation conducted on the output generated by the XRAG system utilizing the BGE-Large retriever.}
In Table~\ref{main_conl_generation_1}, the Dp-RMch metric scores below 0.5, while all other metrics exceed 0.8. This suggests that Dp-RMch is unsuitable for most non-critical scenarios, as it may misclassify responses that are semantically correct.
The probability score \({P}_{sc}\) in  Table~\ref{main_conl_generation_1} and Table~\ref{main_conl_generation_2}  indicates that LLM testing depends on API requests, which may incur packet loss due to unstable requests, long processing times, timeout limits, or server-side errors. 
This reflects a particular weakness of LLM testing, even though the average request success rate reaches over 85\%.

\textbf{i. When the model parameter size is relatively small, directly supplying the large model with golden context may not yield optimal results for these datasets.}
Our analysis reveals that the experimental outcomes in Tables~\ref{main_cong_generation_JINA}, and ~\ref{main_cong_generation_BGE} generally surpass those in Table~\ref{main_cong_generation_Oracle}, with the exception of the results obtained using the DeepSeek-R1-70B model. 
% 删图对应文字
% As depicted in Fig.~\ref{dataset_vue}, 
The golden context typically comprises a single text passage (DQA and NQA) or two passages (HQA), which may provide inadequate information for the large model. 
Additionally, due to the limited reasoning capacity of small-scale LLMs, they may encounter difficulties in analyzing results from such concise texts, particularly for questions necessitating numerical computations and logical reasoning. 
Our retrieval model delivers the top k (top 3) results to the LLM, filtering and ranking context information to provide more comprehensive data. 
Some retrieved data, though not labeled as ``golden context'' for not being direct answers, are highly relevant to the query and contribute positively to the reasoning process.
This observation implies that for any retrieval-augmented generation task, evaluating only the retrieval results may not accurately represent the overall RAG performance; 
concurrent evaluation of both retrieval and large model Q\&A generation is imperative.

\textbf{ii. Choosing the right type of LLM for a RAG task could be superior to merely increasing the model's parameters.}
When comparing with various LLMs, it was observed that the DeepSeek R1-7B and the Deepseek R1-70B inference model exhibited the poor performance. 
The high hallucination rate of the DeepSeek model series~\cite{10.1145/3732796} is fundamentally at odds with the stringent accuracy demands of RAG scenarios.
At Wikipedia datasets, the most consistent performance was observed with the Llama3.1-8B model, which achieved optimal Q\&A results when combined with either JINA-Large or BGE-Large retrieval outputs. 
% Nonetheless, its perplexity (PPL) value was the highest, with a PPL of 181.71 on the HQA dataset, for instance. 
% It is important to note that PPL and Q\&A metrics assess different facets of the model: PPL emphasizes text fluency and generality, whereas ChrF and METEOR focus on the precise alignment between generated text and expected answers. 
% While Llama3.1-8B may excel in producing content closely aligned with expected answers in specific tasks, it may not manage semantic coherence and text fluency as effectively as other models in broader and more complex generation tasks.
Nonetheless, the perplexity (PPL) value of DeepSeek R1-7B and Deepseek R1-70B was the lowest. 
It is important to note that PPL and Q\&A metrics assess different facets of the model: PPL emphasizes text fluency and generality, whereas ChrF and METEOR focus on the precise alignment between generated text and expected answers. 
While the DeepSeek series of models may not be closely aligned with expected answers in specific tasks, yet their strengths lie in superior semantic coherence and text fluency.

% 通过表1中的结果可以看出，使用先进orchestrator的RAG方法在生成质量上有显著提升。例如，SIM-RAG在HQA数据集上，表现远高于其他方法（包括ORACLE）。
% 这表明迭代式的orchestrator在寻找最优的检索结果和生成结果的迭代过程中，能够更有效地整合检索到的信息，从而提升生成的准确性和相关性。
% 此外，RRFusion和Adapt-RAG也表现出了良好的性能，表明并行和混合式的orchestrator相比传统的Sequential orchestrator在处理复杂查询时更加具有优势。

\subsection{Performance of RAG Orchestrator}
% \subsection{Orchestrators Evaluation on ConG}
\label{Orchestrators Evaluation on ConG}
% \textbf{Performance of RAG orchestrator}
As shown in Table~\ref{agentic_rag}, RAG methods employing advanced orchestrators demonstrate substantial improvements in generation quality. 
In agentic RAG methodologies, SIM-RAG and Adapt-RAG demonstrate superior performance, whereas Self-RAG underperforms compared to the BGE retrieval results.
RRFusion, as a traditional parallel retrieval method, exhibits a performance gap when compared to agentic RAG approaches such as Adapt-RAG and SIM-RAG.

\textbf{i. Multi-hop question answering is more suitable for iterative orchestrators.}
In Table~\ref{agentic_rag}, SIM-RAG substantially outperforms other methods on the HQA dataset, which contains a high volume of multi-hop queries.
% This indicates that iterative orchestrators can more effectively integrate retrieved information through their cyclical process of optimizing retrieval results and generated content, thereby enhancing the accuracy and relevance of final outputs.
For multi-hop question-answering tasks, an iterative orchestrator can progressively retrieve information from each hop of the multi-hop data through their cyclical process of optimizing retrieval results and generated content, until the retrieved content is sufficient to answer the question or no more information can be obtained.
% Furthermore, Adapt-RAG demonstrates strong performance, which benefits from it combines both conditional and iterative orchestrators. 
% When dealing with complex multi-hop problems, the iterative retrieval component proves particularly effective.

\textbf{ii. Conditional orchestrators often rely on heuristic decision logic, resulting in relatively poor transferability.}
In Table~\ref{agentic_rag}, Self-RAG demonstrates underwhelming performance, which stems from the heuristic-based classifier within its conditional orchestrator. 
The classifier in conditional orchestrator often requires specific adjustments based on data characteristics and application scenarios, and fails to deliver effective classification when handling more complex datasets.

% Retrieval-oriented metrics assess context quality and consist of Context Relevance (Up-CRel) and Context Conciseness (Up-CCns), which come from UpTrain.  
% Response-oriented metrics include Response Relevance (Dp-ARel), Response Completeness (Up-RCmp) from DeepEval, Response Conciseness (Up-RCnc), Response Relevance (Up-RRel), and Response Validity (Up-RVal) and Response Matching (Up-RMch) from Uptrain. 
% Combined metrics evaluate the impact of retrieval on final responses and include Context Precision (Dp-CPre), Context Recall (Dp-CRec), Context Relevance (Dp-CRel), Response Consistency (Up-RCns),  Context Utilization (Up-CUti), and Factual Accuracy (Up-FAcc), Faithfulness (Dp-Faith), and Hallucination (Dp-Hall). The metrics prefixed with `Up' originate from UpTrain, while those prefixed with `Dp' are from DeepEval. 

\addtocounter{changecounter}{-1}
\begin{table}[t]
  \centering
    % \footnotesize
    \scriptsize
    % \tiny
    \caption{
      \changemark[consumption table]{
        Latency and total token consumption summary} of models on the full test set of the FQA dataset. 
        All methods were evaluated under identical hardware (RTX 4090, 20 vCPU Intel(R) Xeon(R) Platinum 8470Q, 30GB RAM).
      }
  \renewcommand\arraystretch{1.2}
  \setlength{\tabcolsep}{0.55mm}{
    \begin{tabular}{|l|ccccc|c|}
      \specialrule{1pt}{0pt}{0pt}
    \multirow{2}{*}{\textbf{Methods}} &
      \multicolumn{5}{c|}{\textbf{Latency(s)}} &
      \multirow{2}{*}{\textbf{Token(k)}}  \\ \cline{2-6}
              & \multicolumn{1}{c|}{\textbf{Index}}   & \multicolumn{1}{c|}{\textbf{Pre-retrieval}} & \multicolumn{1}{c|}{\textbf{Retrieval}}  & \multicolumn{1}{c|}{\textbf{Post-processor}} & \multicolumn{1}{c|}{\textbf{Generation}}& \\ \hline
              % \textbf{ORACLE}  & \multicolumn{1}{c|}{N/A}  & \multicolumn{1}{c|}{N/A}      & \multicolumn{1}{c|}{N/A}   & \multicolumn{1}{c|}{N/A}   & \multicolumn{1}{c|}{} & \multicolumn{1}{c|}{} \\ \hline
              \textbf{BGE-Large}    & \multicolumn{1}{c|}{2087}  & \multicolumn{1}{c|}{\usym{2717}}      & \multicolumn{1}{c|}{1641}   & \multicolumn{1}{c|}{\usym{2717}}     &  \multicolumn{1}{c|}{296} &  \multicolumn{1}{c|}{161} \\ \hline
              \textbf{JINA-Large}    & \multicolumn{1}{c|}{2105}  & \multicolumn{1}{c|}{\usym{2717}}      & \multicolumn{1}{c|}{1459}   & \multicolumn{1}{c|}{\usym{2717}}     &  \multicolumn{1}{c|}{256} &  \multicolumn{1}{c|}{165} \\ \hline
              \textbf{SBPT} & \multicolumn{1}{c|}{2087} & \multicolumn{1}{c|}{183}     & \multicolumn{1}{c|}{1489} & \multicolumn{1}{c|}{\usym{2717}}    &  \multicolumn{1}{c|}{301} &  \multicolumn{1}{c|}{232}  \\  \hline
              \textbf{HyDE} & \multicolumn{1}{c|}{2087} & \multicolumn{1}{c|}{492}     & \multicolumn{1}{c|}{1410} & \multicolumn{1}{c|}{\usym{2717}}    &  \multicolumn{1}{c|}{288} &  \multicolumn{1}{c|}{441}  \\  \hline
              \textbf{RRFusion} & \multicolumn{1}{c|}{2087} & \multicolumn{1}{c|}{\usym{2717}}     & \multicolumn{1}{c|}{7119} & \multicolumn{1}{c|}{\usym{2717}}    &  \multicolumn{1}{c|}{256} &  \multicolumn{1}{c|}{98} \\  \hline
              \textbf{StParser} & \multicolumn{1}{c|}{2087} & \multicolumn{1}{c|}{\usym{2717}}     & \multicolumn{1}{c|}{1921} & \multicolumn{1}{c|}{\usym{2717}}    &  \multicolumn{1}{c|}{159} &  \multicolumn{1}{c|}{63}  \\  \hline
              \textbf{BGE-RRK} & \multicolumn{1}{c|}{2087} & \multicolumn{1}{c|}{\usym{2717}}     & \multicolumn{1}{c|}{1836} & \multicolumn{1}{c|}{1232}    &  \multicolumn{1}{c|}{235}  & \multicolumn{1}{c|}{163} \\  \hline
              \textbf{JINA-RRK} & \multicolumn{1}{c|}{2087} & \multicolumn{1}{c|}{\usym{2717}}     & \multicolumn{1}{c|}{2106} & \multicolumn{1}{c|}{1387}    & \multicolumn{1}{c|}{253} & \multicolumn{1}{c|}{165}  \\  \specialrule{1pt}{0pt}{0pt}
              \multirow{2}{*}{\textbf{Methods}} &
      \multicolumn{5}{c|}{\textbf{Latency(s)}} &
      \multirow{2}{*}{\textbf{Token(k)}}  \\ \cline{2-6}
              & \multicolumn{1}{c|}{\textbf{Index}}   & \multicolumn{1}{c|}{\textbf{Critic}} & \multicolumn{1}{c|}{\textbf{Classifier}}  & \multicolumn{1}{c|}{\textbf{Retrieval}} & \multicolumn{1}{c|}{\textbf{Generation}}& \\ \hline
              \textbf{Adapt-RAG}  & \multicolumn{1}{c|}{2087}  & \multicolumn{1}{c|}{1215}      & \multicolumn{1}{c|}{1215}   & \multicolumn{1}{c|}{1215}   & \multicolumn{1}{c|}{3118} & \multicolumn{1}{c|}{179}\\ \hline
              \textbf{SIM-RAG}    & \multicolumn{1}{c|}{2087}  & \multicolumn{1}{c|}{59}      & \multicolumn{1}{c|}{\usym{2717}}   & \multicolumn{1}{c|}{4085}     &  \multicolumn{1}{c|}{5346} & \multicolumn{1}{c|}{871}\\ \hline
              \textbf{Self-RAG} & \multicolumn{1}{c|}{2087} & \multicolumn{1}{c|}{\usym{2717}}     & \multicolumn{1}{c|}{42} & \multicolumn{1}{c|}{1881}    & \multicolumn{1}{c|}{225} & \multicolumn{1}{c|}{98}  \\  \specialrule{1pt}{0pt}{0pt}
    \end{tabular}
  }
  \label{Calculate Source}
\end{table}

\subsection{Latency and Token Consumption}
\label{Overhead Analysis}
\changemark[Latency_and_token_2]{
  As shown in Table~\ref{Calculate Source}, 
  % different retrieval methods exhibit distinct latency and LLM token consumption. 
  taking the vanilla RAG methods (e.g., BGE-Large and JINA-Large) as a reference, pre-retrieval methods such as SBPT (71k) and HyDE (280k) introduce only a slight increase of LLM token usage. 
  While advanced retrieval methods (e.g., RRFusion and StParser) come at the cost of considerably greater retrieval latency (7119s for RRFusion, 1921s for StParser), they compensate by consuming fewer LLM tokens, thanks to heuristic optimizations applied during retrieval.
  Post-processors add some latency during the reranking stage. 
  Orchestrators typically involve complex retrieval logic, including multiple retrieval steps and multiple rounds of LLM interactions, resulting in higher latency and greater token consumption.
  Considering the performance (as shown in Table~\ref{agentic_rag} and Table~\ref{advanced_conr_retrieval}) of retrieval modules alongside their latency and LLM token costs (as shown in Table~\ref{Calculate Source}), 
  The rerank method is highly recommendable, as it consistently demonstrates stable performance gains across all datasets (e.g., Hit@1: +73.9\% for BGE-RRK vs. BGE-Large on FQA) without incurring substantial increases in latency (+11.8\% for BGE-RRK vs. BGE-Large) or token consumption (+1.2\% for BGE-RRK vs. BGE-Large). 
  In latency-insensitive scenarios, employing an orchestrator is also a favorable option, where a more sophisticated retrieval workflow can trade computational resources for enhanced retrieval and generation performance.
}

% \subsection{Retrieval and Generation Evaluation on CogL}
% \label{Retrieval and Generation Evaluation on CogL}
% In the benchmark experiments evaluated by LLM retrieval and response metrics (Table~\ref{main_conl_generation_1} and  Table~\Ref{main_conl_generation_2}), we set \(E_{sp}\) to 100. 
% For the evaluation of RAG failures, we set \(E_{sp}\) to 20, considering the test quantity to be adequate, given that we have specifically curated datasets to investigate failures. 
% Tables~\ref{main_conl_generation_1} and ~\ref{main_conl_generation_2} present the results of a Cognitive LLM evaluation conducted on the output generated by the XRAG system utilizing the BGE-Large retriever.
% In Table~\ref{main_conl_generation_1}, the Dp-RMch metric scores below 0.5, while all other metrics exceed 0.8. This suggests that Dp-RMch is unsuitable for most non-critical scenarios, as it may misclassify responses that are semantically correct.
% The probability score \({P}_{sc}\) in  Table~\ref{main_conl_generation_1} and Table~\ref{main_conl_generation_2}  indicates that LLM testing depends on API requests, which may incur packet loss due to unstable requests, long processing times, timeout limits, or server-side errors. 
% This reflects a particular weakness of LLM testing, even though the average request success rate reaches over 85\%.  

\section{Additional Discussion \& Analysis}
\label{Analysis and Discussion}

\textbf{Q1: Whether longer queries have a greater impact?}
The illustration (as shown in Fig.~\ref{dataset_vue_sec3}) indicates a positive correlation between query length and the performance of several metrics, notably Hit@1 and METEOR, which demonstrate an ascending trajectory with increased query length. 
This correlation implies that the utilization of longer queries may augment the efficacy of retrieval and query-answering systems. 
Such enhancements are likely due to the enriched information that longer queries provide, facilitating a more precise interpretation of queries.

\textbf{Q2:Whether more context has a greater impact?}
The illustration depicted in Fig.~\ref{dataset_vue_sec4} demonstrates that the performance of query-answering systems remains relatively stable as the number of retrieval contexts increases. 
This observation suggests that augmenting the quantity of retrieval contexts may not significantly enhance model performance.
Upon analyzing the mean value ranges, it is evident that for instance, in the ROUGE-1 metric, the mean value only moderately increases from 0.31 to 0.41 with the augmentation of retrieval contexts. 
This modest variation implies that the impact of increasing the number of contexts on performance improvement is limited. Furthermore, the observation of maximum values reaching 1.0 in multiple figures indicates that the key to enhancing performance may lie in improving the precision of retrieval. This involves identifying a few contexts most relevant to the query, rather than merely increasing the number of retrieved contexts. 
The ChrF++ and METEOR metrics exhibit relatively stable performance across varying numbers of retrieval contexts. 
This suggests that these metrics are not particularly sensitive to variations in the number of contexts. 
Consequently, this underscores the importance of focusing on enhancing the relevance and precision of retrieval results in the optimization of the retrieval module within RAG systems, rather than increasing the quantity of retrieved contexts.

\textbf{Q3: Whether problem hard induce challenges in both retrieval and generation processes?}
\changemark[difficulty]{
As shown in Tables~\ref{difficult_R_results} and ~\ref{difficult_G_results}, consistent with expectations, both retrieval and generative models achieve higher scores on easy datasets compared to harder datasets. 
From an evaluation perspective, more stringent metrics—such as F1 and Hit@1 for retrieval, and ChrF and METEOR for generation—are notably sensitive to dataset difficulty (e.g., On the Hit@1 metric, BGE-Large scores 97.00 on the easy dataset but only 70.00 on the hard one, a decline of approximately 28\%). 
This sensitivity occurs because stricter metrics are more effective at detecting performance declines, which are inherent as question difficulty increases.
% Therefore, in tasks requiring classification of questions by difficulty, employing such stringent evaluation metrics may facilitate more effective difficulty categorization.
% From the perspective of model performance, under datasets of comparable difficulty, BGE-Large outperforms JINA-Large, and GPT outperforms ds-70b. 
% This indicates that once the dataset type is fixed, question difficulty affects only the absolute scores of the models, without altering their relative performance rankings.
}

\addtocounter{changecounter}{-1}
\begin{table}[t]
  \centering
    % \footnotesize
    \scriptsize
    % \tiny
    \caption{
      \changemark[difficulty_r_table]{
        Performance of different retrieval models on subsets of varying difficulty (Easy/Med./Hard) in HotpotQA, utilizing GPT-4o-mini as the generative model.}
      }
  \renewcommand\arraystretch{1.25}
  \setlength{\tabcolsep}{1.9mm}{
    \begin{tabular}{|c>{\raggedright\arraybackslash}p{2cm}|*{4}{>{\centering\arraybackslash}p{1.05cm}|}}
      \specialrule{1pt}{0pt}{0pt}
    \multicolumn{2}{|l|}{\multirow{2}{*}{\textbf{Difficulty \& Models}}} & \multicolumn{4}{c|}{Conventional Retrieval Evaluation (ConR Evaluator)} \\ \cline{3-6}% \textbf{ORACLE}  & \multicolumn{1}{c|}{N/A}  & \multicolumn{1}{c|}{N/A}      & \multicolumn{1}{c|}{N/A}   & \multicolumn{1}{c|}{N/A}   & \multicolumn{1}{c|}{} & \multicolumn{1}{c|}{} \\ \hline
    &  & \multicolumn{1}{c|}{\textbf{F1}} & 
    \multicolumn{1}{c|}{\textbf{Hit@1}} & 
    \multicolumn{1}{c|}{\textbf{Hit@10}} & 
    \textbf{MAP} \\ \hline        
    \multirow{2}{*}{\rotatebox{90}{\texttt{\textbf{Easy}}}} 
    & \multicolumn{1}{|l|}{\textbf{BGE-Large}} & 36.21 & 97.00 & 99.00 & 18.39 \\ \cline{2-6}
     & \multicolumn{1}{|l|}{\textbf{JINA-Large}} & \multicolumn{1}{c|}{35.48} & \multicolumn{1}{c|}{88.00} & \multicolumn{1}{c|}{99.00} & 17.17 \\ \hline
    \multirow{2}{*}{\rotatebox{90}{\texttt{\textbf{Med.}}}} 
    & \multicolumn{1}{|l|}{\textbf{BGE-Large}} & \multicolumn{1}{c|}{32.38} & \multicolumn{1}{c|}{86.00} & \multicolumn{1}{c|}{100.00} & 16.68 \\ \cline{2-6}
     & \multicolumn{1}{|l|}{\textbf{JINA-Large}} & \multicolumn{1}{c|}{29.30} & \multicolumn{1}{c|}{73.00} & \multicolumn{1}{c|}{96.00} & 14.56 \\ \hline
    \multirow{2}{*}{\rotatebox{90}{\texttt{\textbf{Hard}}}} 
    & \multicolumn{1}{|l|}{\textbf{BGE-Large}} & \multicolumn{1}{c|}{28.74} & \multicolumn{1}{c|}{70.00} & \multicolumn{1}{c|}{93.00} & 14.66 \\ \cline{2-6}
     & \multicolumn{1}{|l|}{\textbf{JINA-Large}} & \multicolumn{1}{c|}{27.02} & \multicolumn{1}{c|}{61.00} & \multicolumn{1}{c|}{91.00} & 12.59\\ 
     \specialrule{1pt}{0pt}{0pt}
  \end{tabular}
  }
  \label{difficult_R_results}
\end{table}

\addtocounter{changecounter}{-1}
\begin{table}[t]
  \centering
    % \footnotesize
    \scriptsize
    % \tiny
    \caption{
      \changemark[difficulty_g_table]{
        Performance of different Generation modules on subsets of varying difficulty (Easy/Med./Hard) in HotpotQA, utilizing BGE-Large as the generative model.}
      }
  \renewcommand\arraystretch{1.25}
  \setlength{\tabcolsep}{1.9mm}{
    \begin{tabular}{|c>{\raggedright\arraybackslash}p{2cm}|*{4}{>{\centering\arraybackslash}p{1.05cm}|}}
      \specialrule{1pt}{0pt}{0pt}
    \multicolumn{2}{|l|}{\multirow{2}{*}{\textbf{Difficulty \& Models}}} & \multicolumn{4}{c|}{Conventional Generation Evaluation (ConG Evaluator)} \\ \cline{3-6}% \textbf{ORACLE}  & \multicolumn{1}{c|}{N/A}  & \multicolumn{1}{c|}{N/A}      & \multicolumn{1}{c|}{N/A}   & \multicolumn{1}{c|}{N/A}   & \multicolumn{1}{c|}{} & \multicolumn{1}{c|}{} \\ \hline
    &  & \multicolumn{1}{c|}{\textbf{ChrF}} & 
    \multicolumn{1}{c|}{\textbf{METEOR}} & 
    \multicolumn{1}{c|}{\textbf{R1}} & 
    \textbf{{CER}\(^{\circ}_{\downarrow}\)} \\ \hline        
    \multirow{2}{*}{\rotatebox{90}{\texttt{\textbf{Easy}}}} & \multicolumn{1}{|l|}{\textbf{GPT-4o mini}} & 35.56 & 43.36 & 25.02 & 11.27 \\ \cline{2-6}
     & \multicolumn{1}{|l|}{\textbf{DeepSeek R1-70B}} & \multicolumn{1}{c|}{30.07} & \multicolumn{1}{c|}{37.89} & \multicolumn{1}{c|}{20.55} &18.97 \\ \hline
    \multirow{2}{*}{\rotatebox{90}{\texttt{\textbf{Med.}}}} & \multicolumn{1}{|l|}{\textbf{GPT-4o mini}} & \multicolumn{1}{c|}{32.53} & \multicolumn{1}{c|}{39.70} & \multicolumn{1}{c|}{20.79} & 14.70 \\ \cline{2-6}
     & \multicolumn{1}{|l|}{\textbf{DeepSeek R1-70B}} & \multicolumn{1}{c|}{26.85} & \multicolumn{1}{c|}{34.72} & \multicolumn{1}{c|}{16.56} & 20.22 \\ \hline
    \multirow{2}{*}{\rotatebox{90}{\texttt{\textbf{Hard}}}} & \multicolumn{1}{|l|}{\textbf{GPT-4o mini}} & \multicolumn{1}{c|}{25.42} & \multicolumn{1}{c|}{28.52} & \multicolumn{1}{c|}{16.25} & 13.16 \\ \cline{2-6}
     & \multicolumn{1}{|l|}{\textbf{DeepSeek R1-70B}} & \multicolumn{1}{c|}{22.93} & \multicolumn{1}{c|}{28.12} & \multicolumn{1}{c|}{14.31} & 17.86 \\ 
     \specialrule{1pt}{0pt}{0pt}
  \end{tabular}
  }
  \label{difficult_G_results}
\end{table}

\addtocounter{changecounter}{-1}
\begin{figure}[t]
    \centering
    % First row
    \begin{minipage}[t]{0.32\linewidth}
        \centering
        \includegraphics[width=\linewidth]{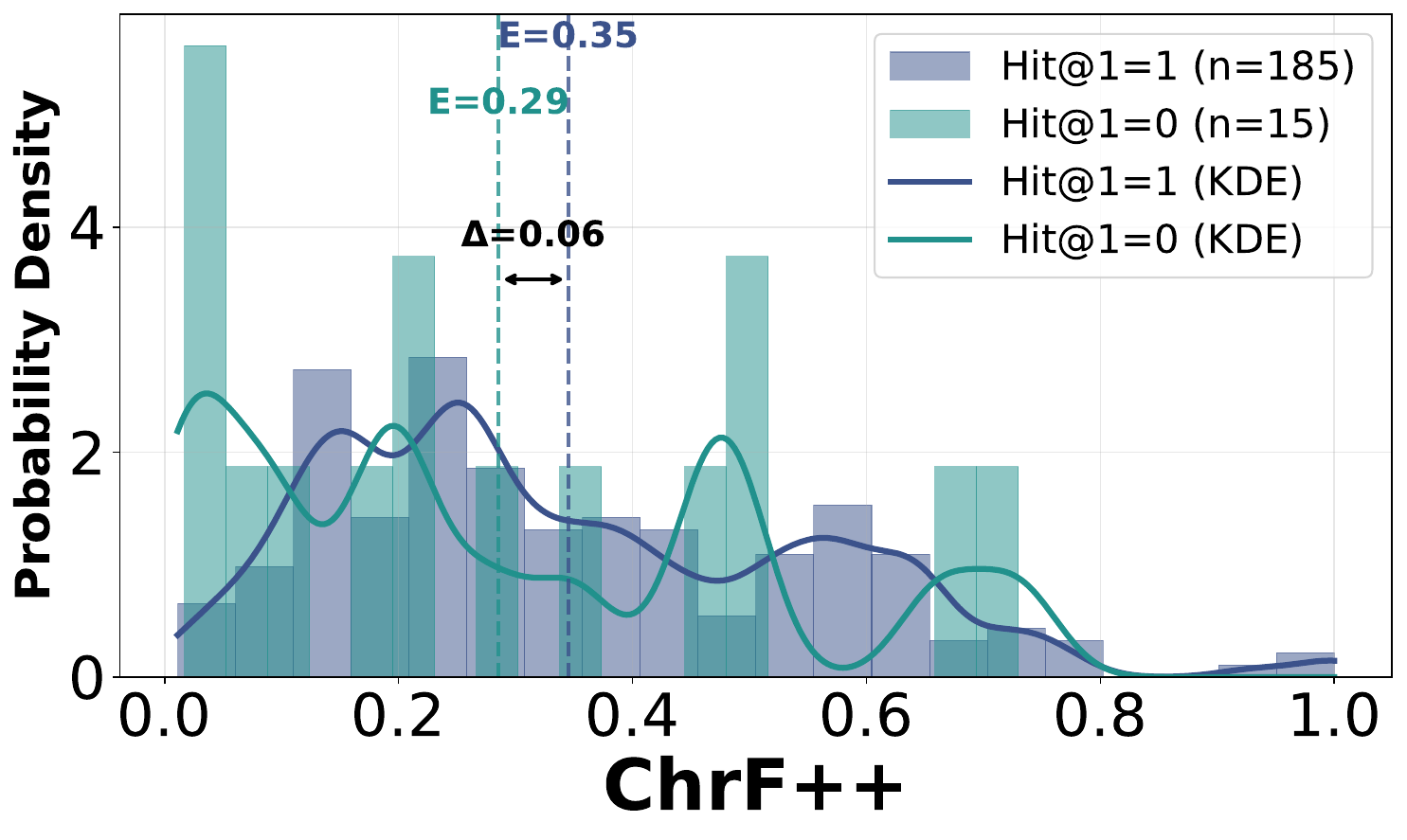}
        \par\vspace{2pt}
        {\scriptsize (a) Easy}
    \end{minipage}
    \hfill
    \begin{minipage}[t]{0.32\linewidth}
        \centering
        \includegraphics[width=\linewidth]{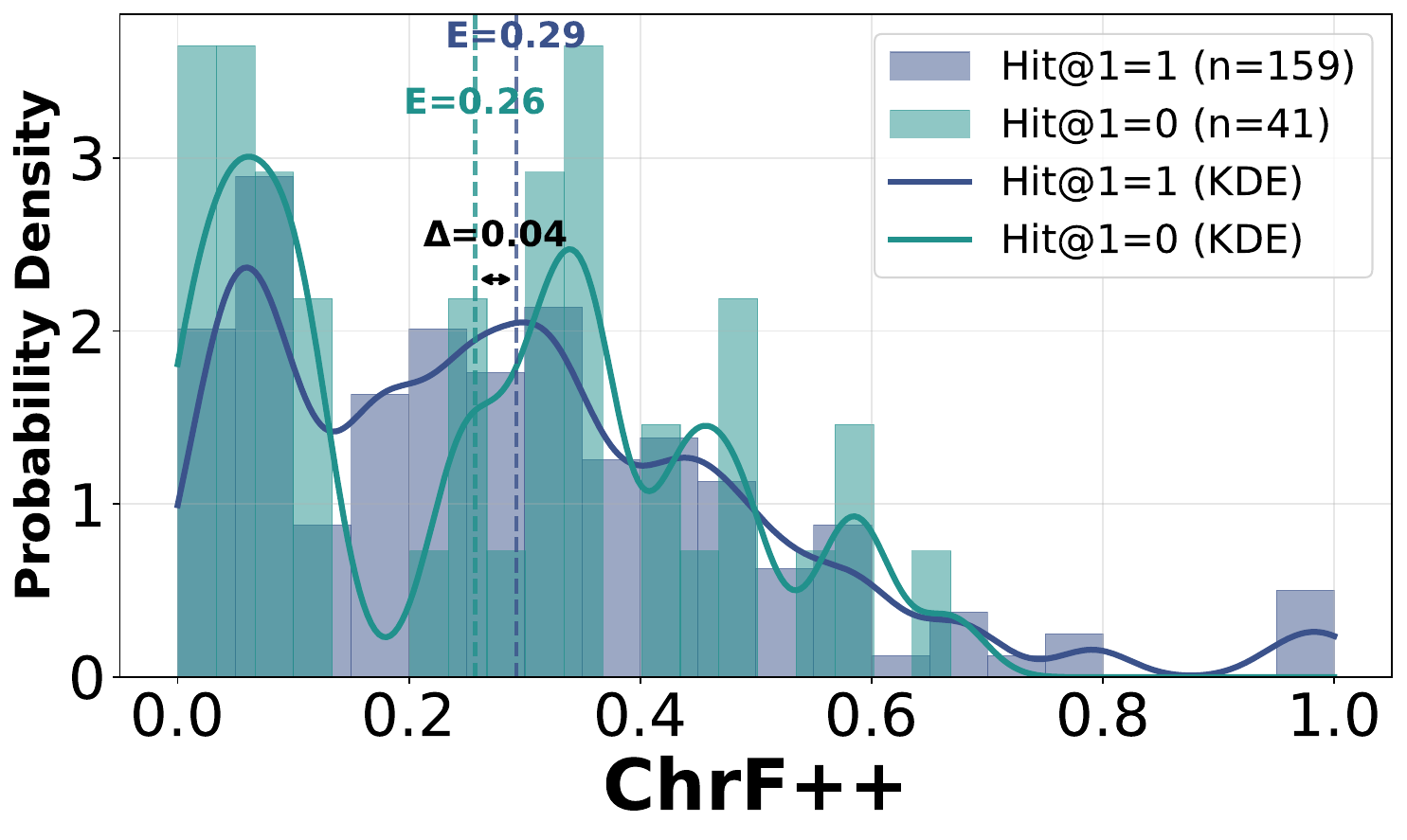}
        \par\vspace{2pt}
        {\scriptsize (b) Medium}
    \end{minipage}
    \hfill
    \begin{minipage}[t]{0.32\linewidth}
        \centering
        \includegraphics[width=\linewidth]{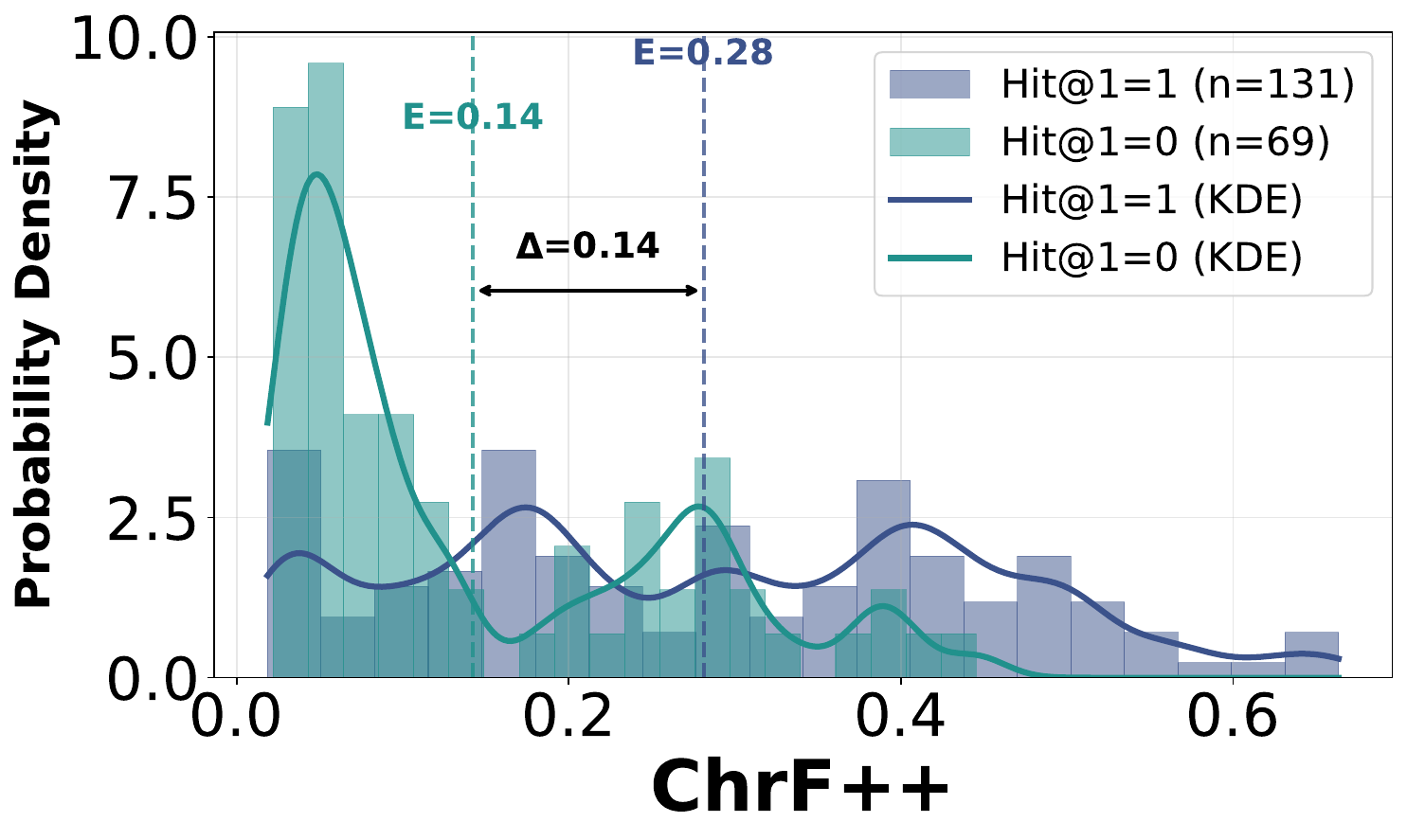}
        \par\vspace{2pt}
        {\scriptsize (c) Hard}
    \end{minipage}    
    \caption{\changemark[Q4_fig]{The relationship between retrieval success and generation quality is examined for the subsets (easy, medium, hard) of HotpotQA, using GPT-4o-mini as the generative model.}}
    \label{dataset_difficulty_fig}
\end{figure}

\addtocounter{changecounter}{-1}
\begin{figure}[t]
    \centering
    % First row
    \begin{minipage}[t]{0.32\linewidth}
        \centering
        \includegraphics[width=\linewidth]{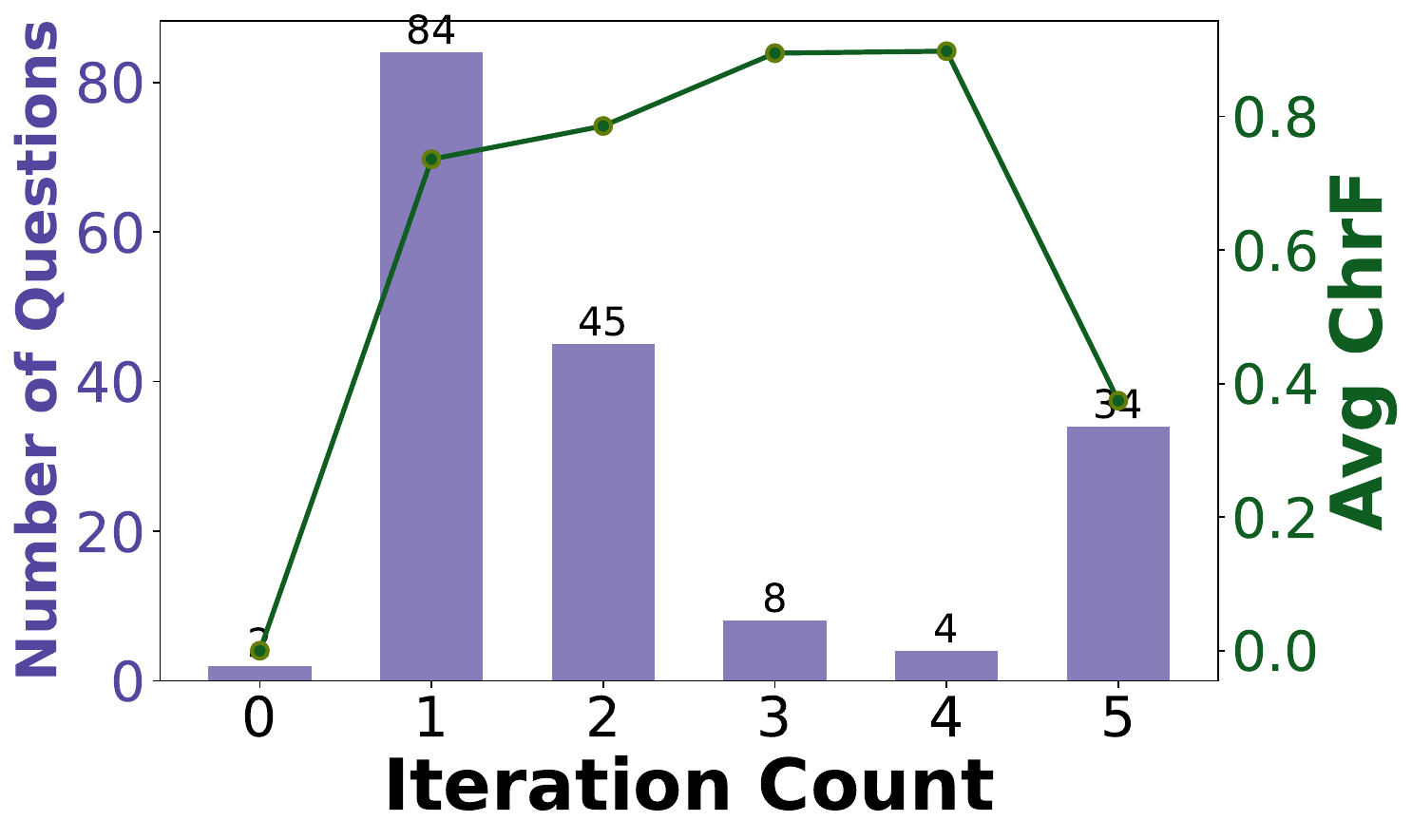}
        \par\vspace{2pt}
        {\scriptsize (a) Easy}
    \end{minipage}
    \hfill
    \begin{minipage}[t]{0.32\linewidth}
        \centering
        \includegraphics[width=\linewidth]{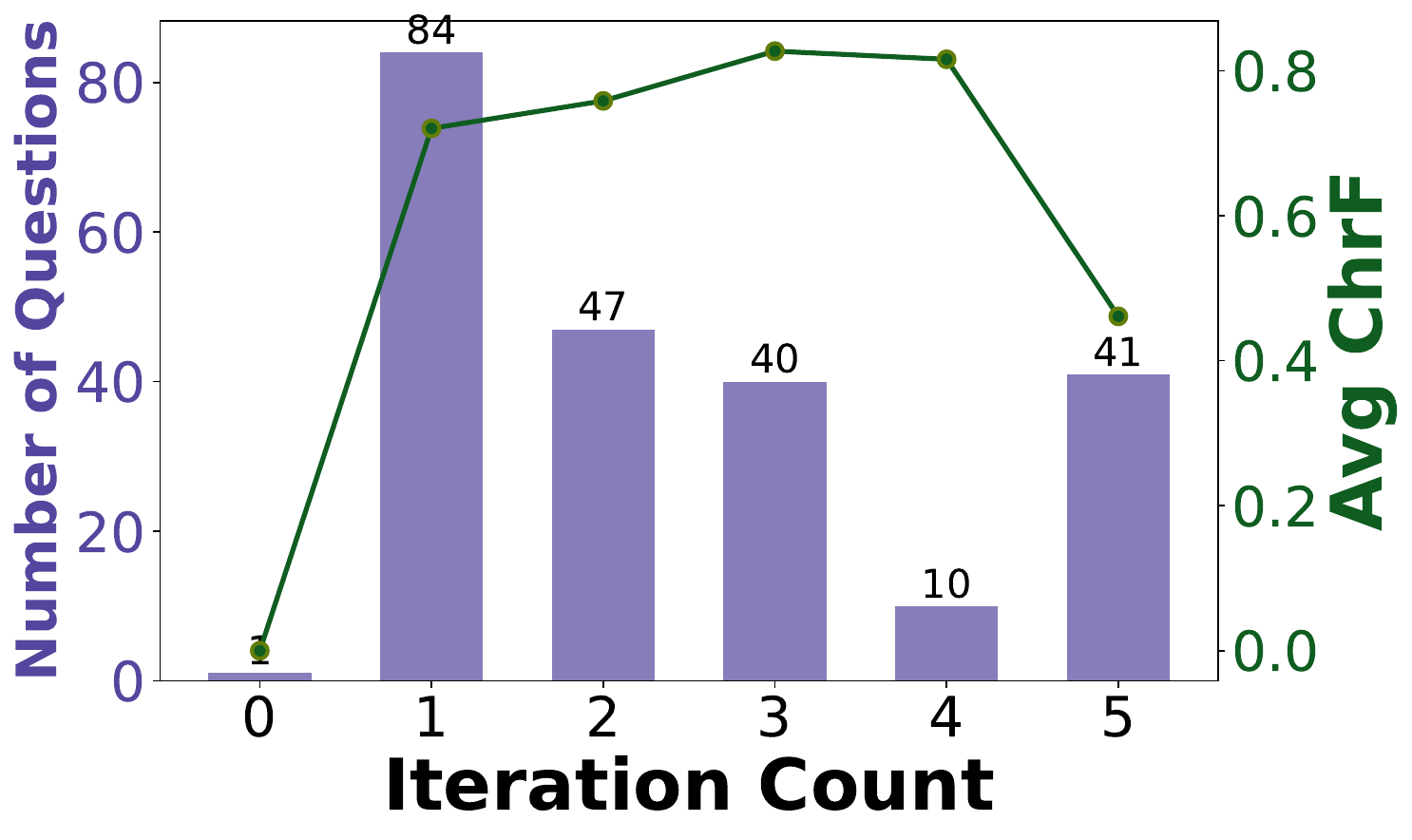}
        \par\vspace{2pt}
        {\scriptsize (b) Medium}
    \end{minipage}
    \hfill
    \begin{minipage}[t]{0.32\linewidth}
        \centering
        \includegraphics[width=\linewidth]{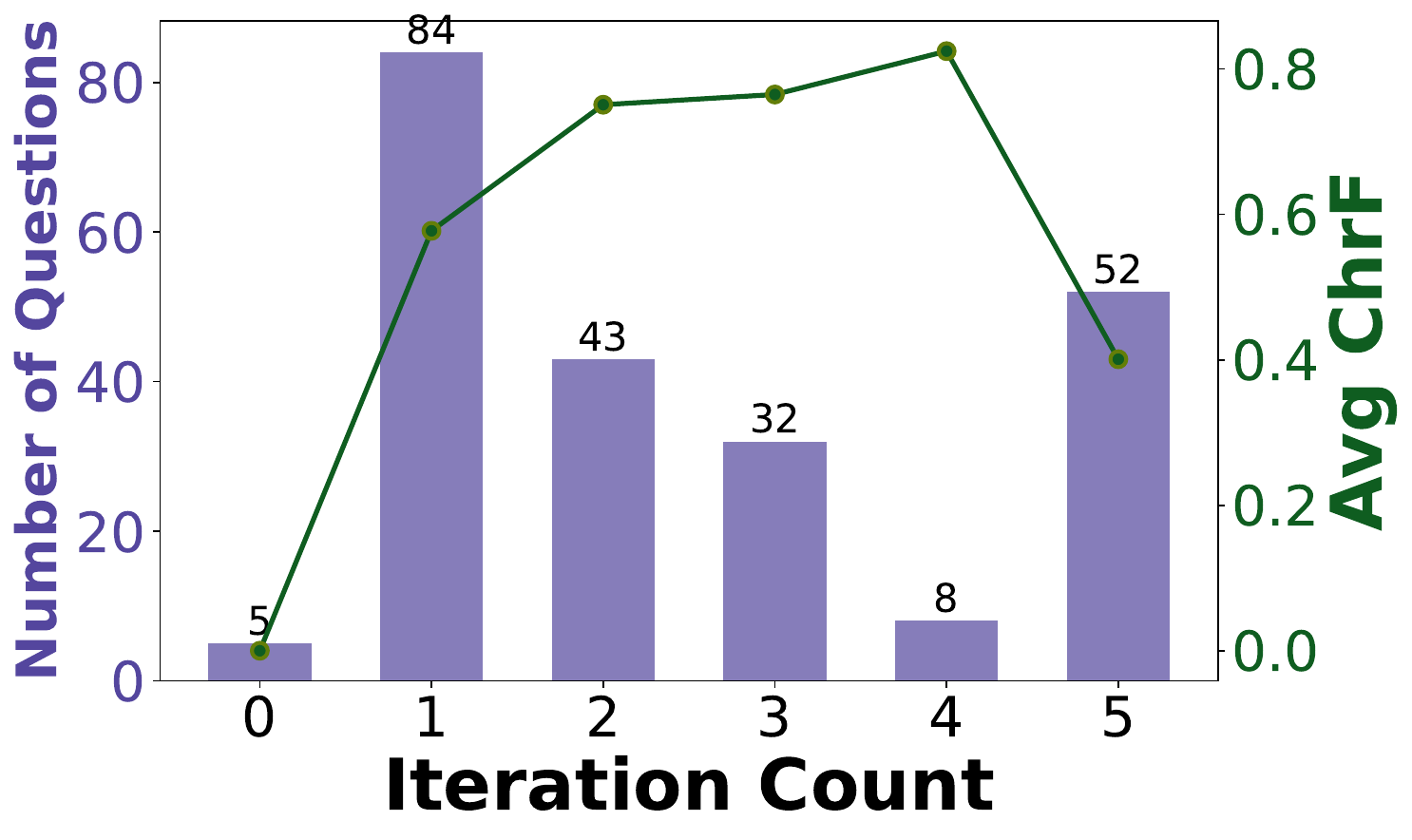}
        \par\vspace{2pt}
        {\scriptsize (c) Hard}
    \end{minipage}    
    \caption{\changemark[Q5_fig]{The relationship between generation quality and SIM-RAG (Maximum iteration limit: 5) iterative rounds is examined for the subsets (easy, medium, hard) of HotpotQA.}}
    \label{iterations_fig}
\end{figure}

\textbf{Q4: Whether superior retrieval results ensures superior generation?}
\changemark[hit_chrf_fig]{
  % 图7中是对HotpotQA数据集不同难度的子集上，检索指标Hit1是否命中和生成指标chrf之间的关系。从图中可以看出，在hard数据集上，当Hit1指标命中时生成指标的期望为0.28比Hit1指标未命中时搞了0.14。这说明，当Hit1命中时，统计上模型回答的更好。但也包含Hit1命中了但生成质量较差的情况，以及Hitler没有命中但是生成质量较好的情况。
  % Figure~\ref{dataset_difficulty_fig} illustrates the relationship between the retrieval metric Hit@1 and the generation metric ChrF across easy, medium, and hard subsets of the HotpotQA dataset. 
  % The results demonstrate that  Hit@1 hits correlate with statistically superior generation performance. 
  As shown in Figure~\ref{dataset_difficulty_fig},on the hard dataset, when the model hits at Hit@1, the mean of Chrf++ metric (0.28) is 0.14 higher than Chrf++ when the model non-hits (0.14).
  This indicates that successful retrieval substantially enhances answer quality for difficult questions.
  % However, the data also reveal instances where Hit@1 hits accompany poor generation quality, as well as cases where Hit@1 misses coincide with satisfactory answers.
  On the easy and medium datasets, the gap in answer quality between Hit@1 hits and non-hits is relatively small (0.06 for easy, 0.04 for medium). Moreover, when Hit@1 hits, there are a small number of samples with poor answer quality (chrF < 0.1), indicating that a good retrieval result does not necessarily yield a correct answer. This explains the phenomenon observed in Section~\ref{Generation Evaluation on ConG}: "Retrieved context outperforms golden context".
}

\textbf{Q5: Whether more iterations of RAG orchestrators 
lead to better generation results?}
\changemark[iteration_fig]{
  % 图8中是在HotpotQA难中易三个子数据集上SIM-RAG迭代次数的实验结果。从图中可以看出，难度较高的数据集需要更多的迭代次数。同时，随着迭代次数的增加，模型的生成质量有明显提升，尤其是在难度较高的数据集上，迭代带来的提升更加明显。这进一步验证了多跳问答任务更适合采用迭代式的orchestrator。
  % Figure~\ref{iterations_fig} presents the experimental results of SIM-RAG iteration counts across easy, medium, and hard subsets of the HotpotQA dataset. 
  As shown in Figure~\ref{iterations_fig}, the number of queries requiring 5 iterations was higher on the hard dataset (52) than on the easy (34) and medium (41) datasets, indicating that more challenging subsets require greater iteration counts. 
  When the iteration limit has not been reached (1$\leq$ count$\leq$4), the quality of answer results improves markedly with increased iterations on the three subsets of HotpotQA dataset. 
  This validates the suitability of iterative orchestration for multi-hop question answering tasks.
  For questions that are too difficult to answer by SIM-RAG, unsatisfactory answers may still be obtained after reaching the maximum iteration limit, which is why performance degrades significantly once the iteration count reaches 5.
}

% \subsection{Cognitive LLM Evaluation (CogL Evaluator)}
% \label{CONGL}
% For the Cognitive LLM Evaluation, token costs are substantial, as they stem from both the input and output tokens in the LLM's reasoning and LLM's testing processes. 
% Rule-based metrics assess performance through `strict' token-overlap criteria, whereas model-based evaluations in  Table~\ref{main_conl_generation_1} and Table~\ref{main_conl_generation_2} offer a `soft' token-match approach. 
% Conventional rule-based metrics like EM and ROUGE focus on n-gram matching and fail to capture the overall accuracy of expression. Rule-based metrics are adequate for tasks requiring unique answers. 

% Conventional rule-based metrics like EM and ROUGE focus on n-gram matching and fail to capture the overall accuracy of expression. Rule-based metrics are adequate for tasks requiring precise and unique answers. Cognitive LLM-based metrics assess generative expression quality, offering an advantage as evaluative understanding is complex to quantify with rule-based metrics.

\section{Conclusion}
\label{conclusion}
This work introduces XRAG, an open-source framework for benchmarking advanced RAG systems, which enables researchers to construct RAG systems more efficiently from datasets. 
It facilitates the selection of appropriate retrieval modules and logic based on data characteristics, and allows for the evaluation of retrieval performance.
% Based on the XRAG framework, the study evaluates state-of-the-art retrieval modules and orchestrators, offering evidence-based recommendations and design insights.
In the future, we plan to enhance XRAG in the following aspects:
% the integration of trainable RAG components to address complex data scenarios;
i. Enable support for RAG component training to handle more complex data scenarios.
ii. Integrate a wider range of benchmark datasets for evaluation, such as OpenQA~\cite{MallenAZDKH23,JoshiCWZ17}, long-form Q\&A~\cite{StelmakhLDC22,MinKLLYKIZH23}, and multiple-choice Q\&A~\cite{HendrycksBBZMSS21,HuangWZGZ022}. 
\changemark[robustness assessment]{iii. Add the robustness evaluation related to dirty and malformed queries.}
% XRAG equips researchers with a unified framework for constructing and benchmarking retrieval-augmented generation modules and offers researchers a standardized RAG framework and evaluation tools, covering around 50 retrieval and generation metrics. 
We encourage contributions from the open-source community to advance the XRAG framework.
Our goal is to refine the XRAG framework by delivering a more efficient and dependable platform with comprehensive evaluation and development tools for RAG research community.

\section*{AI-Generated Content Acknowledgement}
In the preparation of this paper, Deepseek-R1 was used exclusively for the purpose of polishing and refining the linguistic style of certain sentences to improve readability and fluency
in section~\ref{Introduction}, ~\ref{XRAG}, and~\ref{conclusion}. 
All core intellectual contributions, including but not limited to:
\begin{itemize}
\item Original Ideas and Novel Viewpoints: Any innovative concepts, hypotheses, and critical analysis are the product of the authors' own work.
\item Experimental Data and Results: All code, figures, findings, and interpretations thereof were generated, designed, collected, and analyzed by the authors.
\item Substantive Descriptions of the Work: The description of the methodology, experimental procedures, and discussion of the research implications were entirely authored by the research team.
\end{itemize}

\bibliographystyle{./IEEEtran} 
\bibliography{IEEEabrv,ICDE}

\end{document}